\documentclass[10pt,journal,compsoc]{IEEEtran}
\usepackage{comment}
\usepackage{balance}
\usepackage{booktabs} % For formal tables
\usepackage{graphicx}
\usepackage{amsmath, amssymb, amsthm}
\usepackage{thmtools}
\usepackage{bm}
\usepackage[export]{adjustbox}
\usepackage{pythonhighlight}
\usepackage{subcaption}
\usepackage{multirow}
\usepackage{tabularx}
\usepackage{enumitem}
\usepackage{cleveref}
\usepackage{url}
\usepackage{array}
\usepackage{pifont}
\usepackage{pgfplots}
\usepackage{nicefrac}
\usepackage[algo2e,ruled,vlined]{algorithm2e}
\usepackage{algorithm}
\usepackage{dsfont}
\usepackage{framed}
\usepackage{mdframed}

\newenvironment{dbleftbar}{%
  \MakeFramed {\advance\hsize-\width \FrameRestore}}%
 {\endMakeFramed}

\declaretheoremstyle[bodyfont=\normalfont]{mystyle}
\declaretheorem[numbered=no, style=mystyle, name=Example, postheadhook=\begin{dbleftbar}\rule{0pt}{1ex}, prefoothook=\smallskip\end{dbleftbar}]{Ex}
\declaretheorem[numbered=no, style=mystyle, name=Note, postheadhook=\begin{dbleftbar}\rule{0pt}{1ex}, prefoothook=\smallskip\end{dbleftbar}]{note}

\newenvironment{minipeqn}[1][]{\begin{minipage}[#1]{.45\columnwidth}\begin{equation}}{\end{equation}\end{minipage}}

\crefformat{section}{\S#2#1#3}
\crefformat{subsection}{\S#2#1#3}
\crefformat{subsubsection}{\S#2#1#3}

\newcommand{\sd}[1]{\textcolor{black}{#1}}

\newcommand{\dat}{\textsf{DATE}}
\newcommand{\badge}{\textsf{BADGE}}
\newcommand{\bate}{\textsf{bATE}}
\newcommand{\gate}{\textsf{gATE}}
\newcommand{\update}{\textsf{upDATE}}

\usepackage{tikz}

\AtBeginDocument{%
  \providecommand\BibTeX{{%
    \normalfont B\kern-0.5em{\scshape i\kern-0.25em b}\kern-0.8em\TeX}}}

\ifCLASSOPTIONcompsoc
  \usepackage[nocompress]{cite}
\else
  \usepackage{cite}
\fi
\ifCLASSINFOpdf
\else
\fi
\begin{document}
\title{Active Learning for Human-in-the-Loop \\ Customs Inspection}
\author{Sundong~Kim,        
        Tung-Duong~Mai,
        Sungwon~Han,
        Sungwon~Park,
        Thi~Nguyen~D.K,
        Jaechan~So,
        Karandeep~Singh,
        and~Meeyoung~Cha,~\IEEEmembership{Member,~IEEE}% <-this % stops a space
\IEEEcompsocitemizethanks{\IEEEcompsocthanksitem S. Kim, K. Singh, and M. Cha are with the Data Science Group, Institute for Basic Science, Daejeon, Republic of Korea, 34126.
\IEEEcompsocthanksitem T. Mai, S. Han, S. Park, D. Nguyen, J. So, and M. Cha are with KAIST, Daejeon, Republic of Korea, 34141.
\IEEEcompsocthanksitem Contact: Sundong Kim, sdkim0211@gmail.com}% <-this % stops a space
% \thanks{Manuscript received May 31, 2021; revised Nov 27, 2021; accepted Jan 15, 2022.}
} % }}

\IEEEtitleabstractindextext{%
\begin{abstract}
We study the human-in-the-loop customs inspection scenario, where an AI-assisted algorithm supports customs officers by recommending a set of imported goods to be inspected. If the inspected items are fraudulent, the officers can levy extra duties. Th formed logs are then used as additional training data for successive iterations. Choosing to inspect suspicious items first leads to an immediate gain in customs revenue, yet such inspections may not bring new insights for learning dynamic traffic patterns. On the other hand, inspecting uncertain items can help acquire new knowledge, which will be used as a supplementary training resource to update the selection systems. Based on multiyear customs datasets obtained from three countries, we demonstrate that some degree of exploration is necessary to cope with domain shifts in trade data. The results show that a hybrid strategy of selecting likely fraudulent and uncertain items will eventually outperform the exploitation-only strategy.
%
% \medskip
% \sundong{copied from KDD paper}
% Intentional manipulation of invoices that lead to undervaluation of trade goods is the most common type of customs fraud to avoid ad valorem duties and taxes. To secure government revenue without interrupting legitimate trade flows, customs administrations around the world strive to develop ways to detect illicit trades. This paper proposes \algo, a model of Dual-task Attentive Tree-aware Embedding, to classify and rank illegal trade flows that contribute the most to the overall customs revenue when caught. The strength of \algo~comes from combining a tree-based model for interpretability and transaction-level embeddings with dual attention mechanisms. To accurately identify illicit transactions and predict tax revenue, \algo~learns simultaneously from illicitness and surtax of each transaction. With a five-year amount of customs import data with a test illicit ratio of 2.24\%, \algo~shows a remarkable precision of 92.7\% on illegal cases and a recall of 49.3\% on revenue after inspecting only 1\% of all trade flows. We also discuss issues on deploying \algo~in Nigeria Customs Service, in collaboration with the World Customs Organization.
\end{abstract}

\begin{IEEEkeywords}
Customs selection; Fraud detection; Active learning; Online learning; Human-in-the-loop; Import control;
\end{IEEEkeywords}}

\maketitle

% is not selected <OR> if conference mode is selected - because compsoc
% conference papers position the abstract like regular (non-compsoc)
\IEEEdisplaynontitleabstractindextext

% \fi
\IEEEpeerreviewmaketitle

\section{Introduction}
\label{sec:introduction}

Suppose you are given the task of developing an AI-based selection system to assist customs officers working on-site who inspect goods based on recommendations. With the increasing prevalence of online retail, the kinds and amounts of trade traffic are growing astronomically, as are emerging fraudulent trades that try to deceive the system with sophisticated tactics. For example, during the COVID-19 pandemic, the World Customs Organization reported increased numbers of attempted fraud and tax evasion incidents~\cite{wco}. A detection algorithm that is tailored to past logs will inevitably degrade over time. However, relearning the algorithm entirely may waste domain knowledge that has been accumulated over decades. How should the system balance between exploiting existing knowledge and exploring new trends? \looseness=-1

As illustrated in this story, machine learning models in online prediction settings must adapt well to changes in the input, a challenge known as \emph{concept drift}~\cite{conceptdrift2018}. In the context of customs operations, the list of countries procuring a particular product will change over time, and products that are foreign to systems may be declared (e.g., new technology). Even a well-trained machine learning model can fall into the trap of confirmation bias and may not capture these changes. Particularly for situations in which manual labeling is expensive, it can be challenging to make significant changes to the model's working logic. 

To mitigate this problem, \emph{active learning} techniques can help users decide how to query and interactively annotate data points in light of unknown concepts~\cite{settles2009active}. In the recent active learning setting, the model encourages querying uncertain samples while ensuring sample diversity. However, in customs risk management systems, queried data are often subject to evaluation. The system needs to be \emph{profitable} while securing knowledge for the future. Therefore, the learning model cannot fully follow the exploration principles of active learning. To describe the problem, we introduce a case study in which customs administrations maintain an AI-based selection model to support officers' collecting duties.  

\begin{figure}[htpb!]
  \centering
  \includegraphics[width=\linewidth]{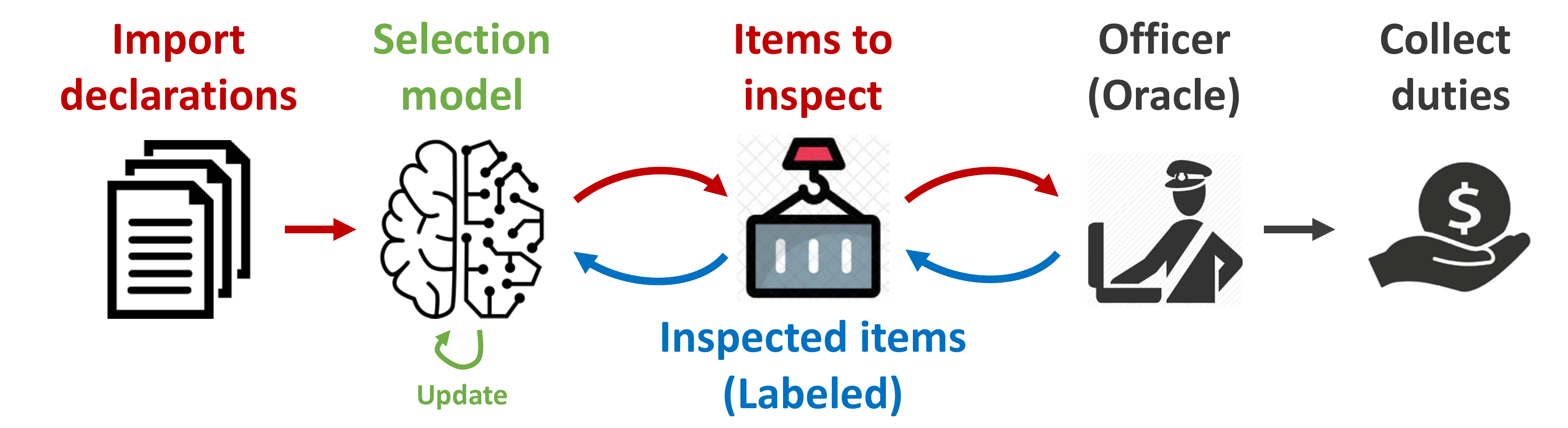}
  \caption{Illustration of the customs clearance process. }
  \label{fig:customs-scenario}
%   \vspace{-5mm}
\end{figure}

Figure~\ref{fig:customs-scenario} depicts a customs clearance process. Importers need to specify the trade items' information in import declaration forms to trade goods across borders. We hypothesize that a trade selection model plays a role in prioritizing items for inspection. Officers follow the recommendation to manually inspect the authenticity of the chosen items and levy additional tariffs if there is fraud. In most customs offices, only a small subset of the import goods (say, 1--5\%) are inspected due to the large trade volume. Once items are inspected, whether there is fraud or not, the performance of the customs offices is evaluated; at this time, the selection model's parameters can be updated with new knowledge. Many customs offices worldwide set aside a small set of random samples to be inspected to learn new fraud patterns~\cite{han2014kcs}. This paper aims to innovate the sample selection strategy, together with the existing inspection process.

%Overall, devising a selection strategy that can maximize the current performance while securing its knowledge for the future is crucial for maintaining the customs selection system in the long run.

We propose a hybrid selection strategy to maximize long-term revenue from fraud detection while maintaining a high income during a short-term inspection. By leveraging the concept of \emph{exploration}, our model remains up to date against concept drifts. In contrast, the concept of \emph{exploitation} maintains high-quality precision for current fraudulent transactions. As an alternative to the random exploration used by customs, we propose the \gate{} exploration methodology, built upon a \emph{state-of-the-art} active learning approach~\cite{Ash2020badge}. \gate{} is designed to select the most informative samples in a diverse way to capture the dynamically changing traits of trade flows. Tested on actual multiyear actual trade data from \emph{three} African countries, we empirically demonstrate that the proposed hybrid model outperforms state-of-the-art models in detecting fraud and securing revenue. 
%Especially, our framework stands out for the country in which the \emph{state-of-the-art} exploitation model failed.  
% Removed newline due to formatting later. %
Our key contributions are as follows:
\begin{enumerate}[leftmargin=*]
\item We define the problem of concept drifts in the context of customs fraud detection, i.e., a dynamic trade selection setting that adaptively acknowledges new trends in data while securing revenue collected from fraud detection.
\item We propose a novel hybrid sampling strategy based on active learning that combines exploration and exploitation strategies.
\item The experiments demonstrate the long-term benefit of exploration strategies on real trade logs from three African countries.
\item We prepare the codes for simulating customs trade selection, considering the needs of customs administration. See the reproducibility section.
%\texttt{\url{http://bit.ly/tkde21-codes}}. % compatible with synthetic dataset 
\end{enumerate}
% Developed a software \ks{are we packaging it as a software?} 

Since 2019, we have collaborated with customs communities represented by the World Customs Organization, and their partner countries, including the Nigeria Customs Service. Namely, our prior work \dat{} classifies and ranks illegal trade flows that contribute most to the overall customs revenue when they are identified~\cite{kim2020date}. \dat{} is open-source\footnote{\url{http://bit.ly/kdd20-date}} and is being studied widely to advance data analytics capabilities in customs organizations. However, in the process of piloting \dat{} in a live environment, we found that concept drift could be fatal to the model performance. Considering that various fraud detection algorithms were tested in an offline setting~\cite{han2014kcs, kultur2017hybrid, vanhoeyveld2020belgian, kim2020date}, it is very likely that the model will suffer from confirmation bias in a live setting. We also show that our \dat{} model suffers from confirmation bias in Figure~\ref{fig:exploitation-fails}(c). Avoiding bias in the model is the primary motivation for extending the research. 

In contrast to our prior work, we propose a \emph{hybrid} algorithm with a new exploration strategy \gate{} and refine our experiments from 80-20\% data splitting to long-term simulation with consecutive inspections. We also tested our algorithm in datasets from multiple countries. \looseness=-1

\section{Related Work}
\label{sec:relatedwork}

\subsection{Customs Fraud Detection} 
Earlier research on customs fraud detection focused on rule-based or random selection algorithms~\cite{han2014kcs, kultur2017hybrid}. While they are intuitive and widely used, these classical methods need to relearn patterns periodically, leading to high maintenance costs. The application of machine learning in customs administration has been a closed task primarily due to the proprietary nature of the data. Several recent studies have shown the use of off-the-shelf machine learning techniques such as XGBoost and the support vector machine (SVM)~\cite{chen2016xgboost, vanhoeyveld2020belgian}. Recently, dual attentive tree-aware embedding (\dat{}) was proposed, employing transaction-level embeddings in customs fraud detection~\cite{kim2020date}. This new model provides interpretable decisions that can be checked by customs officers and yield high revenue through the collected tax.

However, these algorithms are expected to face performance degradation over a long period due to their limited adaptability to uncertainty, diversity, and concept drifts in trade data. We introduce the concept of exploration to remain up to date against concept drifts to address this issue. \looseness=-1 %\smallskip

\subsection{\sd{Concept Drift}}
\sd{Concept drift describes unexpected changes in the underlying distribution of streaming data over time~\cite{conceptdrift2018}. Past research has studied three aspects of concept drift: drift detection, drift understanding, and drift adaptation~\cite{ddm}. Several methods are available for adapting existing learning models to concept drift. The most straightforward way is to retrain a model with the latest data and replace the obsolete model parameters when drift is observed~\cite{paired2008}. In cases with recurring drift, ensemble methods are known to be effective. A classic example involves utilizing tree-based classifier and replacing an obsolete tree with a new tree~\cite{gomes2017adaptive}. Voting schemes have also been applied to manage base classifiers by adding them to ensembles~\cite{xu2017alternating}. The requirement of maintaining a set of pre-defined classifiers is a major drawback of these methods.}

\sd{For stream applications, where only a fraction of the given data is annotated by human effort, one can consider sample selection via active learning to maintain the optimal level of performance. This situation is often subject to the concept drift problem~\cite{zliobaite2014activedrift}. Since our customs trade selection problem also falls under this setting, we consider active learning as a potential solution.
}

\subsection{Active Learning}
Active learning enables an algorithm to elicit ground truth labels for uncertain data instances and enhance its performance~\cite{settles2009active,ren2020activesurvey}. It has been utilized in training models to deal with high-dimensional data~\cite{gal2017bayesian}, to offer long-term benefits~\cite{moon2020icassp,chen2020icml}, to select appropriate data instances to speed up the model training~\cite{song2020cikm}, and to train the model with a limited budget~\cite{zhang2018kdd}.  

For example, one study proposed a way to measure the `informativeness' of given samples~\cite{huang2014tpami}. Others have proposed collecting as much information as possible by prioritizing inspection of uncertain samples~\cite{houlsby2011bald, gal2016uncertainty, yoo2019loss}. 
Another line of research has focused on improving diversity by strategically collecting samples to represent the overall data distribution. Diversity-based algorithms include region-based active learning~\cite{cortes2020icml} and core-set-based approaches~\cite{sener2018coreset}. Recent research has also focused on the concurrent inclusion of uncertainty and diversity aspects~\cite{Ash2020badge,zhdanov2019diverse}. \looseness=-1

Collectively, these approaches share common limitations in practical use. First, active learning research has assumed an offline setting~\cite{kirsch2019batchbald, song2020ada, abbas2020active, Ash2020badge}. For example, \textsf{HAL} showed that including simple exploration helps margin sampling in a skewed dataset~\cite{abbas2020active}, and \textsf{BADGE} showed the effectiveness of sampling uncertain data points in a diverse way~\cite{Ash2020badge}. However, their evaluations are based on fixed test data, which cannot accommodate concept drifts in real active learning scenarios. Real-world logs exhibit substantial changes and dynamics over time, as we will demonstrate in this paper, making most static machine learning models obsolete. \looseness=-1

% To deploy in real-world customs practice, the proposed algorithm should perform robustly in \emph{dynamic} settings. The underlying data distribution often changes in the presence of concept drift, and the machine learning model can become easily outdated.

Second, extant models separate the processes of exploitation (i.e., inspection) and exploration (i.e., annotation). In practice, every manual inspection or annotation is a cost in which the budget is often limited (for example, by inspection officers in customs). Given the shared budget, an exploration-oriented active learning algorithm is unlikely to succeed if it is learned separately. This constrained optimization setting has not been handled in conventional approaches. Our work addresses these two realistic settings.

\section{Customs Trade Selection}
\label{sec:problemdefinition}
% The objective of customs administration

The customs administration aims to detect fraudulent transactions and maximize the tax revenue from illicit trades --- this is the customs fraud detection problem~\cite{vanhoeyveld2020belgian, kim2020date}. Given an import trade flow $\mathcal{B}$, the main goal is to predict both the fraud score $y^{cls}$ and the raised revenue $y^{rev}$ obtainable by inspecting each transaction $\mathbf{x}$. 
Given the limited budget of inspection and annotation, we address the problem of devising an efficient selection strategy to identify suspicious trades and increase revenue as follows:
%, which we define as the \emph{customs trade selection} problem as follows:
%
%However, as transaction volume is huge, customs administration cannot practically screen all suspicious transactions. While some customs offices had conducted 100\% manual inspection, this has become costly and time-consuming with astronomically growing trade volume. Furthermore, customs offices are most recently compelled to cut down on many intensive operations in light of the health pandemic and social distancing. There is a growing need to develop an efficient selection strategy to identify suspicious trades and increase revenue. We formulate the \emph{customs trade selection} problem as follows:
%
\begin{quote}
%\begin{Thm}
\textbf{Customs trade selection problem}.~~~\\ Given trade flows $\mathcal{B}$, construct a sample selection strategy $f$ that maximizes the detection of fraudulent transactions and the associated tax revenue.
%\end{Thm}
\end{quote}

\noindent
The trade flows $\mathcal{B}$ consist of the online stream of trade records\footnote{These terms are used interchangeably: transactions, items, goods.}, including the importer ID, commodity ID (such as the Harmonized System Codes), and declared price of goods. The characteristic distributions of illicit transactions are assumed to change over time (see  Sec.~\ref{sec:experiments:settings:datasets}).
% \begin{table}[h!]
% \centering
% \caption{Notations used throughout the paper.}
% \label{tab:notations}
% % \resizebox{\linewidth}{!}{%
% \small
% \begin{tabular}{p{0.1\columnwidth}| p{0.81\columnwidth} } \toprule
%     Symbol & Definition \\ \midrule
%     $\mathcal{B}$ & An import flow. \\
%     $\mathbf{x}$ & An imported item from an import flow $\mathcal{B}$. \\
%     $y^{cls}$ & A binary label denoting the item $\mathbf{x}$ is fraud. \\
%     $y^{rev}$ & A non-negative value (label) denoting the item $\mathbf{x}$'s additional revenue upon its inspection.\\  \midrule
%     $\mathcal{B}_t$ & Imported items arrived during an unit interval before time $t$. \\ 
%     $\mathcal{B}_t^{S}(f)$ & Items selected by strategy $f$ at time $t$. These items are subject to inspection. After inspection, their labels $\bm{y}^{cls}$ and $\bm{y}^{rev}$ are obtained. \\ 
%     $\mathcal{B}_t^{F}(f)$ & A subset of $\mathcal{B}_t^{S}(f)$ in which items are considered as frauds. \\
%     $\mathcal{B}_t^{U}(f)$ & A subset of $\mathcal{B}_t^{S}(f)$ in which items are considered uncertain. \\ \midrule
%     $f$ & A customs selection strategy. \\
%     $X_t$ & Training data at time $t$. $X_t$ is used to update the parameters of the strategy $f$.  \\
%     $r_t$ & An inspection (selection) rate at time $t$. \\
%     $ m $ & An evaluation metric for $\mathcal{B}_t^{S}(f)$, (e.g., \textsf{Revenue@k\%}). \\
%     \bottomrule
% \end{tabular}
% % }
% \end{table}

\begin{table}[h!]
\centering
\caption{Notation used throughout the paper.}
\label{tab:notations}
\resizebox{\linewidth}{!}{%
\small
\begin{tabular}{p{0.09\columnwidth} p{0.91\columnwidth} } \toprule
    Symbol & Definition \\ \midrule
    $\mathcal{B}$ & Import flows. \\
    $\mathbf{x}$ & A transaction (item) from import flows $\mathcal{B}$. \\
    $y^{cls}$ & A binary label denoting that item $\mathbf{x}$ is fraudulent. \\
    $y^{rev}$ & A non-negative value (label) denoting item $\mathbf{x}$'s additional revenue upon its inspection.\\  \midrule
    $\mathcal{B}_t$ & Import flows arriving during an unit interval before time $t$. \looseness=-1 \\ 
    $\mathcal{B}_t^{S}(f)$ & A set of items selected by strategy $f$ at time $t$. These items are subject to inspection. After inspection, their labels $\bm{y}^{cls}$ and $\bm{y}^{rev}$ are obtained. For simplicity, we denote it as $\mathcal{B}_t^{S}$.\\ 
    $\mathcal{B}_t^{F}$ & A subset of $\mathcal{B}_t^{S}$ in which items are considered fraudulent. \\
    $\mathcal{B}_t^{U}$ & A subset of $\mathcal{B}_t^{S}$ in which items are considered uncertain. \\ \midrule
    $f$ & A customs selection strategy. \\
    $X_t$ & Training data at time $t$. $X_t$ is used to update the parameters of strategy $f$.  \\
    $r_t$ & An inspection (selection) rate at time $t$. \newline ($0\% \leq r_t \leq 100\%$) \\
    $ m $ & An evaluation metric for $\mathcal{B}_t^{S}(f)$. \newline (e.g., \textsf{Revenue@k\%}). \\
    \midrule
\end{tabular}
}
\footnotesize
\begin{tabular*}{\columnwidth}{ll} 
 \begin{minipeqn}
  \mathcal{B}_t^{S} = \{x | x \in \mathcal{B}_t; f\},
\end{minipeqn}&
\begin{minipeqn}[b]
  \mathcal{B}_t^{S} \subset \mathcal{B}_t,
\end{minipeqn}\\
\begin{minipeqn}
  |\mathcal{B}_t^{S}| = r_t\cdot|\mathcal{B}_t|,
\end{minipeqn}&
 \begin{minipeqn}[b]
    \mathcal{B}_t^{U} \subset \mathcal{B}_t^{S},
\end{minipeqn}\\
\begin{minipeqn}
  \mathcal{B}_t^{F} \subset \mathcal{B}_t^{S},
\end{minipeqn}&
 \begin{minipeqn}[b]
  \mathcal{B}_t^{S} = \mathcal{B}_t^{F} \cup \mathcal{B}_t^{U},
\end{minipeqn} \\
\begin{minipeqn}
  \mathcal{B} = \bigcup_{t} \mathcal{B}_t,
\end{minipeqn}&
 \begin{minipeqn}[b]
  X_t = \bigcup\limits_{t'<t} \mathcal{B}_{t'}^{S}(f).
\end{minipeqn} \\ \bottomrule
\end{tabular*}
\end{table}

% \noindent
% \end{table}

% \vspace{-5mm}

% \begin{Property}  % \begin{dwf} % \end{dwf}
% \text{Eq.~\ref{} are several properties using notations in Table~\ref{tab:notations}.}
% \noindent
% \begin{tabularx}{\textwidth}{@{}XXX@{}}
% % \caption{}
% \label{tab:properties}
% \vspace{-6pt}
%   \begin{equation}
%   \mathcal{B}_t^{S} = \{x | x \in \mathcal{B}_t; f\},
%   \qquad
%   \mathcal{B}_t^{S} \subset \mathcal{B}_t,
%   \qquad
%   |\mathcal{B}_t^{S}| = r_t\cdot|\mathcal{B}_t|,
%   \end{equation}
% %   \vspace{0.5pt}
%   \begin{equation}
%   \mathcal{B}_t^{U} \subset \mathcal{B}_t^{S},
%   \qquad
%   \mathcal{B}_t^{F} \subset \mathcal{B}_t^{S},
%   \qquad
%   \mathcal{B}_t^{S} = \mathcal{B}_t^{F} \cup \mathcal{B}_t^{U},
%   \end{equation}
% %   \vspace{0.5pt}
%   \begin{equation}
%   \mathcal{B} = \bigcup_{t} \mathcal{B}_t,
%   \qquad
%   X_t = \bigcup\limits_{t'<t} \mathcal{B}_{t'}^{S}(f).
%   \end{equation}  
% \vspace{-5mm}
% \end{tabularx}
% \end{Property}

    % |\mathcal{B}_t^{S}| &= r_t\cdot|\mathcal{B}_t| \\
    % \mathcal{B}_t^{U} & \subset \mathcal{B}_t^{S} \\
    % \mathcal{B}_t^{F} & \subset \mathcal{B}_t^{S} \\
    % \mathcal{B}_t^{S} &= \mathcal{B}_t^{F} \cup \mathcal{B}_t^{U} \\
    % X_t &= \bigcup\limits_{t'<t} \mathcal{B}_{t'}^{S}(f) \\
%\vspace{-5mm}

\subsection{Active Selection for the Online Setting}
\label{sec:active}
% This is a section for describing active learning,~\cite{vanhoeyveld2020belgian, kim2020date}

Extant research on customs fraud detection mainly concentrates on the static setting, in which a model is trained on large training batches and deployed for fraud detection without further updates~\cite{vanhoeyveld2020belgian, kim2020date}. We consider a practical online setting where the characteristic distribution of trade flows $\mathcal{B}$ and the traits of illicit trades change over time. This is the case for the \textbf{active customs trade selection problem}. The active customs selection problem requires the selection strategy to help the model update and adapt for new fraud types. All inspected items can bring additional information, and strategically choosing the right items to maximize the model performance is handled in this problem. 

%Unlike the customs fraud detection problem that only requires selecting the maximum fraudulent transactions for every timestamp, the active customs selection problem also requires the selection strategy to help the model update and adapt for new types of fraud. All inspected items can bring additional information, but choosing items that aid the model performance with the selection strategy to improve the model's future performance is crucial in this problem. 

We formally define the active customs trade selection problem as follows: At each time $t$, given a batch of items $\mathcal{B}_t$ from trade flows $\mathcal{B}$, based on a strategy $f$ trained with $X_t$, customs officers select a batch of items $\mathcal{B}_t^S$ to inspect physically. After inspection, the annotated results are used to update the strategy $f$. We evaluate the model from timestamp $t_k$ onward. The goal is to devise a \emph{strategy} \textit{$f^*$} that maximizes the precision and revenue in the long-term:
% \begin{equation}
% f^* = \operatorname*{argmax}_f{\dfrac{1}{T-t_0+1}\sum_{t=t_0}^{T}{m(\mathcal{B}_t^s(f))}}
% \end{equation}
\begin{equation}
f^* = \operatorname*{argmax}_f{\sum_{t\geq t_k}{m(\mathcal{B}_t^s(f))}},
\end{equation}
where $m$ is the evaluation metric, which is the precision or revenue from fraud detection. Table~\ref{tab:notations} lists the notation used frequently throughout the paper, and the main training process for fraud detection with active customs selection is described in Algorithm~\ref{alg:active_custom_select}.
\begin{algorithm}[h]
\small
\DontPrintSemicolon
\SetNoFillComment
\begin{flushleft}
\SetAlgoLined
\smallskip
\textbf{Input:} Previous inspection histories $\mathcal{H}$, initial inspection rate $r_0$, target inspection rate $r$, unlabeled datastream of new items in each timestamp $\mathcal{B}_t$ \\
\textbf{Output:} Items for inspection $\mathcal {B}_t^S$ in each timestamp t

\smallskip

{\small \tcc{Considering a weekly inspection is made.}}

Initialize the training data $X_1$ from inspection histories $\mathcal{H}$;

\For{$t = t_1,\cdots,$}{
Obtain the batch of new items $\mathcal{B}_t$;

Determine the weekly inspection budget $r_t$, using $r_0$ and $r$;

{\small \tcc{Selection by the algorithm}}
Train the selection strategy \textit{f} with $X_t$; 

Based on \textit{f}, select a set $\mathcal {B}_t^S$ of $r_t|\mathcal{B}_t|$ items for inspection; 
{\small \tcc{Inspection by officers}}
Obtain the ground-truth annotation $(\textbf{x}_i, y_i^{cls}, y_i^{rev})$ for each item $\textbf{x}_i \in \mathcal{B}_t^S$ after manual inspection;

Evaluate the results by precision and revenue;

Add the newly annotated items into the training data:
$X_{t+1} = X_t \cup \mathcal{B}_t^S$;
  
}
\caption{Active Customs Trade Selection}
\label{alg:active_custom_select}
\end{flushleft}
\end{algorithm}

\begin{figure*}[!t]
  \centering
  \includegraphics[width=\linewidth]{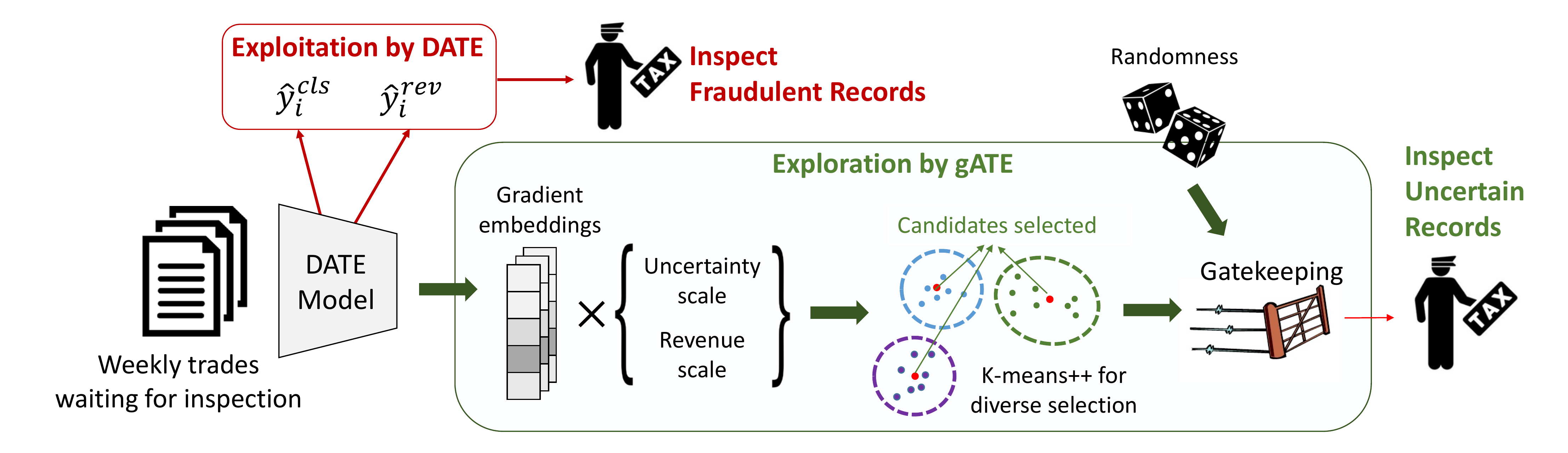}
  \caption{Illustration of the hybrid selection framework. 
  The state-of-the-art exploitation model \dat{} first computes whether the input trade records are fraudulent. Then, we backpropagate it with pseudo-labels to generate a gradient embedding and rescale it with its uncertainty and revenue. Then, the $k$-means++ algorithm is applied to select diverse yet uncertain samples for inspection. A gating unit decides which items to explore.
  }
  \label{fig:hybrid_architecture}
\end{figure*}
\section{Hybrid Selection Strategy}
\label{sec:model}
% Section for describing specific strategy (proposed)
% Introduce how the \update{} algorithm works with small illustrative examples. Provide a summary of the algorithm: The algorithm accentuates sample uncertainty and diversity. 
The quality of the active customs trade selection problem depends on having a good selection strategy \textit{f}. We propose a new strategy that combines two approaches: \textit{exploitation} and \textit{exploration}. The exploitation approach selects the most likely fraudulent and highly profitable items to secure short-term revenue for customs administration. The exploration approach, in contrast, selects uncertain items at the risk of temporary revenue regret, yet potentially detects novel fraud patterns in the future. Our algorithm mixes these two components to gain long-term benefits and secure immediate revenue from imbalanced customs datasets. Figure~\ref{fig:hybrid_architecture} illustrates the overall framework of the proposed model.

\subsection{Exploitation Strategy: Customs fraud detection tailored to maximize the short-term revenue}
\label{sec:model:exploit}
We employ the current state-of-the-art algorithm in illicit trade detection, \dat{}~\cite{kim2020date}, as a baseline for this research. It is a tree-enhanced dual-attentive model that optimizes the dual objectives of (1) illicit transaction classification and (2) revenue prediction. We leverage the predicted fraud score of \dat{} for our exploitation strategy. We update the \dat{} model at each timestamp and select the most suspicious items as per the inspection budget (see Algorithm~\ref{eq:date_exploit}).

% \brian{Selected items are added into the training set for the next update.}

\begin{algorithm}[H]
\small
\SetNoFillComment
\begin{flushleft}
\smallskip
 \textbf{Input:} Training set $X_t$, items received $\mathcal{B}_t$, inspection rate $r_t$ \\
 \textbf{Output:} A batch of selected items $\mathcal{B}_t^S$ \\
 \SetAlgoLined
 \tcc{Corresponds to the selection part in Alg.~\ref{alg:active_custom_select}.}
 Train the \dat{} model using training set $X_t$;\\
 Perform prediction on $\mathcal{B}_t$, obtain the predicted annotation $(\textbf{x}_i, \hat{y}_i^{cls}, \hat{y}_i^{rev})$ for each item $\textbf{x}_i \in \mathcal{B}_t$;\\
 Obtain the set $\mathcal{B}^S_t$ of $r_t|\mathcal{B}_t|$ items with the highest fraud score $\hat{y}_i^{cls}$;\\
 \caption{Exploiting suspicious items by \dat{}}
 \label{eq:date_exploit}
\end{flushleft}
\end{algorithm}

\subsection{Exploration Strategy: Customs fraud detection adapt to concept drift and aiming at long-term revenue}
\label{sec:model:exploration}

The exploitation strategy selects the more familiar and highly suspicious transactions for inspection; therefore, it tends to underperform over time as trade patterns gradually change. In contrast, our hybrid strategy chooses to add a small portion of new and uncertain trades as a learning sample in the training data, which gradually affects the model's future prediction performance. Since fraud types are constantly evolving, the model performance might drop over time. We propose an exploration strategy to select uncertain trade items, with additional consideration of diversity and revenue, to resolve these issues.

\subsubsection{Exploration in light of uncertainty and diversity}
\label{sec:model:embedding} 
% Alleviating the unbalance between uncertainty and diversity is the key to active learning.
One approach to detecting new fraud types is to utilize uncertainty in the query strategy. Selecting items for which the model is least confident can provide more information on similar new observations. However, this strategy can create an unfavorable scenario where newer labeled data do not include diverse transaction types and labels for identical transactions continue to accumulate. Considering this, we include the concept of diversity along with uncertainty in our selection strategy; i.e., we choose the most diverse samples possible for stable and fast exploration~\cite{kirsch2019batchbald,Ash2020badge}.
We take in gradient embedding and $k$-means++ initialization from \badge{}~\cite{Ash2020badge} in our exploitation model \dat{} to determine which trades should be queried for inspection by considering uncertainty and diversity concepts. The detailed implementation of each concept is described below. \smallskip
%  \badge{} \cite{Ash2020badge}, the state-of-the-art active learning query strategy utilizes the gradient embedding and $k$-means++ initialization to choose groups of points that are both diverse and uncertain. 

% In the DATE model, after the fusion layer, the network performs dual task learning (i.e. illicitness and revenue prediction) with task-specific layers \cite{kim2020date} (Figure~\ref{fig:date_architecture}). 

\noindent\textbf{Uncertainty.} 
If a sample generates a large gradient loss and, consequently, a large parameter update, the item potentially contains useful information. This means that the magnitude of \emph{gradient embedding} reflects the uncertainty of the model on samples. With this motivation, we aim to choose trade flows with uncertainty using the magnitude of gradient embedding. At time $t$, for each trade item $\textbf{x}_i$ in $\mathcal{B}_t$, the illicitness classifier $h_{\theta}$ from the \dat{} model returns its fraud score $\hat{y}_i^{cls}$, which indicates the illicit class of $y_i^{cls}$.
% Neural network $h$ is parameterized by the weights $\theta = (W,V)$ where $W$ is the weight of last layer and $V$ includes weights of all previous layers.
\begin{equation}
     h_{\theta}(\textbf{x}_i) = \sigma(W \cdot z_{\phi}(\textbf{x}_i)), 
\end{equation}
% where $z(x, V)$ is the output after the intermediate layer and $\sigma$ is a sigmoid function. 
where $W$ is a weight matrix that projects the transaction embedding $z_{\phi}$ to the \dat{} illicitness class space.

% For an item $x$, probability scores assigned to candidate labels 1 (illicit) and 0 (licit) is $p_1 = f(x,\theta) = e^{W\cdot z(x, V)}[1+e^{W\cdot z(x, V)}]^{-1}$ and $p_0 = 1 - f(x,\theta) = [1+e^{W\cdot z(x, V)}]^{-1}$ respectively.
% \begin{equation}
% p_c = \dfrac{e^{I(c=1)W\cdot z(x, V)}}{1+e^{W\cdot z(x, V)}} \text{ for } c \in \{0, 1 \}
% \end{equation}
% Cross-entropy loss is calculated by $l_{CE}(f(x;\theta), c) = - \ln(p_c)$:
% \begin{equation}
%     l_{CE}(f(x;\theta), c) = \ln[1+ e^{W\cdot z(x,V)}] - I(c=1)W\cdot z(x, V)
% \end{equation}

The gradient embedding $g_{x_{i}}$ is the gradient of the loss function with respect to $W$ and sample $x_{i}$. Since the received data points are unlabeled (not yet inspected), we predict the pseudo label $\hat{c}_i$ by the fraud score with a threshold of 0.5 (i.e. $\hat{c}_i = \mathds{1}(\hat{y}_i^{cls} \geq 0.5)$). This pseudo label is used to calculate the loss, resulting in the gradient embedding described as:
\begin{equation}
    g_{x_{i}}^{c} = (p_i^c - \mathds{1}(\hat{c}_i = c))
    \cdot z_{\phi}(\textit{x}_i),
\end{equation}
where $c \in \{0,1\}$ corresponds to the two classes and $p_i^c$ is the predicted probability for class c; $p_i^{c=0} = 1 - \hat{y}_i^{cls}$ and $p_i^{c=1} =  \hat{y}_i^{cls}.$ \smallskip

\noindent\sd{\textbf{Diversity}. 
For effective exploration, we inspect items that bring large changes and diverse views to the model. Taking diversity into account, we can avoid a situation where similar items are inspected redundantly. We suggest a batch construction method by selecting the most representative samples through the k-means++ algorithm~\cite{arthur2007kmeans++}, which produces a good initial clustering situation. $K$-means++ obtains a set $\mathcal{B}^s_t$ of $k$ centroids sampled in proportion to the nearest set’s centroids. Samples with small gradients are also unlikely to be chosen, as the distances between them are small. Gradient embedding with $k$-means++ seeding tends to result in the selection of a batch of large and diverse gradient samples. By doing so, our selection strategy can consider both sample uncertainty and batch diversity. 
}

% \noindent\textbf{Diversity}. We create a batch of query items based on gradient embedding with the $k$-means++ algorithm~\cite{arthur2007kmeans++} in connection with the diversity. 
% % In this way, we consider both the magnitude of the gradients and their diverse direction. 
% We obtain the set $\mathcal{B}^s_t$ of $k$ centroids that are sampled with probability proportional to the nearest sets' distances to take diversity into account.
% % of returning centroids are ensured to approximate $k$-means++ objective
% % Hence, we can take diversity into account. 
% Samples with small gradients are also unlikely to be chosen, as the distance between them is small. Thus, gradient embedding with $k$ means++ seeding tends to choose a large, diverse gradient sample.

\subsubsection{Scale uncertainty and revenue effect}
\label{sec:model:bate}
To induce the algorithm to select more uncertain and high-revenue items, we introduce extra weights to amplify the effect of uncertainty and revenue. These weights, called the uncertainty scale and revenue scale, adjust the probability of chosen samples by resizing their gradient embedding vectors. \smallskip

\noindent\textbf{Uncertainty scale.} We magnify the impact of uncertain items by quantifying the model's ability to calibrate an item. We give each item an \emph{uncertainty score} (Eq.~\ref{eq:scale_unc}) such that the score indicates the magnitude of the model's uncertainty about the item. The uncertainty score $unc_i$ is defined as follows:
\begin{equation}
    unc_i = -1.8 \times |\hat{y}_i^{cls}-0.5| + 1. \\
    \label{eq:scale_unc}
\end{equation}
\sd{This concave function maximizes the uncertainty score when the system cannot determine whether an item is fraudulent or not (i.e., the uncertainty score is the largest when the predicted fraud score $\hat{y}_i^{cls}$ is 0.5). We adjust the uncertainty score by using a multiplier -1.8 and set its range between 0.1 and 1, which leads the exploration algorithm to select every item with some chance.\footnote{This setting shows the best results on our datasets. For practical use, the best parameters can be found using the validation set.} Our intention here is to leave some degree of uncertainty even when the base model is overconfident about its prediction results (i.e., $\hat{y}_i^{cls}$ is close to 0 or 1)~\cite{guo2017calibration}}. 

\noindent\textbf{Revenue scale.} Active learning in customs operation requires additional consideration, as revenue needs to be collected as the customs duty. Maximizing the customs duty is one of the top priorities of customs authorities. Therefore, we further amplify the gradient embedding by the \dat{} model's predicted revenue $\hat{y}_i^{rev}$. The distribution of the amount of customs duty is right-skewed, so we take the \emph{log} of the predicted revenue (Eq.~\ref{eq:scale_rev}). We can define the final scale factor $S_i$ of $\textbf{x}_i$ as 
\begin{equation}
    S_{i} = unc_i \cdot log(\hat{y}_i^{rev} + \epsilon).
    \label{eq:scale_rev}
\end{equation}
As a result, the gradient embedding $g_{x_{i}}^{c}$ becomes
\begin{equation}
\label{eqn:embedding}
    g_{x_{i}}^{c}= S_{i} \cdot (p_i^c - \mathds{1}(\hat{c} = c))\cdot z_{\phi}(\textbf{x}_i).  
\end{equation}
$k$ is a constant for computational stability. The algorithm covered in Sections~\ref{sec:model:embedding} to \ref{sec:model:bate} is named \bate{}, inspired by \badge{}~\cite{Ash2020badge} and \dat{}~\cite{kim2020date}.

\subsubsection{Gatekeeping}
\label{sec:model:gate}
In practice, some importers might commit fraud by analyzing and reverse engineering the model's prediction patterns. We can call them adaptive adversaries of the model. In this situation, randomness is known to improve the robustness and competitiveness of the online algorithm~\cite{buchbinder2007online}. With this motivation, we introduce randomness to our sampling strategy. Using the validation performance of the \dat{} model, we establish a gatekeeper. If \textsf{Rev@n\%} is higher than the predefined value of $\theta$, the \bate{} exploration algorithm is used. Otherwise, if the \dat{} models' outputs are highly unreliable, these inputs can be considered an attack, thereby facilitating the random selecting of items for inspection.
%Due to the \dat{} model's large dependency, the \bate{} exploration strategy is expected to work well only with good outputs from the \dat{} model. Therefore, 
\if 0
In customs selection scenarios, one can also assume the presence of fraudsters who act as the adaptive adversaries of our model. Research literature involving online learning algorithms~\cite{buchbinder2007online} has shown that randomness in the selection process improves the competitiveness of an algorithm under an online setting ski-rental problem, as a randomized algorithm is more robust against an adaptive online adversary model. Considering this, we introduce randomness to our sampling strategy.
\fi 
To address the issues above, we propose the final exploration strategy, \gate{}, which is formally written as Algorithm \ref{alg:exploration_strategy}:\newline
\begin{algorithm}[bth]
\small
\SetNoFillComment
\begin{flushleft}
\smallskip
 \textbf{Input:} Training set $X_t$, items received $\mathcal{B}_t$, inspection rate $r_t$\\
 \textbf{Output:} A batch of selected items $\mathcal{B}_t^S$ \\
 \SetAlgoLined
 \tcc{Corresponds to the selection part in Alg.~\ref{alg:active_custom_select}.}
 Train the \dat{} model using training set $X_t$;\\
 Obtain \textsf{Rev@n\%} from validation set;\\
 \eIf{\textsf{Rev@n\%} $> \theta$}{
     Perform prediction on $\mathcal{B}_t$, obtain the predicted annotation $(\textbf{x}_i, \hat{y}_i^{cls}, \hat{y}_i^{rev})$ for each item $\textbf{x}_i \in \mathcal{B}_t$;\\
     Calculate the gradient embedding $g_{x_{i}}$ (Eq. \ref{eqn:embedding});\\
     Obtain the set $\mathcal{B}^S_t$ of $r_t|\mathcal{B}_t|$ items by $k$-means++ initialization;\\
 }{
     Obtain the set $\mathcal{B}^S_t$ of $r_t|\mathcal{B}_t|$ items by random sampling;\\
 }
 \caption{Exploring unknown items by \gate{}}
 \label{alg:exploration_strategy}
\end{flushleft}
\end{algorithm}

\subsection{Hybrid Strategy}
The exploitation-only model can lead to confirmation bias. With a model trained only on historical data and considering the concept drift in customs datasets, the model tends to be unreliable because of outliers. However, a pure exploration strategy cannot secure customs revenue and is unrealistic in the customs setting. Hence, we consider a balance between the two to achieve both short-term and long-term performance. We propose a \emph{hybrid selection strategy} under the online active learning setting that includes two main approaches, \emph{exploitation} and \emph{exploration}, by utilizing \dat{} and \gate{}.

\begin{table*}[t!]
    \centering
    \caption{\sd{Overview of the item-level import data, in which the description and example of each variable are provided.}}\label{tab:data-overview}
    \resizebox{\linewidth}{!}{%
    \begin{tabular}{@{}clllr@{}}
        \toprule
      Type & Variable & Description & Example &  \\ \midrule
% Year  & The year in which the transaction occurred & 2013 \\
% \hline
\multirow{12}{*}[0ex]{Features} & \emph{sgd.id} & An individual numeric identifier for Single Goods Declaration (SGD). & SGD347276 \\ 

 & \emph{sgd.date}  & The year, month and day on which the transaction occurred. & 13-11-28 \\

 & \emph{importer.id}  & An individual identifier by importer based on the tax identifier number (TIN) system. & IMP364856  \\ 
 & \emph{declarant.id}  & An individual identification number issued by Customs to brokers. & DEC795367 \\ 
 & \emph{country} & Three-digit country ISO code corresponding to transaction. & USA \\ 
 & \emph{office.id} & The customs office where the transaction was processed. & OFFICE91 \\ \cmidrule{2-4}
 & \emph{tariff.code} & A 10-digit code indicating the applicable tariff of the item based on the harmonised system (HS). & 8703232926 \\ 
% Receipt number & An individual numeric identifier confirming payment of Customs duty and relevant taxes & RCP287715 \\
% \hline
% Receipt date  & The year, month and day on which payment of Customs duty and relevant taxes were made & 13-10-18 \\
% \hline
% Tariff description  & A description of the item based on the HS coding system\\
% \hline
 & \emph{quantity}  & The specified number of items. & 1 \\ 
 & \emph{gross.weight} & The physical weight of the goods. & 150kg \\ \cmidrule{2-4}
 & \emph{fob.value} & The value of the transaction excluding, insurance and freight costs. & \$350 \\ 
 & \emph{cif.value} & The value of the transaction including the insurance and freight costs. & \$400   \\ 
 & \emph{total.taxes} & Tariffs calculated by initial declaration. & \$50 \\ \midrule
% Inspection information  & A short description of the item based on Customs observation & (1) REFER TO D/C VALUATION AFTER EXAMINATION. \\
% \hline
% Selectivity lane & The relevant Risk Management channel (Green, Blue, Yellow, and Red) used to exit the item from Customs & RED \\
% \hline
\multirow{2}{*}[0ex]{Prediction Target} &  \textbf{\emph{illicit}} & Binary target variable that indicates whether the object has fraud. & 1  \\ 
 & \textbf{\emph{revenue}} & Amount of tariff raised after the inspection, only available on some illicit cases. & \$20  \\  \bottomrule
\end{tabular}}
\end{table*}
To select items that will potentially enhance the model's performance, we design a \gate{} strategy for exploration (\cref{sec:model:embedding}--\ref{sec:model:gate}). We use the \dat{} strategy that exploits historical knowledge to generate the highest possible revenue~\cite{kim2020date} and guarantees short-term revenue. The final selection is made by the hybrid approach, which balances the \gate{} and \dat{} strategies. The hybrid selection logic can be formally represented by Algorithm~\ref{alg:hybrid_strategy}.\looseness=-1
\begin{algorithm}[tbh]
\small
\SetNoFillComment
\begin{flushleft}
\smallskip
 \textbf{Input:} Training set $X_t$, items received $\mathcal{B}_t$, inspection rate: $r_t$, predefined ratio between two strategies $p_1, p_2$ {\small$(p_1 + p_2 = 1)$}\\
 \textbf{Output:} A batch of selected items $\mathcal{B}_t^S = \mathcal{B}_{t}^F$ $\cup$ $\mathcal{B}_{t}^U$ \\
 \SetAlgoLined
 \tcc{Corresponds to the selection part in Alg.~\ref{alg:active_custom_select}.}
 Train the \dat{} model using training set $X_{t}$;\\
 Obtain the set $\mathcal{B}^F_t$ of $p_1r_{t}|\mathcal{B}_t|$ items by the \dat{} strategy;\\
%  Exclude the chosen items from the received data $\mathcal{B}_t$ = $\mathcal{B}_t$ \setminus $\mathcal{B}^F_t$; \\
$\mathcal{B}_{t}$ = $\mathcal{B}_{t} \setminus \mathcal{B}_{t}^F$; \\
Obtain the set $\mathcal{B}^U_t$ of $p_2r_{t}|\mathcal{B}_t|$ items by the \gate{} strategy; \\
 \caption{Hybrid Selection using \dat{} and \gate{}}
 \label{alg:hybrid_strategy}
\end{flushleft}
\end{algorithm}

\section{Experiments}
\label{sec:experiments}

\subsection{Evaluation Settings}
\label{sec:experiments:settings}
\subsubsection{Datasets}
\label{sec:experiments:settings:datasets}
We employed item-level import declarations for three countries in Africa. The import data fields included numeric variables such as the item price, weight, and quantity and categorical variables such as the commodity code (HS code), importer ID, country code, and received office. \sd{After matching the data format for each country, we preprocessed the variables by following the approach used in a previous study~\cite{kim2020date}. For categorical variables, we quantified the risk indicators of the importers, declarants, HS code, and countries of origin from their non-compliance records. For example, importers were ranked by their past fraud rates. The importers, whose ranks were above the 90th percentile, were regarded as high-risk importers, and their risk indicators were given values of 1; otherwise, the values are set to 0. This is called risk profiling, and it is more efficient than one-hot encoding those variables. We also add frequently-used cross features such as \emph{unit.value}~(=\nicefrac{cif.value}{quantity}), \emph{value/kg} (=\nicefrac{cif.value}{gross.weight}), \emph{tax.ratio} (=\nicefrac{total.taxes}{cif.value}), \emph{unit.tax} (=\nicefrac{total.taxes}{quantity}), and \emph{face.ratio} (=\nicefrac{fob.value}{cif.value}). Table~\ref{tab:data-overview} illustrates the import declaration data.} 
\begin{table}[t!]
\centering
% \scriptsize
\caption{Statistics of the datasets}
\label{tab:datastats}
\resizebox{0.9\linewidth}{!}{%
\begin{tabular}{ l | r | r | r } \toprule
    Datasets &  Country \textsf{M} & Country \textsf{N} & Country \textsf{T}  \\ \midrule
    Periods & Jan 13--Dec 16 & Jan 13--Dec 17 & Jan 15--Dec 19 \\ 
    \# imports & 0.42M & 1.93M & 4.17M  \\ 
    \# importers & 41K & 165K & 133K  \\ 
    \# tariff codes & 1.9K & 6.0K & 13.4K  \\ 
    GDP per capita & \$412 & \$2,230 & \$3,317 \\
    Avg. illicit rate &  1.64\% & 4.12\% & 8.16\%  \\ 
    \bottomrule
\end{tabular}
}
\end{table}
\begin{figure}[t!]
\centering
    \begin{subfigure}[b]{0.43\columnwidth}
        \centering
        \includegraphics[width=\linewidth]{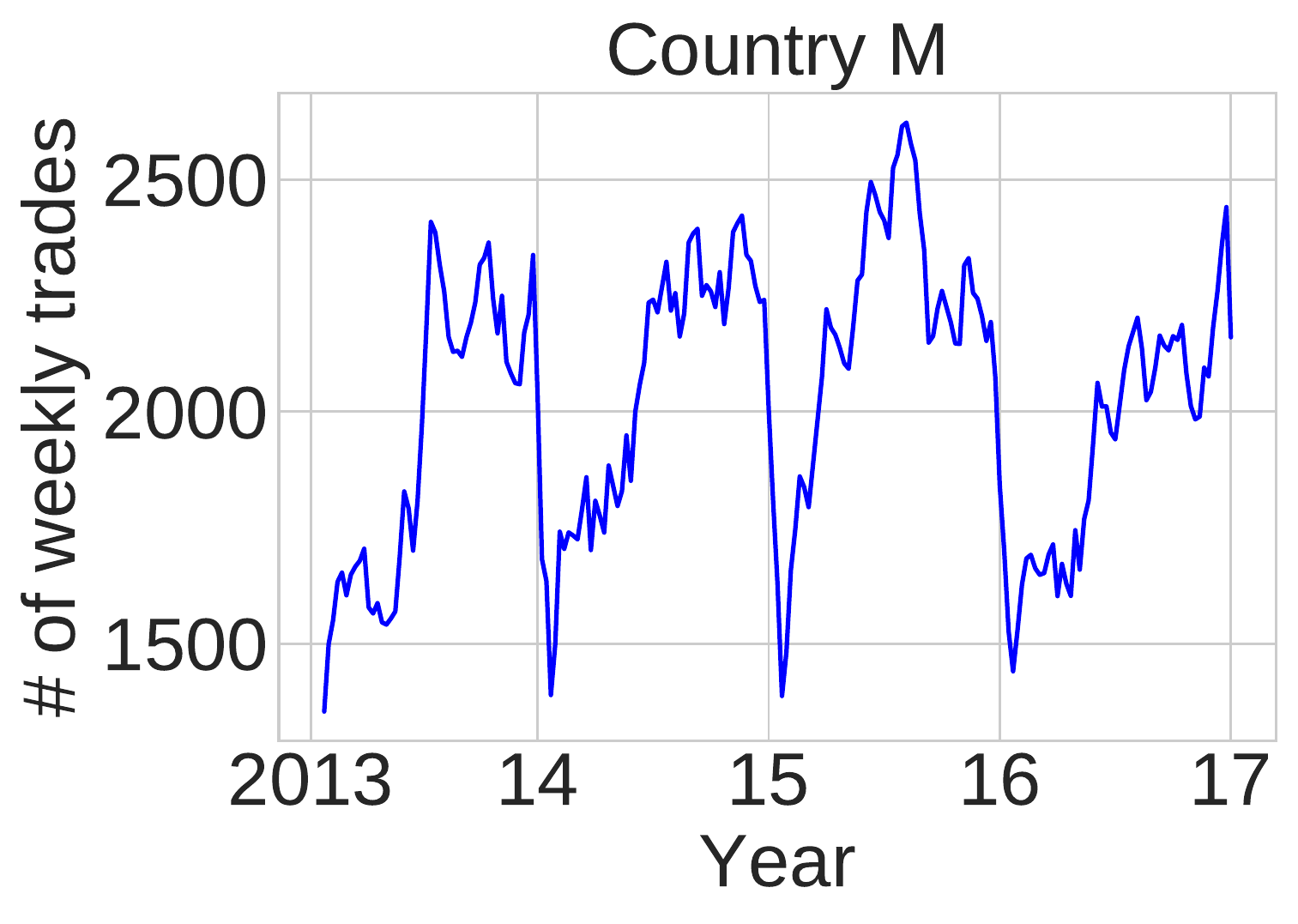}
    \end{subfigure}
    \begin{subfigure}[b]{0.43\columnwidth}
        \centering
        \includegraphics[width=\linewidth]{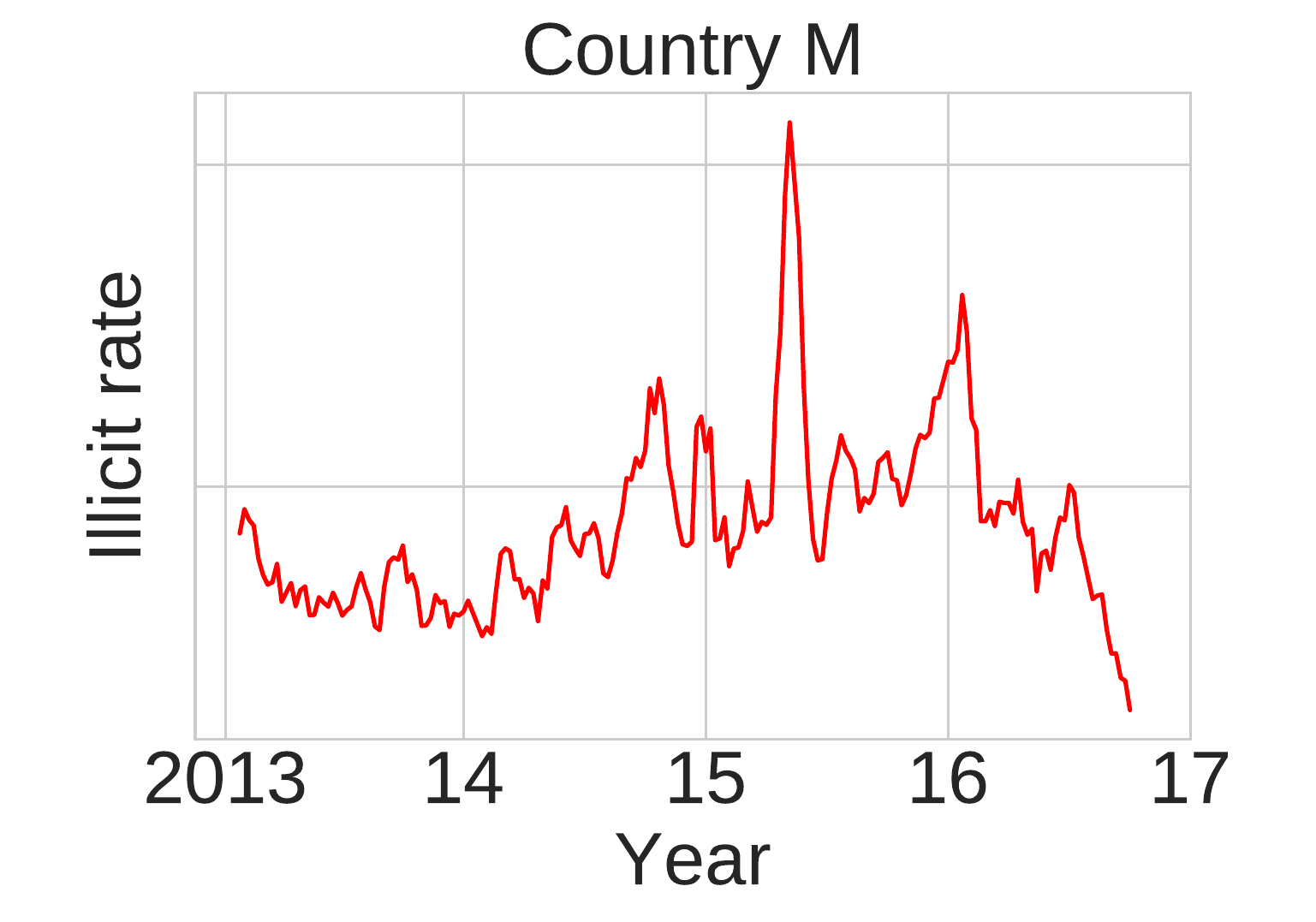}
    \end{subfigure}
    \begin{subfigure}[b]{0.43\columnwidth}
        \centering
        \includegraphics[width=\linewidth]{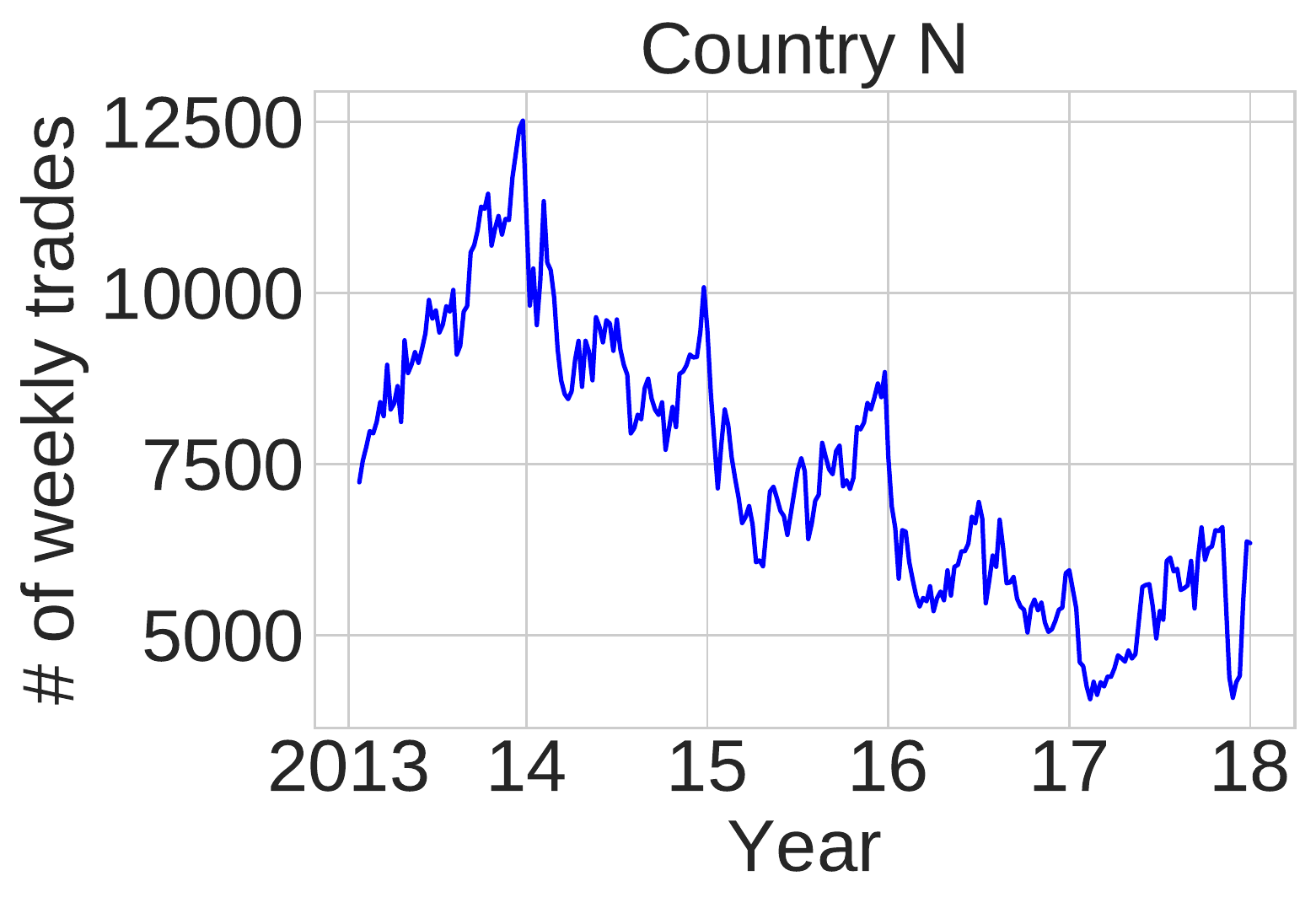}
    \end{subfigure}
    \begin{subfigure}[b]{0.43\columnwidth}
        \centering
        \includegraphics[width=\linewidth]{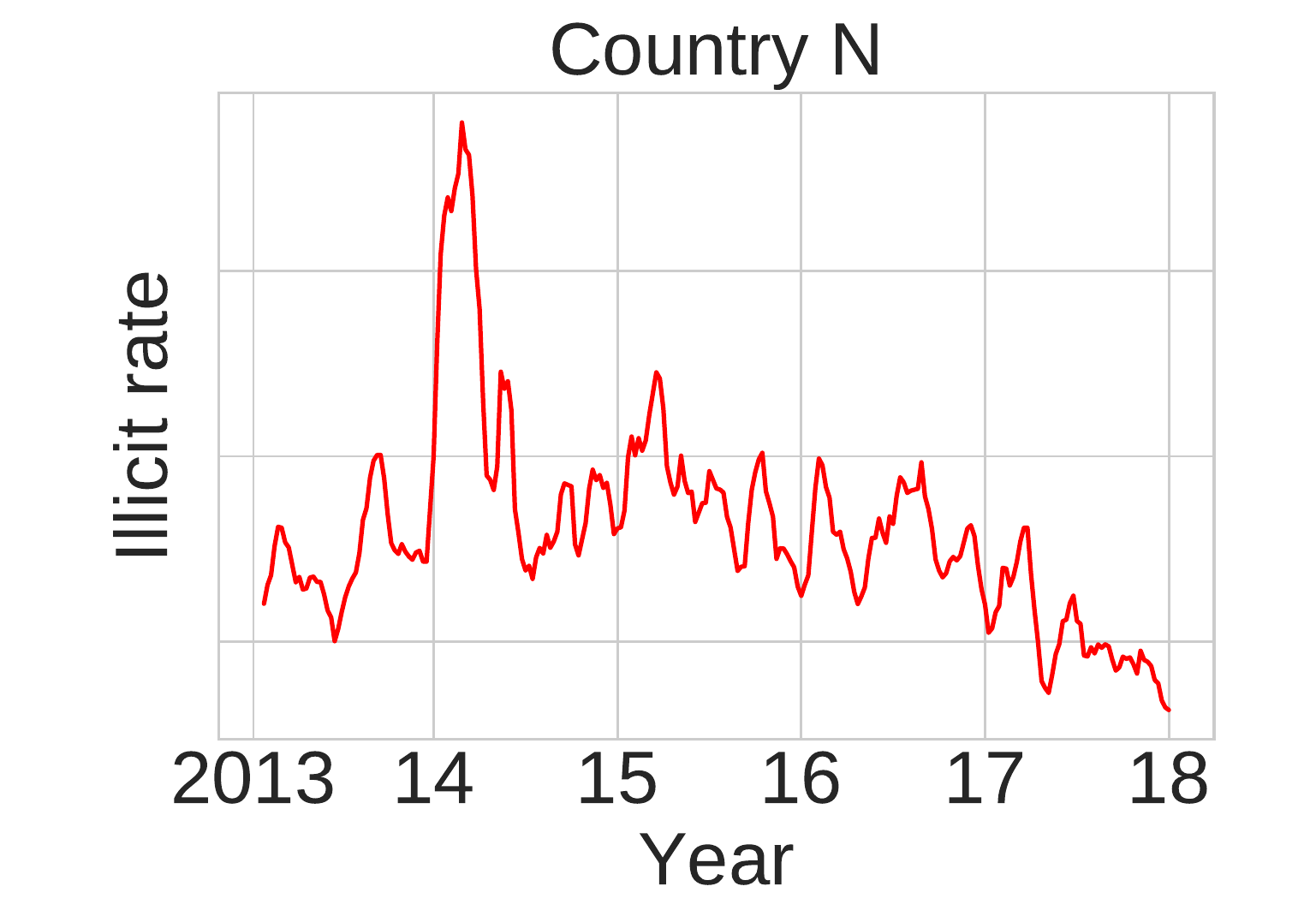}
    \end{subfigure}
    \begin{subfigure}[b]{0.43\columnwidth}
        \centering
        \includegraphics[width=\linewidth]{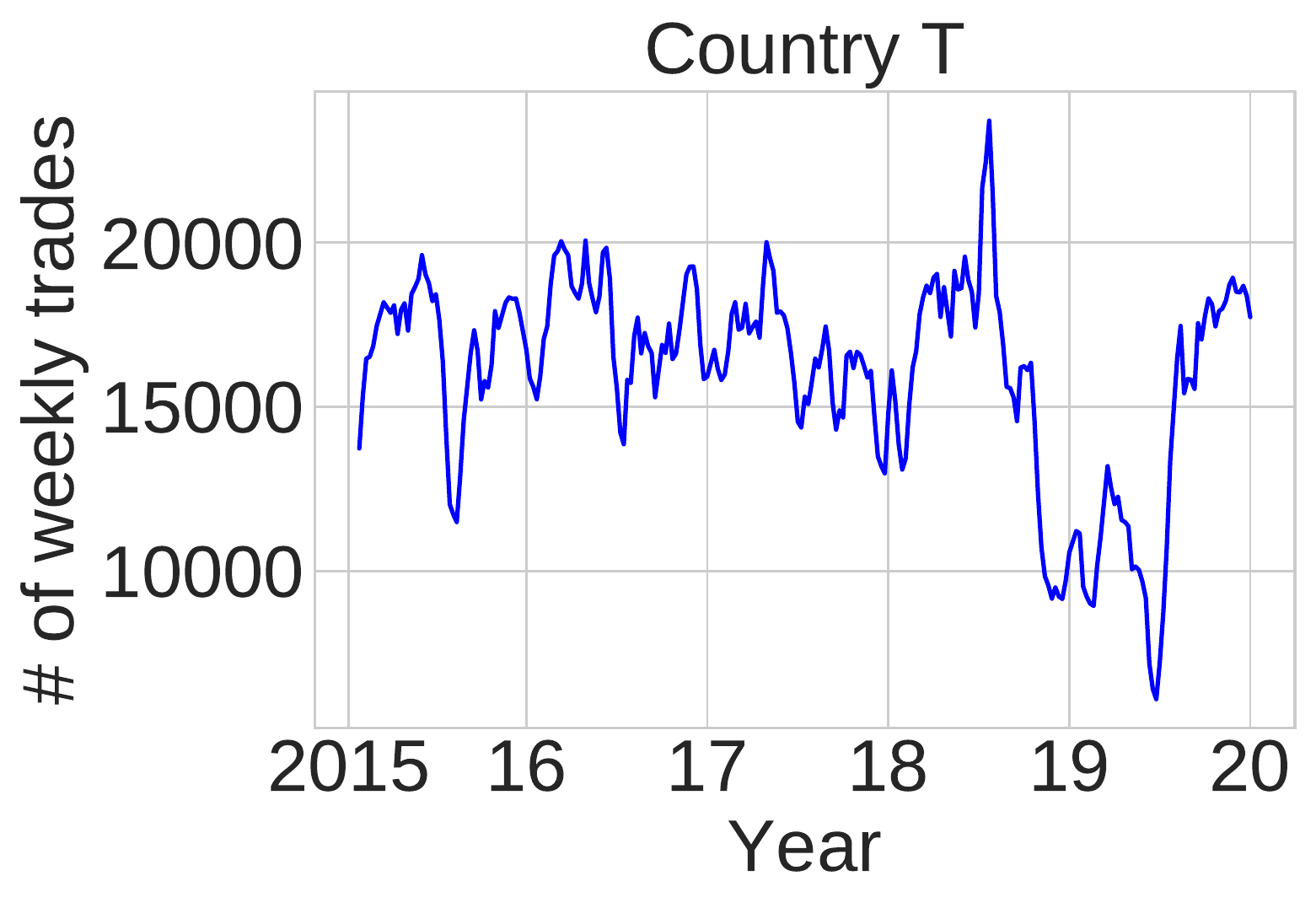}
    \end{subfigure}
    \begin{subfigure}[b]{0.43\columnwidth}
        \centering
        \includegraphics[width=\linewidth]{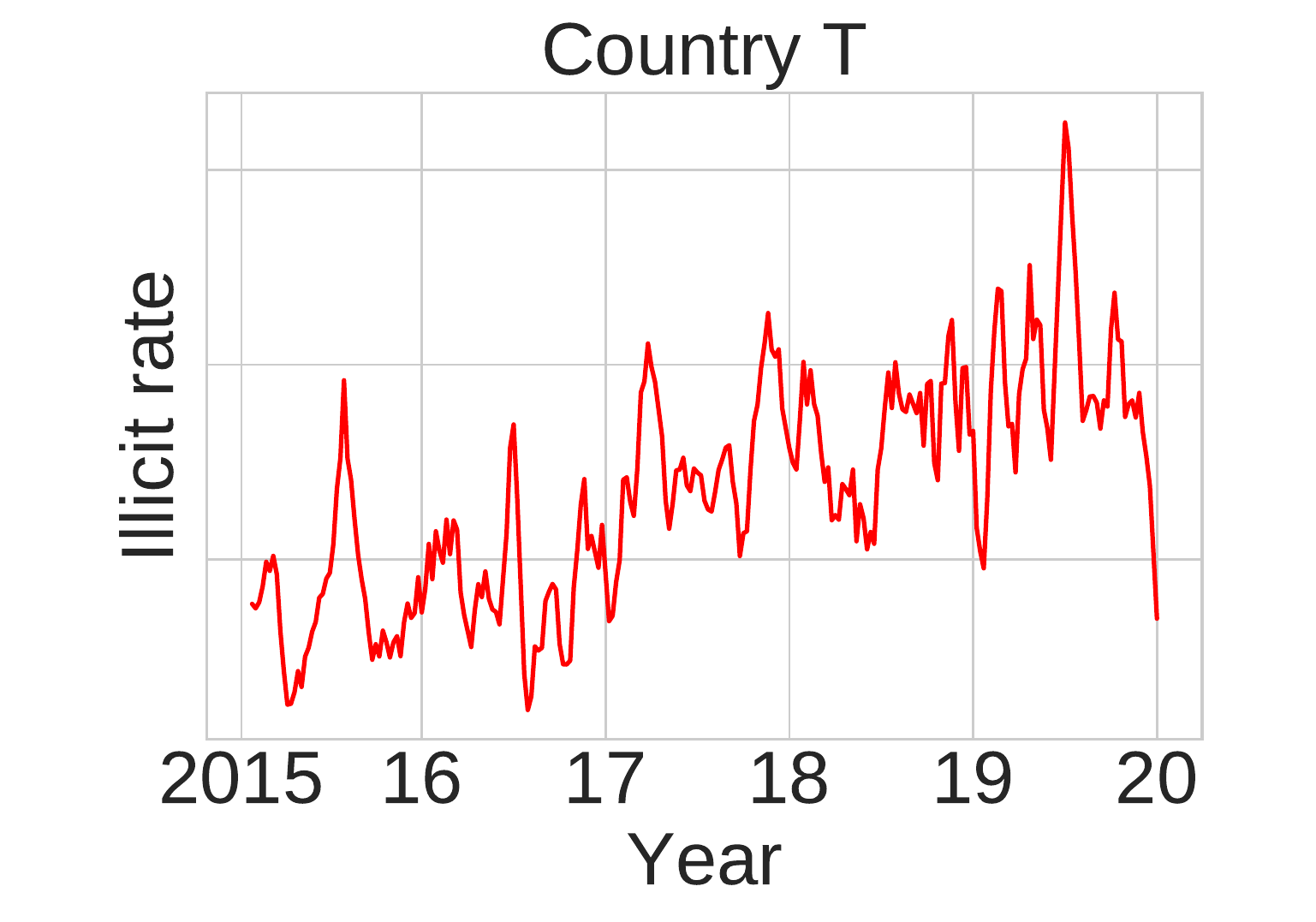}
    \end{subfigure}
    \caption{Number of weekly trades and illicit rate trend over time. The actual values of the illicit trends are hidden due to nondisclosure agreements.}
    \label{fig:weeklystats}
\end{figure}

The three customs were subjected to detailed inspection (i.e., achieving a nearly 100\% inspection rate). However, this practice is not sustainable, and the customs offices of these countries plan to reduce the inspection rate in the future. Due to the manual inspection policy, the item labels and tariffs charged are accurately labeled in these logs at the single-goods level. Table~\ref{tab:datastats} and Figure~\ref{fig:weeklystats} depict the statistics of the data we utilized. %  the existence of concept drift (KL-divergence?)
\begin{figure*}[bth!]
\centering
    \begin{subfigure}[b]{.32\linewidth}
        \centering\captionsetup{width=.95\linewidth}%
        \includegraphics[width=\linewidth]{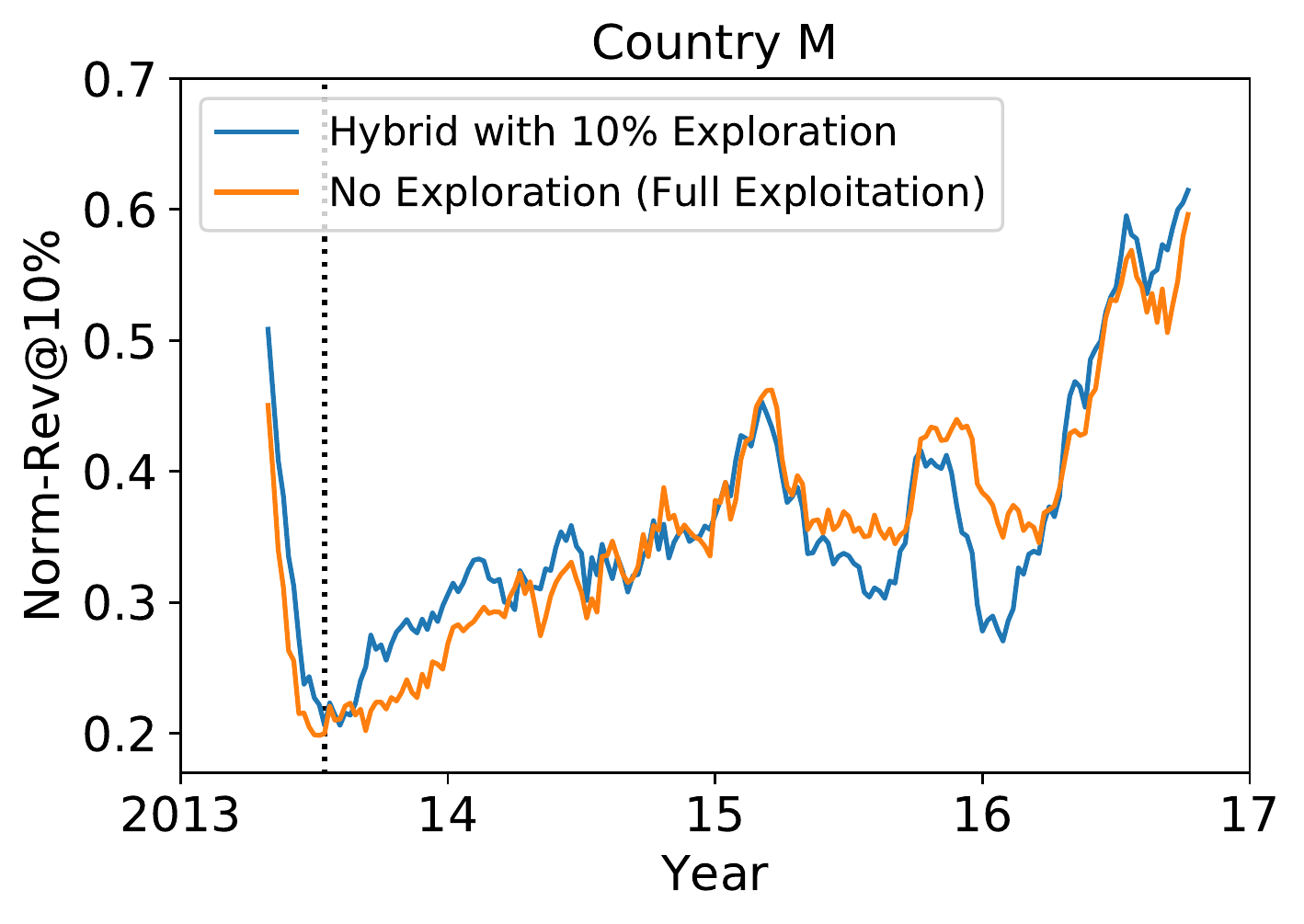}
        \caption{In country \textsf{M}, the performances of both strategies increase over time.}
    \end{subfigure}
    \begin{subfigure}[b]{.32\linewidth}
        \centering\captionsetup{width=.95\linewidth}%
        \includegraphics[width=\linewidth]{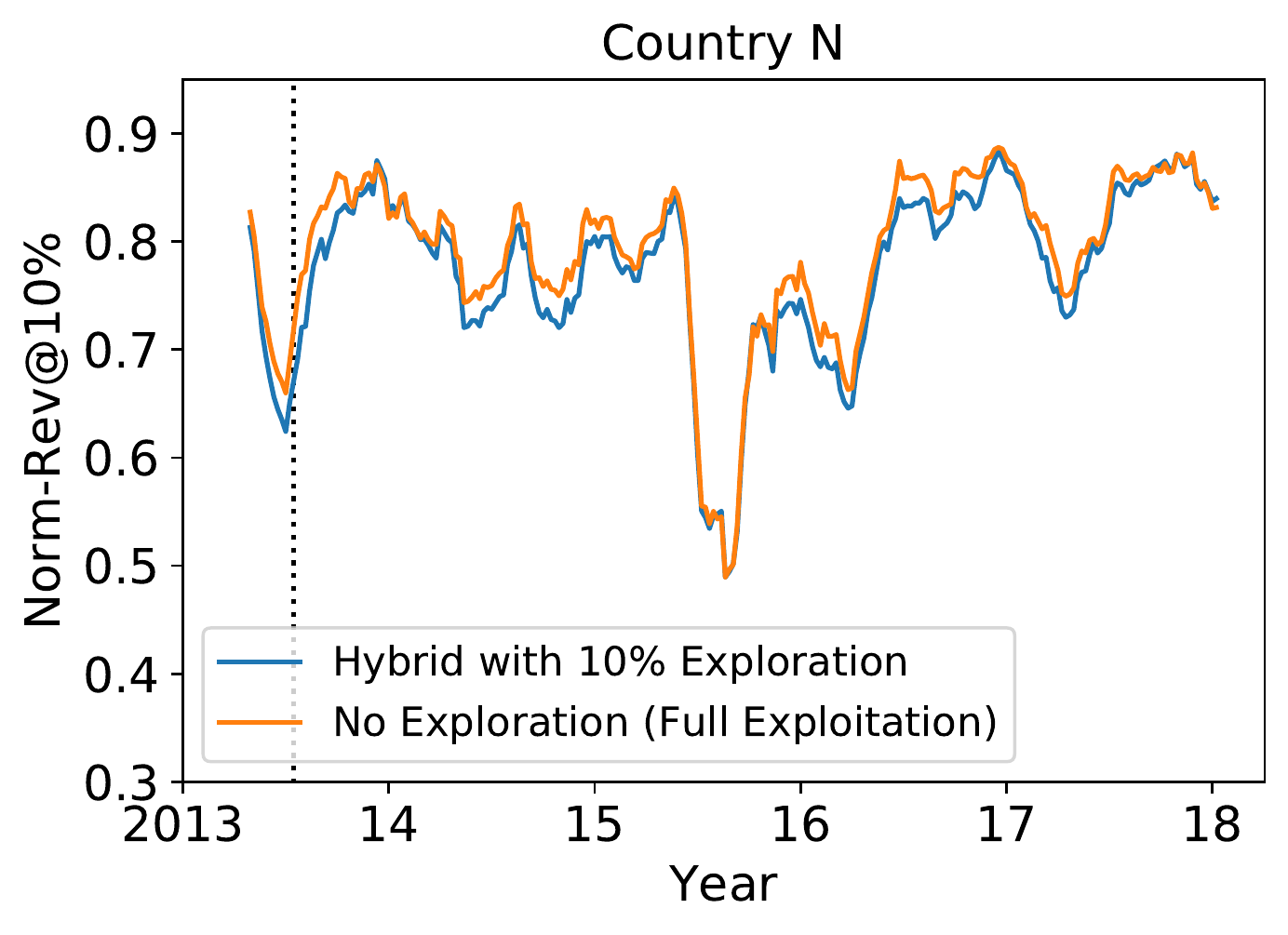}
        \caption{In country \textsf{N}, the performances of both strategies are high and stable.}
    \end{subfigure}
    \begin{subfigure}[b]{.32\linewidth}
        \centering\captionsetup{width=.95\linewidth}%
        \includegraphics[width=\linewidth]{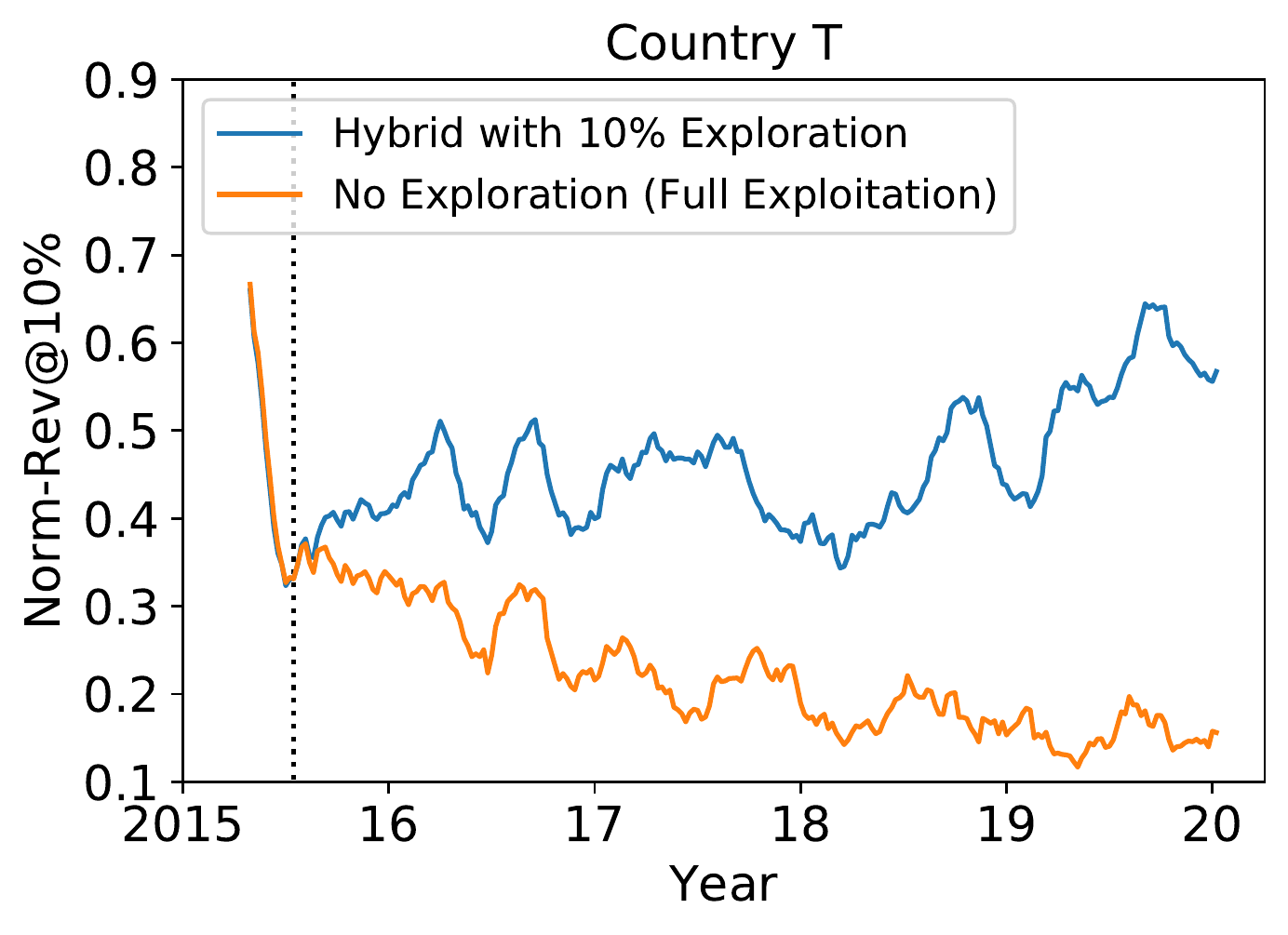}
        \caption{In country \textsf{T}, the exploitation strategy failed, unlike a hybrid model.}
    \end{subfigure}
    \caption{For some cases, the performance of the exploitation strategy \dat{} drops over time, but the performance of hybrid strategies remains stable even for cases in which the exploitation strategy fails. This shows that exploration is necessary for maintaining a selection system in the long run.}
    \label{fig:exploitation-fails}
\end{figure*}
\subsubsection{Long-term simulation setting}
\label{sec:experiments:settings:scenario}
The experiment aims to find the best selection strategy to maintain the customs trade selection model in the \emph{long run}. Therefore, we \sd{simulated} an environment in which a selection model is deployed and maintained for multiple years\footnote{Previous works split the data into training and testing sets on a temporal basis and compared the performance of diverse machine learning models~\cite{kim2020date, vanhoeyveld2020belgian}. However, the algorithm's performance in a static prediction state cannot indicate the model's performance in a real setting when the model is deployed.}. Given that one month of training data is available, the system receives import declarations and selects a batch of items to inspect during the week. A selection model is trained based on a predefined strategy, and the most recent four weeks of data are used to validate the model. By using the inspection results, the model is updated every week.

To simulate a scenario of data providers who are willing to reduce the inspection rate gradually, we implemented several methods to decay the inspection rate over time. In this experiment, we set the target inspection rate to 10\%. Starting with 100\% inspection, we used a linear decaying policy by reducing the inspection rate by 10\% each week. Once the target inspection rate is reached, the system maintains this inspection rate for the remaining period. In Fig.~\ref{fig:exploitation-fails},~\ref{fig:pure-exploration},~\ref{fig:exploration-for-hybrid}, and \ref{fig:synthetic-results}, we use a vertical dashed line to indicate when the decay ends, and the target rate is maintained.\footnote{In countries where the daily import declaration is larger, it would be possible to update the selection strategy every day, and more reliable results could be obtained even with a shorter period.}

\subsubsection{Evaluation metrics}
\label{sec:experiments:settings:metrics}
We evaluate the selection strategy performance by referring to two metrics used in previous work~\cite{kim2020date}: \textsf{Precision@n\%} and \textsf{Revenue@n\%}. Since the underlying data distribution changes each week in an online setting, these value metrics also fluctuate significantly. In other words, unless the illicit rates or item prices are fixed, these two value metrics will be difficult to interpret directly. We used normalized performances by dividing each value metric by the maximum achievable value, namely, the oracle value.
\begin{itemize}[leftmargin=*]
    \item \textbf{\textsf{Norm-Precision@n\%}}: In a situation where n\% of all declared goods are inspected, \textsf{Pre@n\%} indicates how many actual instances of fraud exist among the inspected items. The \textsf{Norm-Pre@n\%} value of the corresponding algorithm is defined as the value obtained by dividing the \textsf{Pre@n\%} of the algorithm by the \textsf{Pre@n\%} of the oracle. 
    \item \textbf{\textsf{Norm-Revenue@n\%}}: \textsf{Rev@n\%} indicates how much revenue (extra tax) can be secured by examining the set of items. The oracle will select the items with the highest revenue. The \textsf{Norm-Rev@n\%} value of the corresponding algorithm is defined as the value obtained by dividing the \textsf{Rev@n\%} of the algorithm by the \textsf{Rev@n\%} of the oracle.
    % \textsf{Rev@n\%} is the total revenue in top n\% transactions identified by a model divided by the total revenue among all transactions. This metric explains how much more customs duties can be generated from the top n\% of transactions than the revenue generated by inspecting the entire transactions. 
\end{itemize}
\begin{Ex}
For example, if a system with a 10\% inspection rate is operating in an environment with a 2\% illicit rate, the \textsf{Pre@10\%} and \textsf{Rev@10\%} of the oracle would be 0.2 and 1, respectively. Let us consider that the deployed selection strategy achieves a \textsf{Pre@10\%} value of 0.18. To prevent any potential interpretation bias caused by the illicit rate that varies from country to country, we divide 0.18 by the performance upper bound of 0.2, which results in 0.9 for \sd{\textsf{Norm-Pre@10\%}}. 
\end{Ex}
\begin{note}
We employ a fully labeled dataset and these metrics as the ground truth information. For countries already maintaining a low inspection rate, these metrics can be modified by conditioning on their observable goods. 
\end{note}
Securing tax revenue was the most critical screening factor for the developing countries we interacted with since their fiscal income depends on customs services~\cite{wco-annual}. Therefore, we mainly used \textsf{Norm-Rev@n\%} in reporting the results in the following sections.
\subsection{When the Exploitation Strategy Fails}
\label{sec:experiments:exploitationFails}
To explore this possibility, we first compare the performance of the pure exploitation strategy \dat{} and a partial-exploitation strategy with some \emph{random} exploration, called a naive hybrid strategy. The naive hybrid strategy uses \dat{} and random exploration at a ratio of $nine$ to $one$. The data reveal that a pure exploitation strategy can lead to a substantial degree of malfunctioning.

\begin{figure}[!t]
  \centering
  \includegraphics[width=.85\linewidth]{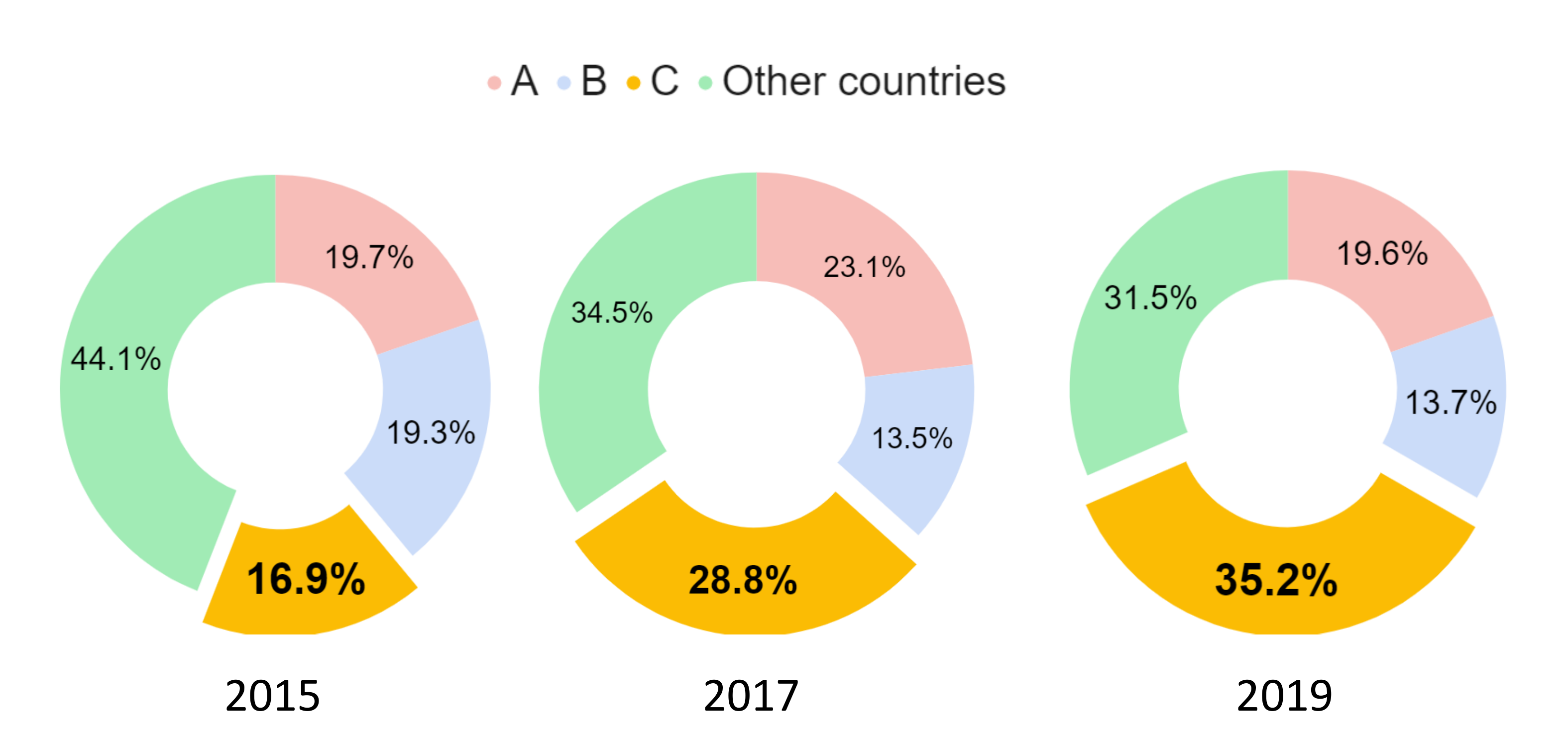}
%   \caption{Over time, novel frauds arise. A fraud detection system that selects based on historical data will not find such frauds.}
  \caption{An example of concept drift in country \textsf{T}: The source country for commodity \textsf{X} (HS-code starting with 620) rapidly changes over the years.}
  \label{fig:domain_shift}
\end{figure}

Figure~\ref{fig:exploitation-fails}(c) shows that for country \textsf{T}, the performance of the \emph{state-of-the-art} \dat{} exploitation strategy drops unexpectedly from time to time, yet the hybrid strategy remains stable. The degradation continues despite the increasing size of the training data, confirming that items chosen for inspection are uninformative and indicating a concept drift in the country's trade pattern. Hence, we conclude that the exploration strategy items significantly boost the performance of the exploitation strategy. Considering that the randomly selected items may affect only 1\% of the total revenue on average, the performance boost arises from inspecting unknown items. \looseness=-1 
 
The longitudinal data also allow us to examine how frequently concept drifts occurred in the trading pattern of country \textsf{T}. Figure~\ref{fig:domain_shift} shows the ratio of each import country for an item with a commodity code starting with 620 in 2015, 2017, and 2019, indicating a significant level of concept drifts in trade rates for the top imported items. Countries \textsf{A} and \textsf{B} used to be where the item was imported the most, but starting in 2017, the shift in import countries sharply changed, and the country \textsf{C} became the dominant source country for imported goods.

\subsection{When the Exploitation Strategy Does Not Fail}
\label{sec:experiments:exploitationNotFails}
Is it common for the performance of the exploitation strategy to decrease over time? We check again to see if these behaviors are common across all countries. Reassuringly, we also observe that the full-exploitation strategy does not always fail. In Figure~\ref{fig:exploitation-fails}(a)--(b), we can see the results obtained from country \textsf{M} and country \textsf{N}. For these countries, maintaining the strategy of screening the most fraudulent items is still valid. However, when we compare the average performance of the exploitation strategy and the hybrid strategy, we can also find that the former does not outperform the latter (\textsf{Norm-Rev@10\%}, Exploitation vs. Hybrid: 59.6\% vs. 61.5\% for country \textsf{M}, 83.2\% vs. 84.0\% for country \textsf{N}; moving average over the previous 13 weeks). It is interesting to see that inspecting a set of random items is even better than inspecting reasonably fraudulent items with high $\hat{y}^{cls}$ values (top 9-10\%) for maintaining a customs trade selection system in the long run. Therefore, how much will the performance improve if the better exploration strategy is used rather than the random strategy? We measured the performance of the proposed exploration strategies. 

% Then, how about their performance on novel frauds?    
% \brian{Can we add more descriptions or evidence of why exploitation strategy is just enough? (e.g., concept drift analysis as in figure 6)} -> ㅠㅠ 

% dropping performance of the two exploitation strategies: select items in order of the high $\hat{y}^{cls}$ value of the XGBoost classifier~\cite{chen2016xgboost} and the \emph{state-of-the-art} \dat{} model~\cite{kim2020date}.   

% To prove that exploration is necessary for,
% we think this result was due to the presence of concept drift.
% We perform our \update{} model on three described dataset. The result is shown in ...
% // Show DATE performance graph for three countries.
% We observed that for the dataset of country \textsf{T}, 
% The performance of the other two countries does not experience an overall decline, evidencing a more stable trade pattern.
\begin{figure*}[t!]
\centering
    \begin{subfigure}[b]{\linewidth}
        \centering
        \includegraphics[width=.30\linewidth]{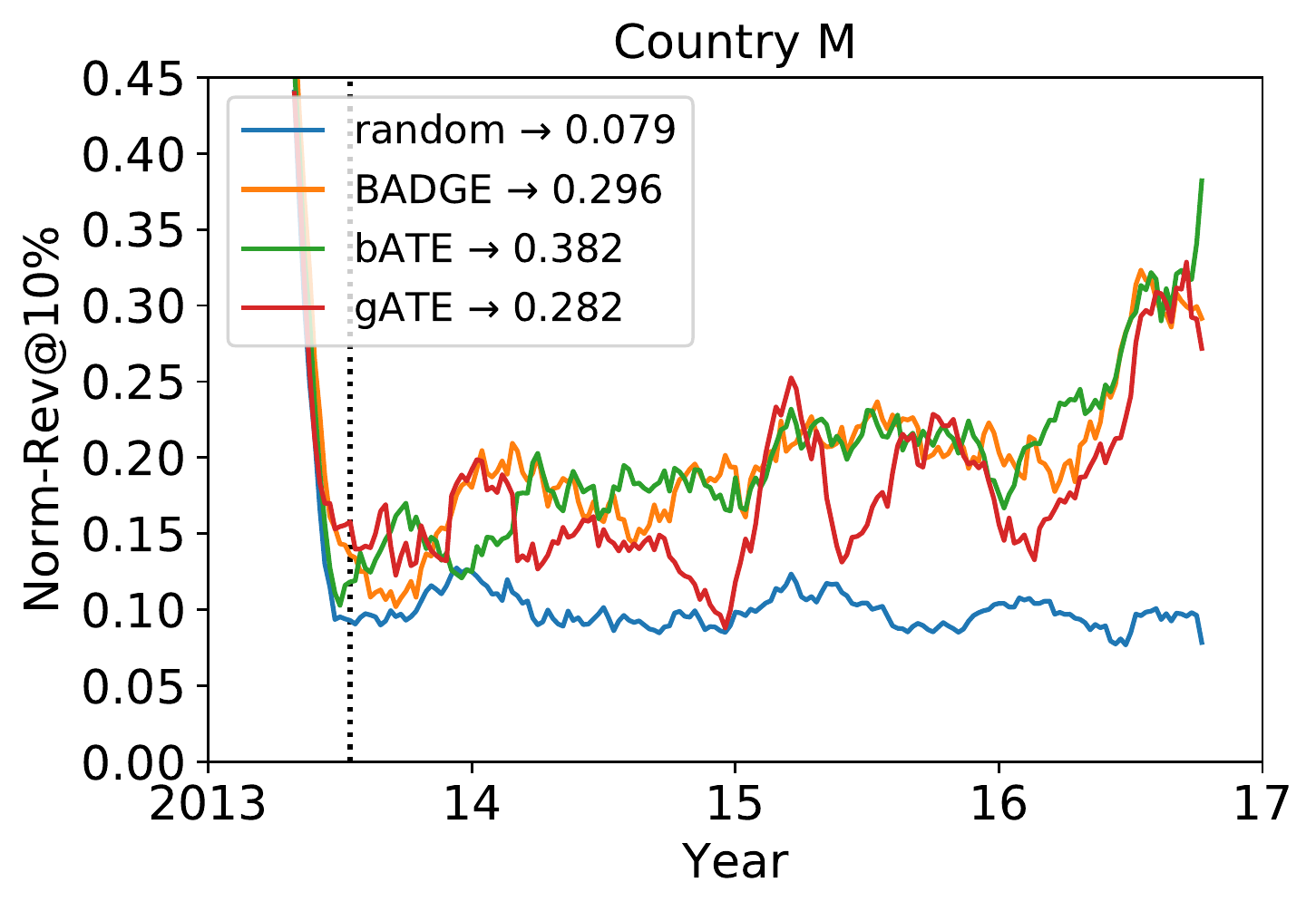}
        \includegraphics[width=.30\linewidth]{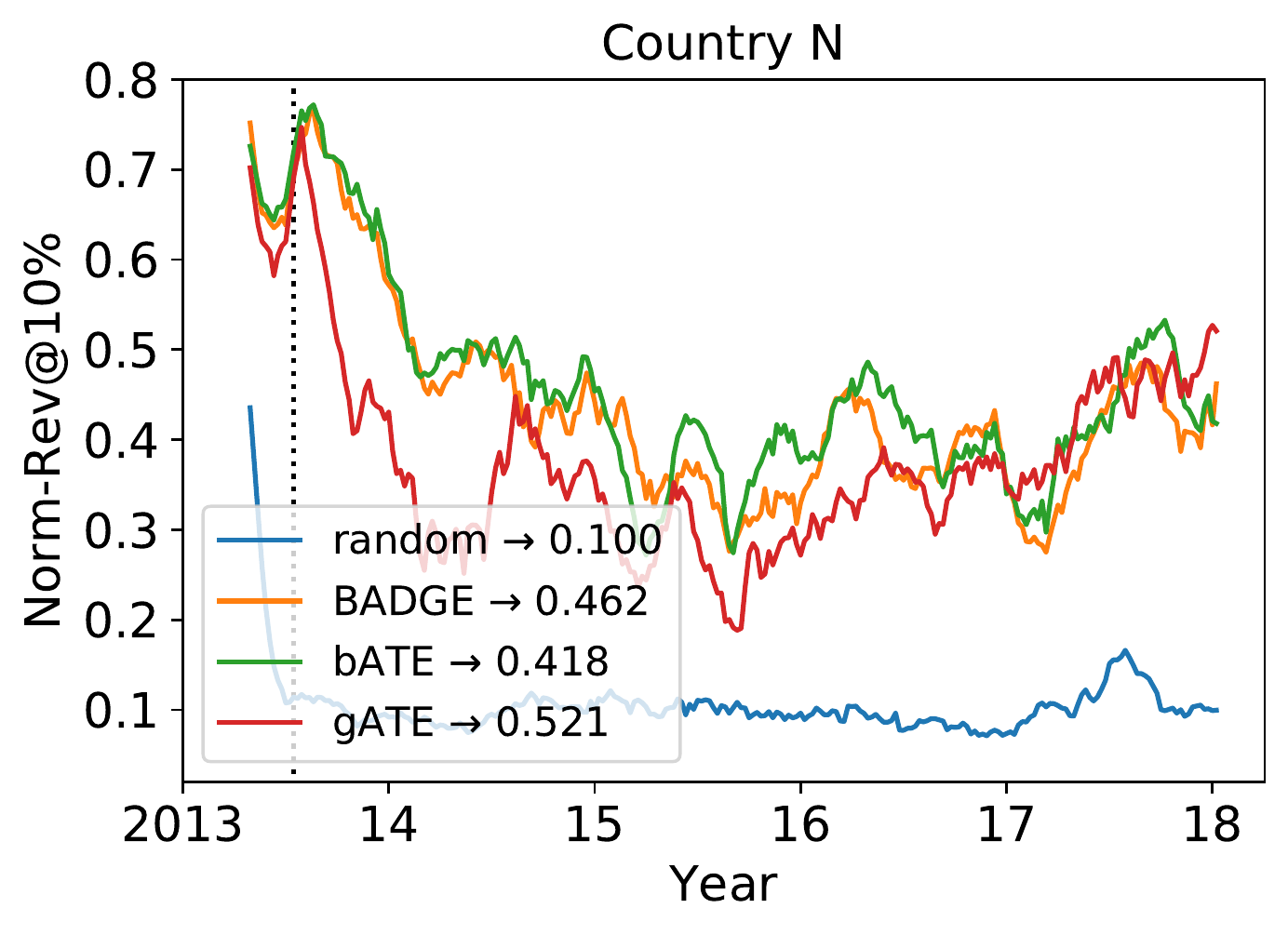}
        \includegraphics[width=.30\linewidth]{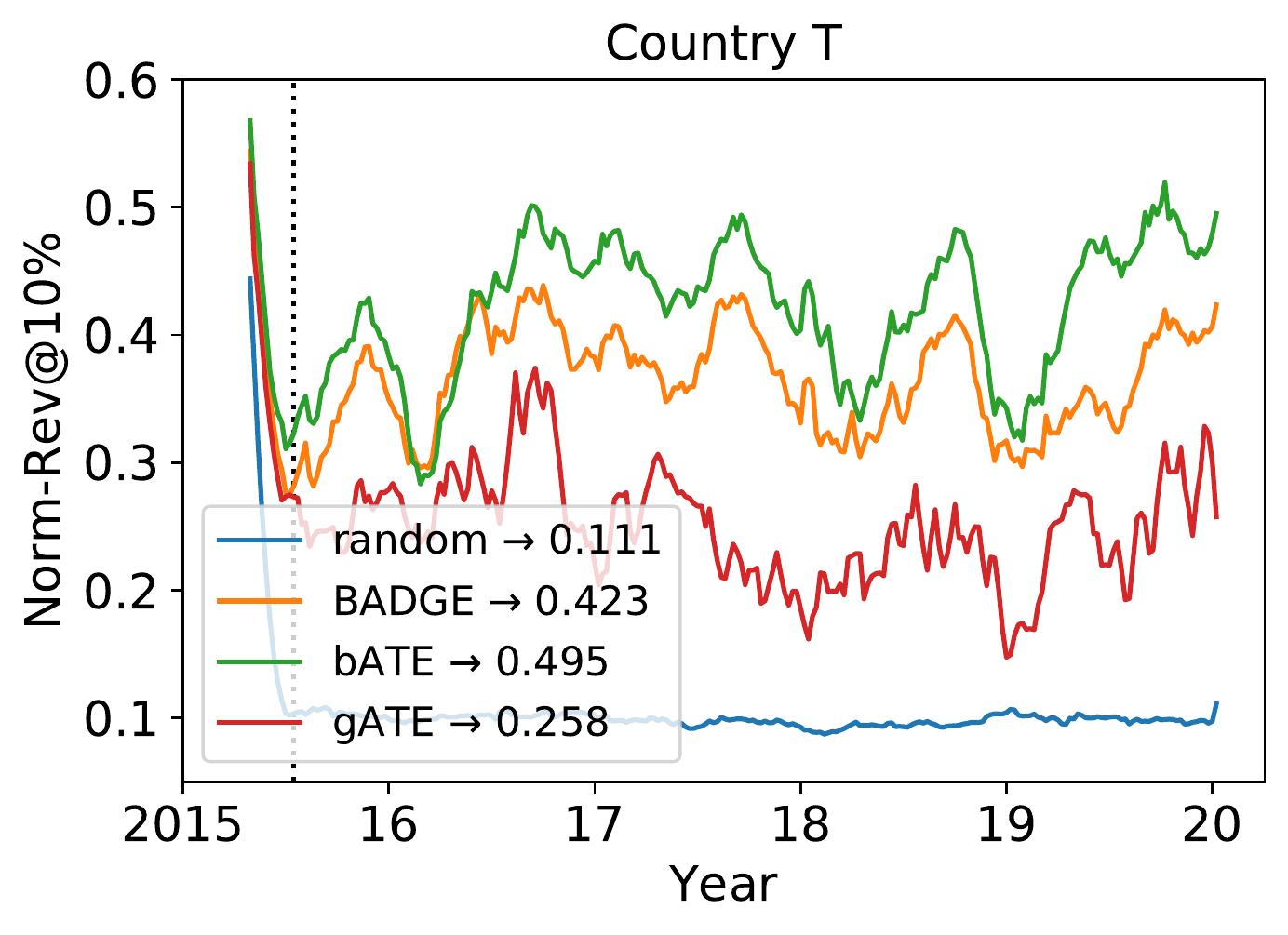}
        \caption{\textsf{Norm-Rev@10\%} performance of four exploration strategies on three country datasets.}
    \end{subfigure}
    \begin{subfigure}[b]{\linewidth}
        \centering
        \includegraphics[width=.30\linewidth]{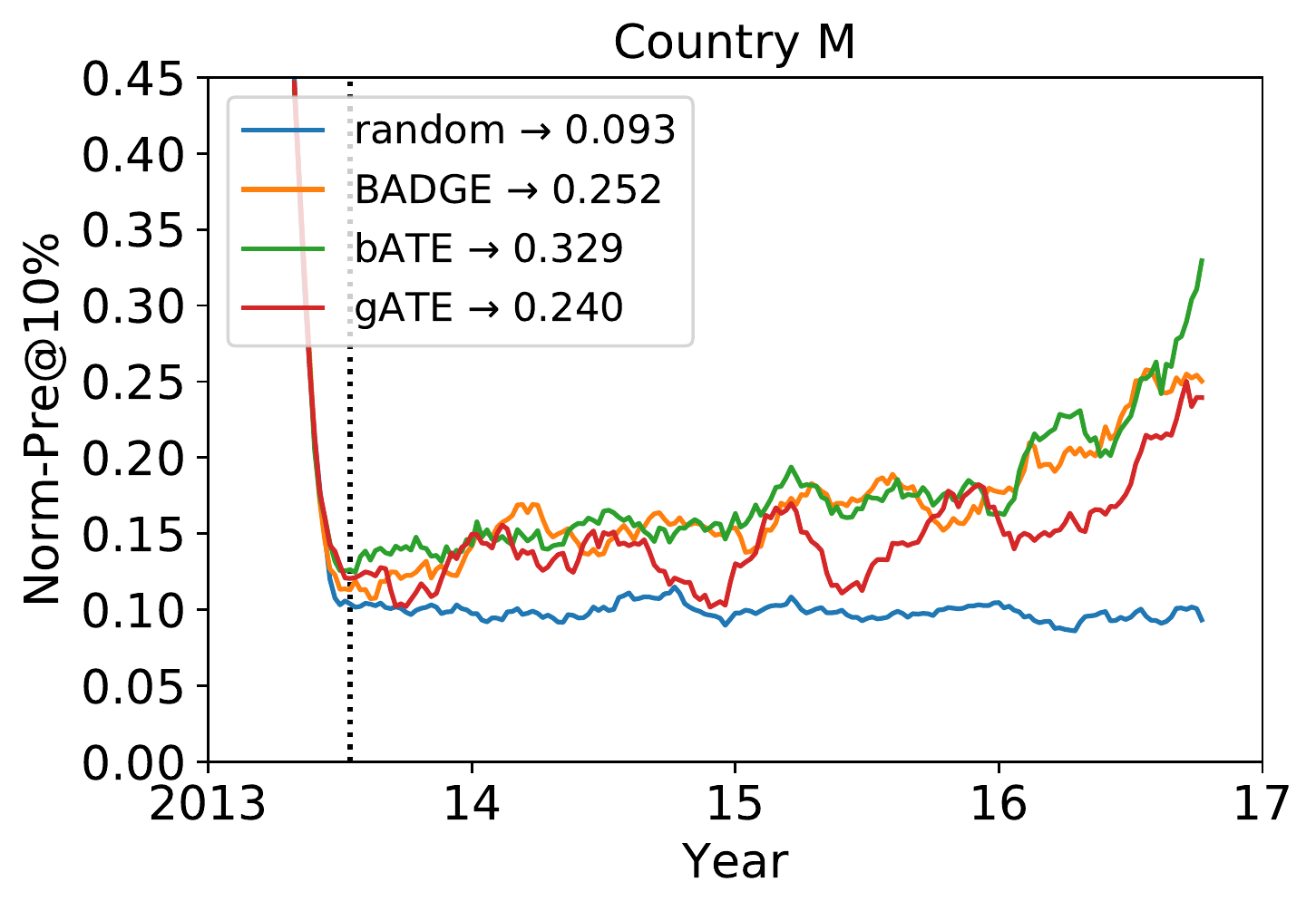}
        \includegraphics[width=.30\linewidth]{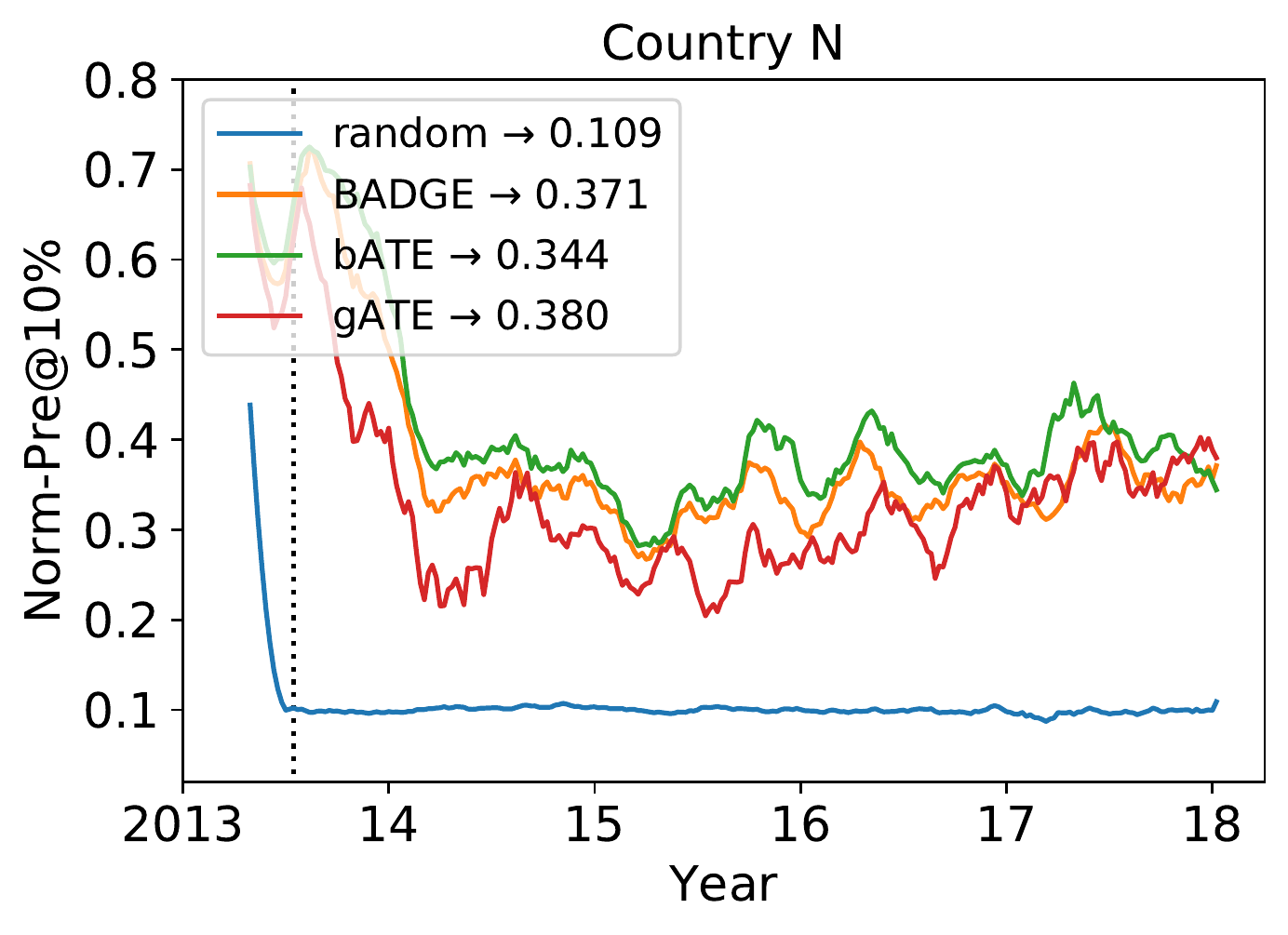}
        \includegraphics[width=.30\linewidth]{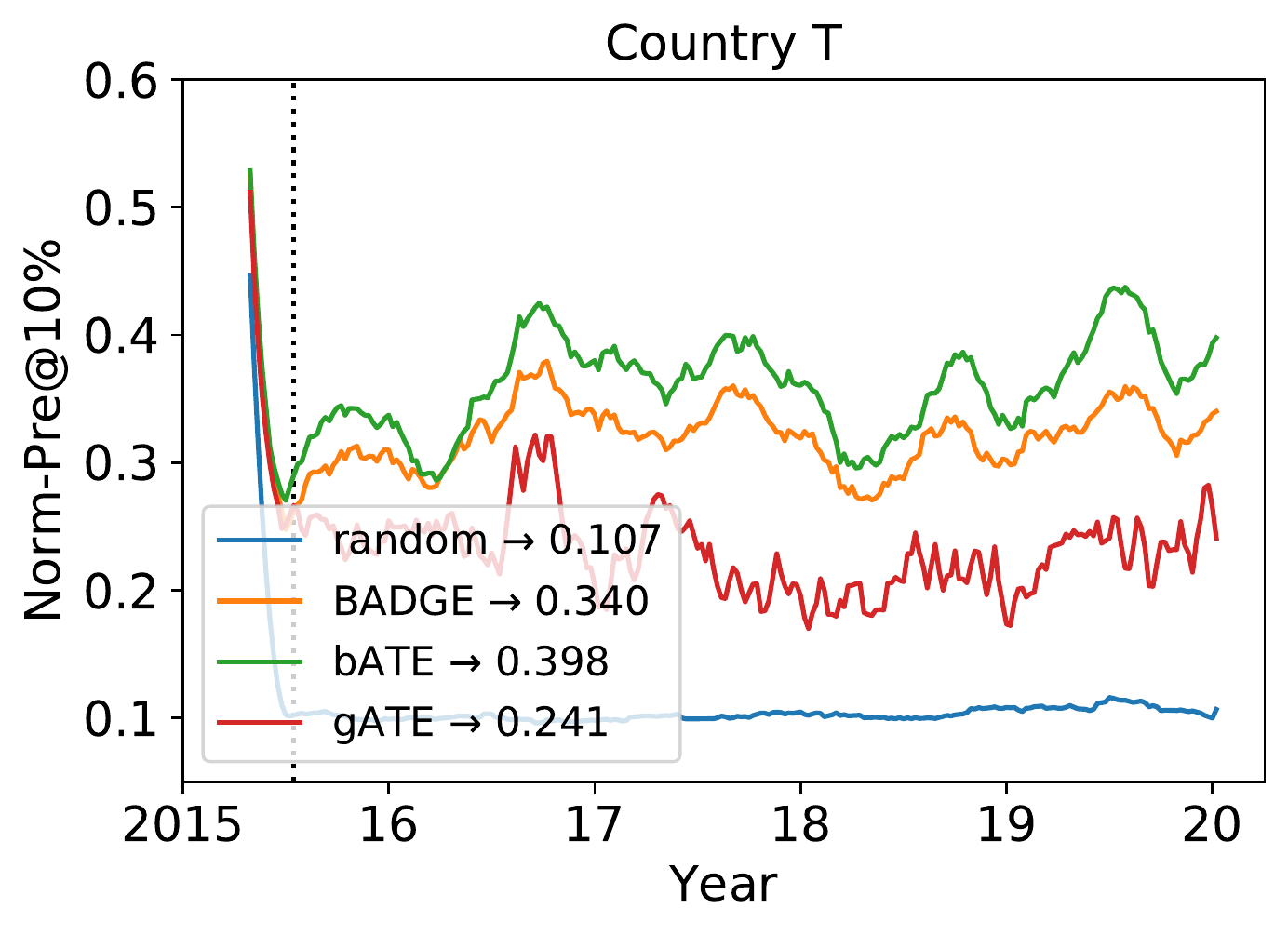}
        \caption{\textsf{Norm-Pre@10\%} performance of four exploration strategies on three country datasets.}
    \end{subfigure}
    \caption{The performance of the advanced exploration strategy outperforms random selection when the customs trade selection system is operated by the exploration strategy. Note that random exploration is widely used in many customs offices. In addition, the performance of \bate{} outperforms \badge{}, suggesting that the introduced scaling components are practical on active customs trade selection settings.}
    \label{fig:pure-exploration}
\end{figure*}
\begin{figure*}[t!]
    \centering
    \begin{subfigure}[b]{\linewidth}
        \centering
        \includegraphics[width=.30\linewidth]{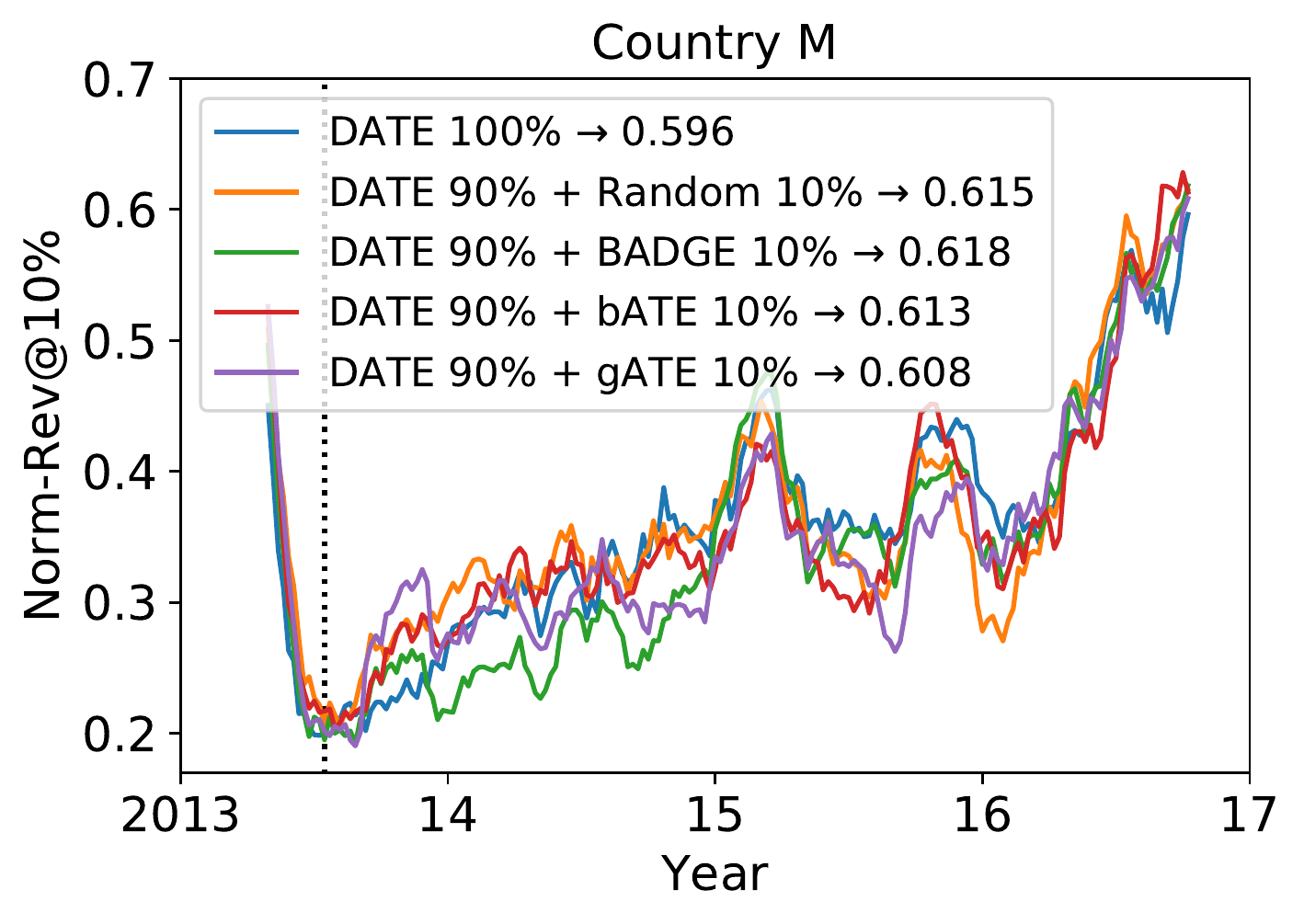}
        \includegraphics[width=.30\linewidth]{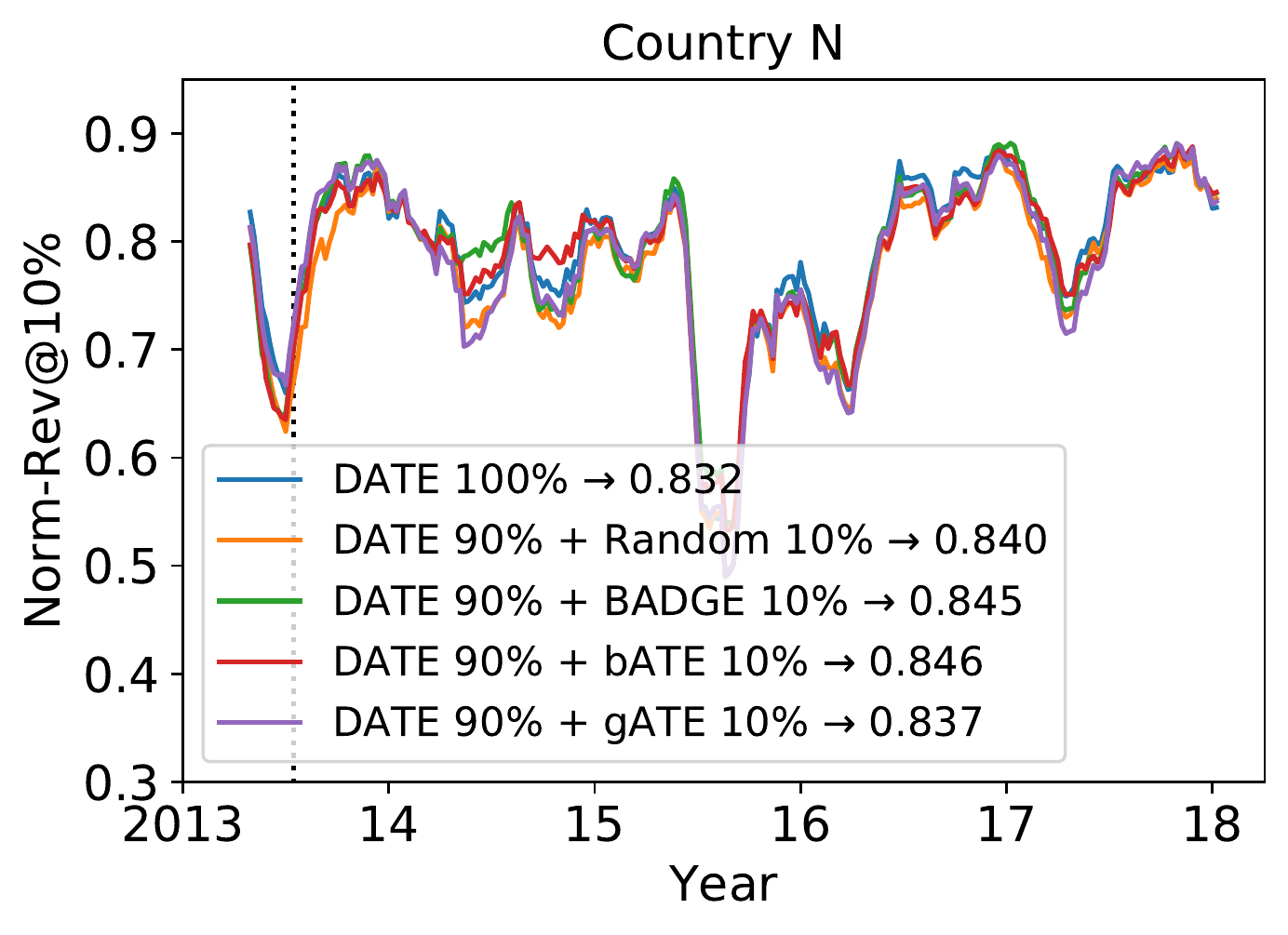}
        \includegraphics[width=.30\linewidth]{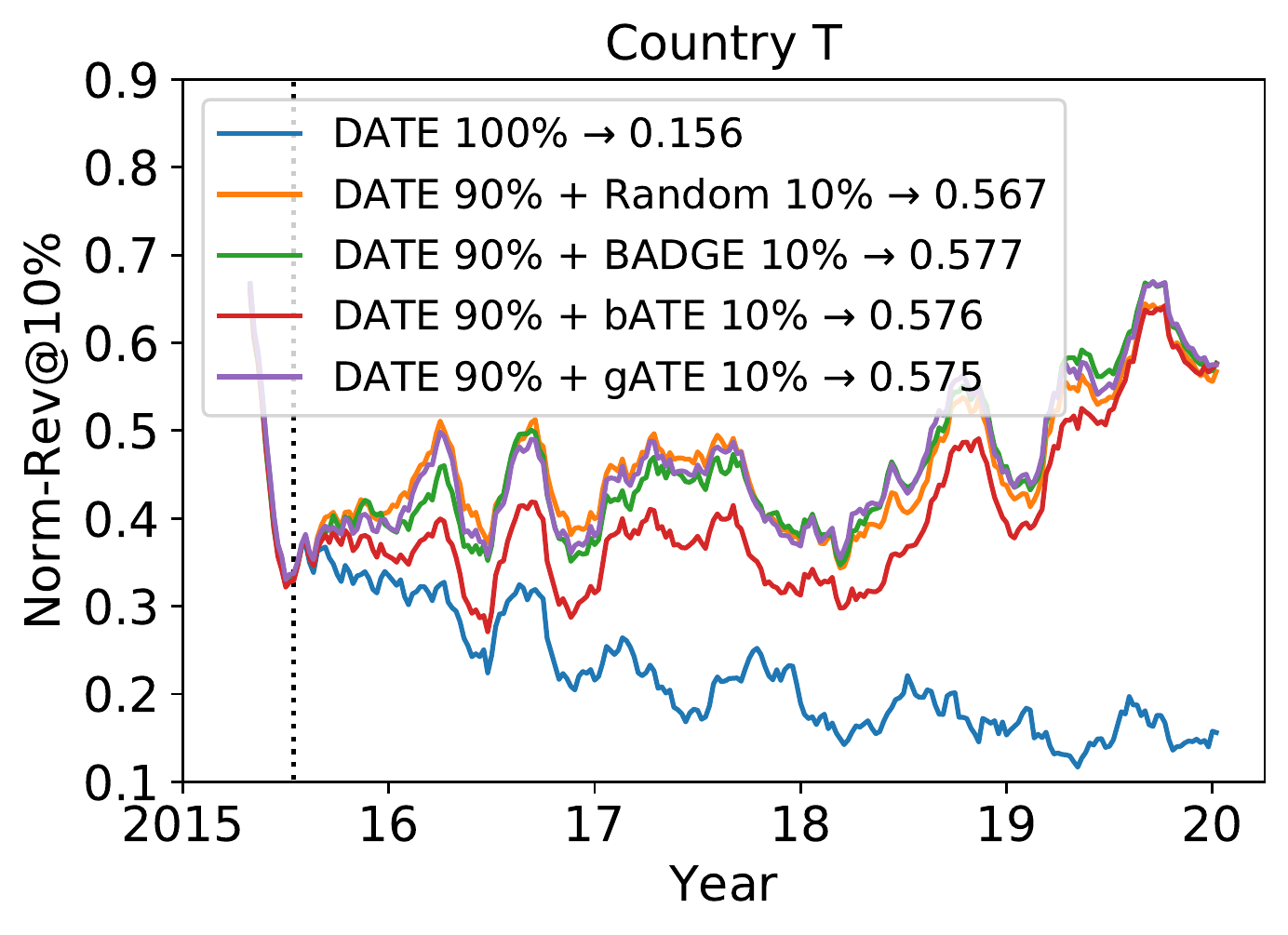}
        \caption{\textsf{Norm-Rev@10\%} performance of four hybrid strategies on three country datasets.}
    \end{subfigure}
    \begin{subfigure}[b]{\linewidth}
        \centering
        \includegraphics[width=.30\linewidth]{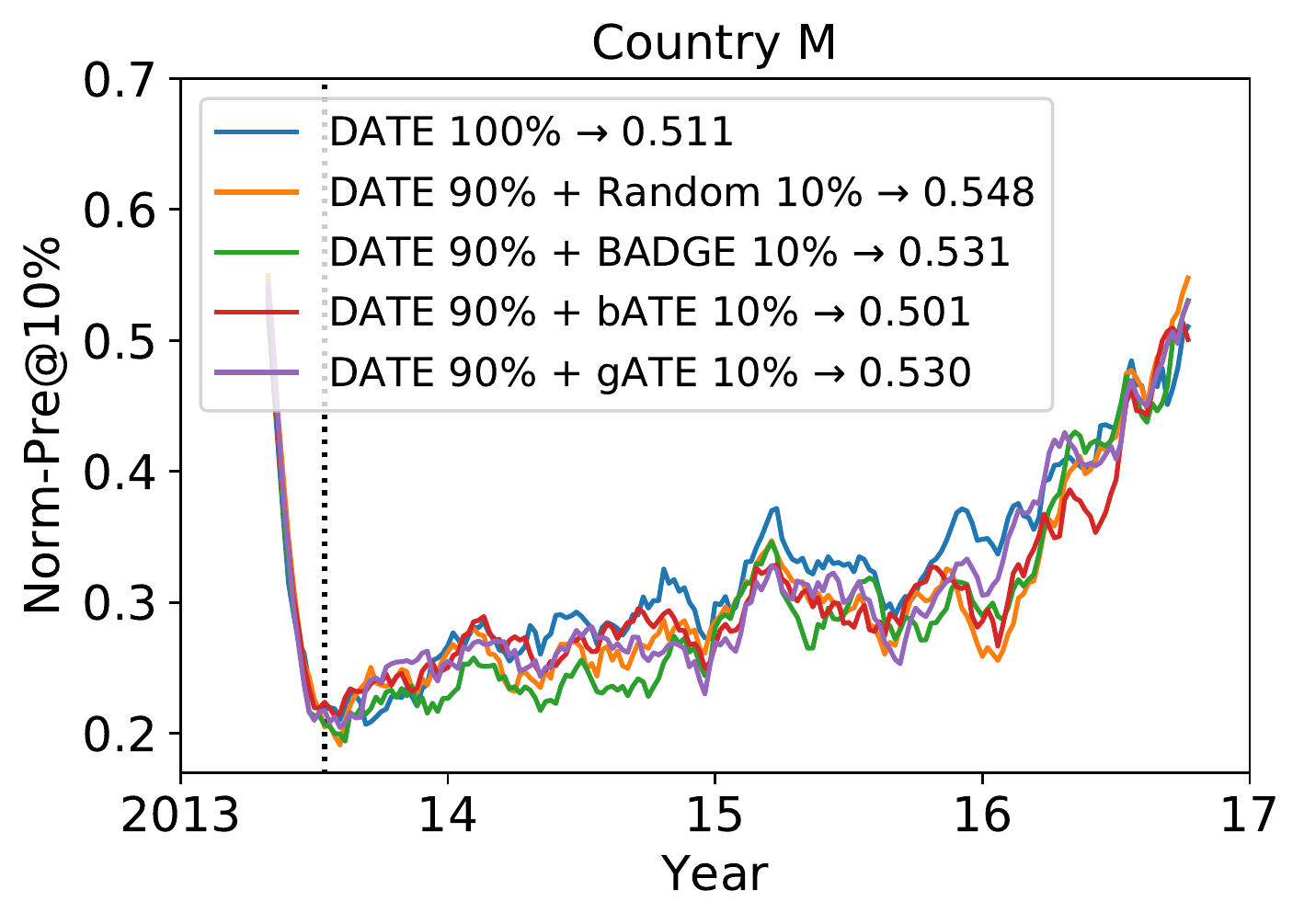}
        \includegraphics[width=.30\linewidth]{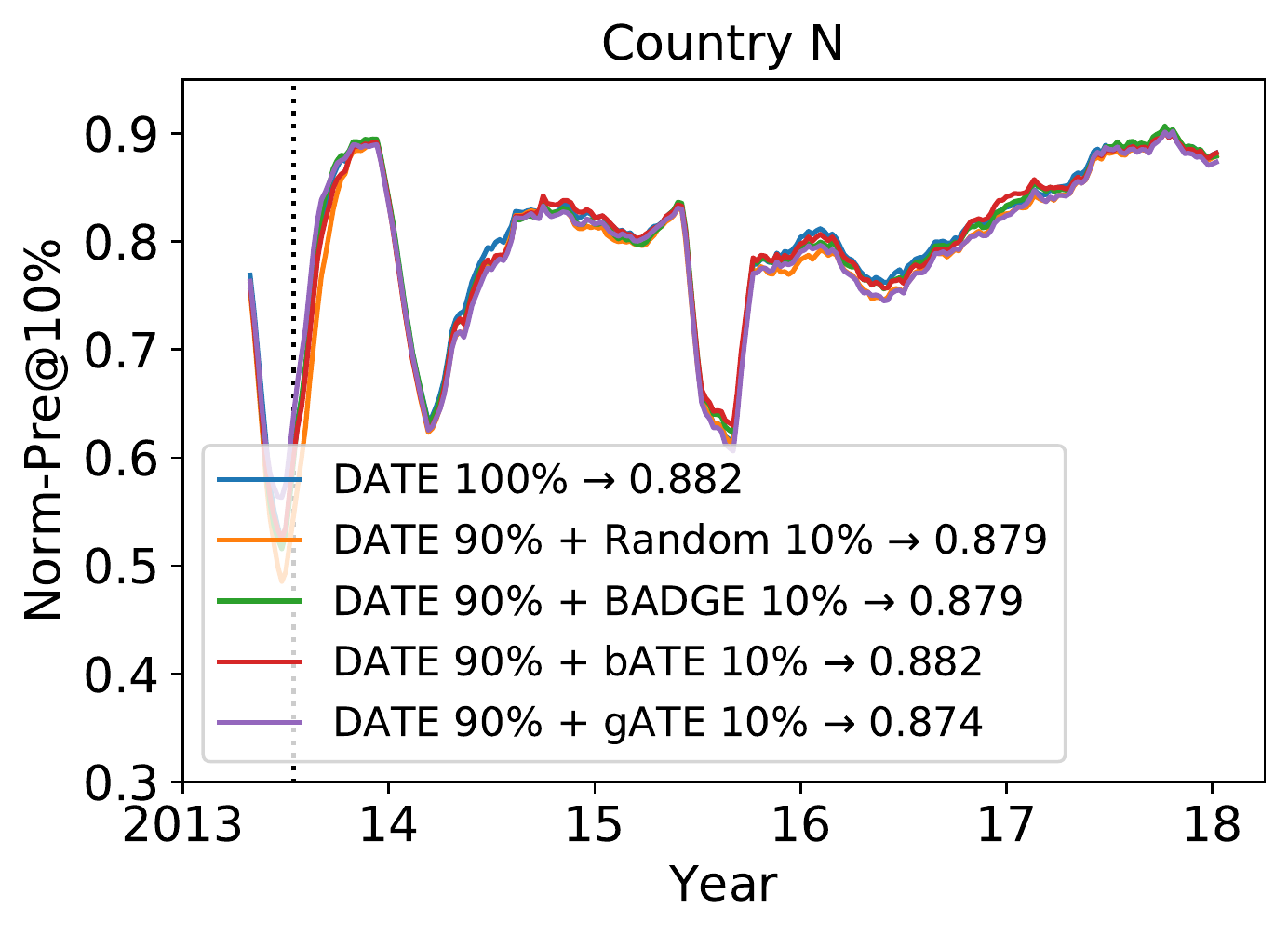}
        \includegraphics[width=.30\linewidth]{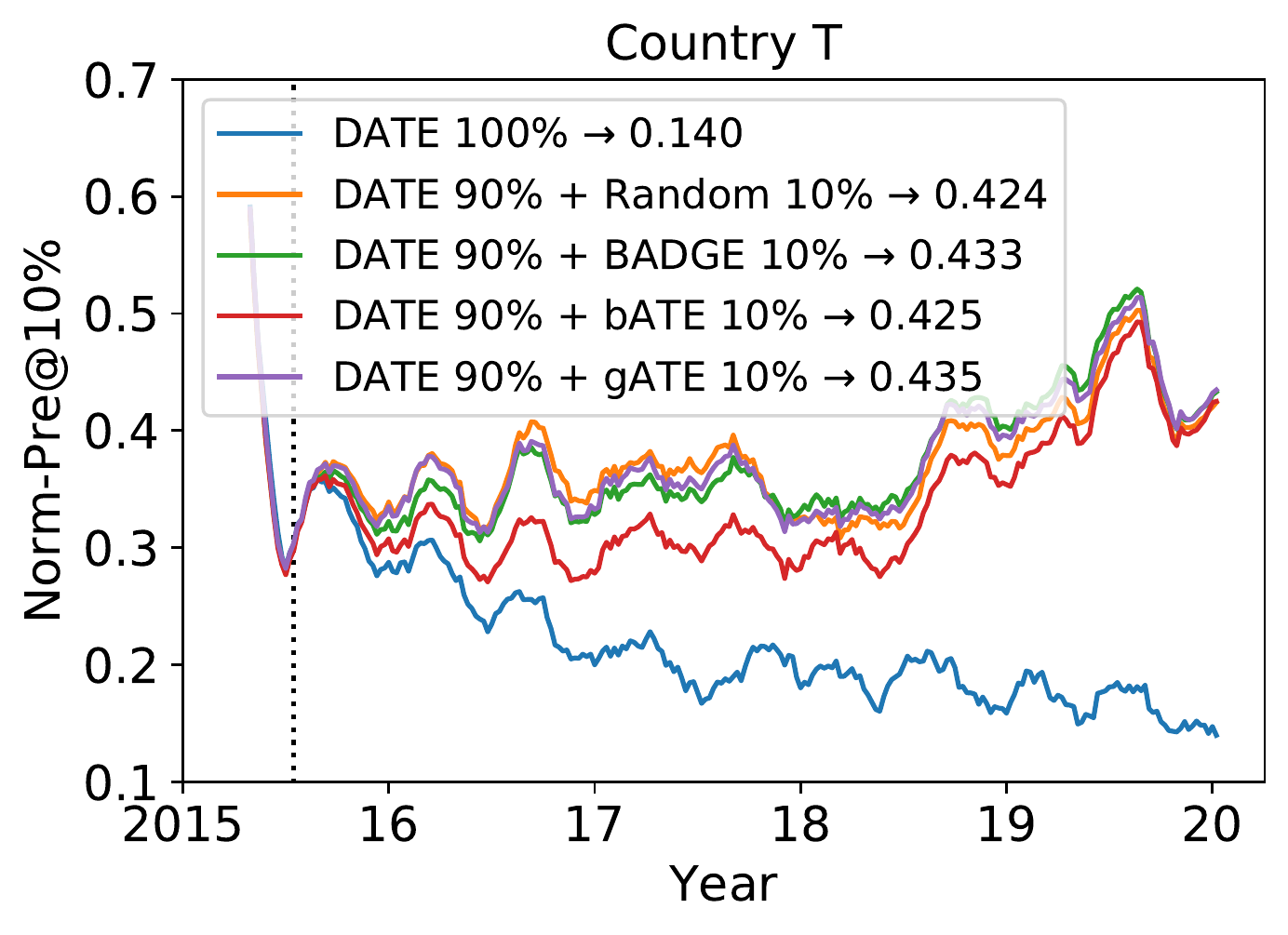}
        \caption{\textsf{Norm-Pre@10\%} performance of four hybrid strategies on three country datasets.}
    \end{subfigure}
    \caption{Hybrid strategies outperforms the \emph{state-of-the-art} fully exploitation strategy \dat{}. We confirm that a robust hybrid model can be made even with a simple exploration strategy.}
    \label{fig:exploration-for-hybrid}
\end{figure*}
\begin{table*}[h!]
\centering
% \small{
\caption{Summary of the overall simulation results. The first number denotes \textsf{Norm-Rev@10\%}, and the second number denotes \textsf{Norm-Pre@10\%} of the model. ($\dag$ Along with three countries dataset, we tested our approach on synthetic data. See the reproducibility section for detail.)}
\label{tab:performance_comparison_table}
\resizebox{\linewidth}{!}{%
\begin{tabular}{c | c c | c c | c c | c c}
\toprule
\multicolumn{1}{c}{} & \multicolumn{2}{|c}{\textbf{Country \textsf{M}}} &
\multicolumn{2}{|c}{\textbf{Country \textsf{N}}} &
\multicolumn{2}{|c}{\textbf{Country \textsf{T}}} & \multicolumn{2}{|c}{\textbf{Synthetic\textsuperscript{$\dag$}}}  \\ \cmidrule{2--3} \cmidrule{3--4} \cmidrule{4--5} \cmidrule{5--6} \cmidrule{6--7} \cmidrule{7--8} \cmidrule{8--9} \cmidrule{9--10}
\multicolumn{1}{c}{\textbf{Model}} &
\multicolumn{1}{|c}{Without \dat{}} &
\multicolumn{1}{c}{With \dat{}} &
\multicolumn{1}{|c}{Without \dat{}} &
\multicolumn{1}{c}{With \dat{}} &
\multicolumn{1}{|c}{Without \dat{}} &
\multicolumn{1}{c}{With \dat{}} &
\multicolumn{1}{|c}{Without \dat{}} &
\multicolumn{1}{c}{With \dat{}} \\
\multicolumn{1}{c}{} &
\multicolumn{1}{|c}{(Fully exploration)} &
\multicolumn{1}{c}{(Hybrid)} &
\multicolumn{1}{|c}{(Fully exploration)} &
\multicolumn{1}{c}{(Hybrid)} &
\multicolumn{1}{|c}{(Fully exploration)} &
\multicolumn{1}{c}{(Hybrid)} & 
\multicolumn{1}{|c}{(Fully exploration)} &
\multicolumn{1}{c}{(Hybrid)}\\
\midrule 
\gate{}         & 0.282 / 0.240   & 0.608 / 0.530   & 0.521 / 0.380   & 0.837 / 0.874   & 0.258 / 0.241   & 0.575 / 0.435  & 0.103 / 0.103 & 0.292 / 0.264 \\
\bate{}         & 0.382 / 0.329   & 0.613 / 0.501   & 0.418 / 0.344   & 0.846 / 0.882   & 0.495 / 0.398   & 0.576 / 0.425  & 0.169 / 0.163 & 0.292 / 0.273 \\
\badge{}        & 0.296 / 0.252   & 0.618 / 0.531   & 0.462 / 0.371   & 0.845 / 0.879   & 0.423 / 0.340   & 0.577 / 0.433  & 0.166 / 0.161 & 0.287 / 0.268 \\
\textsf{Random} & 0.079 / 0.093   & 0.615 / 0.548   & 0.100 / 0.109   & 0.840 / 0.879   & 0.111 / 0.107   & 0.567 / 0.424  & 0.095 / 0.096 & 0.291 / 0.256 \\
\midrule
\dat{} only & - & 0.596 / 0.511 & - & 0.832 / 0.882 & - & 0.156 / 0.146 & - & 0.303 / 0.294 \\
\bottomrule
\end{tabular}
}
\end{table*}
\subsection{Finding the Best Exploration Strategy}
\label{sec:experiments:bestExplore}
A natural question arises regarding which strategies would be best for exploration. We first compared the performance of pure exploration strategies, assuming a need to build a system that tends to explore. 
% This experimental setting may not be very \emph{realistic} for advanced countries that maintain clean enough histories to train a model and would like to exploit their knowledge. However,
This experimental setting is necessary for customs administration where there are not enough import histories available, so customs want to construct the working selection system as quickly as possible. This experimental setting is also widely used in the active learning community~\cite{kirsch2019batchbald, Ash2020badge} to compare performances between pool-based active learning algorithms. We performed experiments with four exploration strategies, including our proposed model designed in Section~\ref{sec:model:exploration}.
\begin{itemize}[leftmargin=*]
    \item \textbf{\textsf{Random}}~\cite{han2014kcs}: Known to be used as an exploration strategy to detect novel fraud in the production systems of some countries.
    \item \textbf{\textsf{BADGE}}~\cite{Ash2020badge}: State-of-the-art active learning approach that selects items considering uncertainty and diversity.
    \item \textbf{\textsf{bATE}}: Explores by considering predicted revenue as well as item uncertainty and item diversity. 
    \item \textbf{\textsf{gATE}}: Strategically determines whether the exploration strategy is random or \textsf{bATE}, depending on the performance of the base model.
\end{itemize}

% Figure~\ref{fig:pure-exploration} shows the performance of exploration strategies. We report each strategy's average performance in the last quarters (13 weeks) with overall trends.

Figure~\ref{fig:pure-exploration} shows the 13-week moving average performance of the exploration strategies. The results show that the three advanced strategies, \badge{}, \bate{}, and \gate{}, outperform the random strategy by a large margin. \bate{} is the top-performing strategy in countries \textsf{M} and T, and \gate{} performs the best in country \textsf{N}. In the point of view of active learners, these results suggest that the introduced scaling components in our method play a role in constructing the working customs trade selection system more rapidly.

However, it turns out that the performance of the full-exploration strategy is not comparable to the full-exploitation strategy. In our experiments, the performance of the full-exploitation strategy reaches 0.844 in country N (Table 4), while the performance of the full-exploration strategy is 0.52 in the same country, which is not impressive in itself. \sd{Although a set of explored samples consists of items with uncommon HS codes or under-invoiced item near the decision boundary, they are not always frauds. Including these items is helpful to the model training process to some extent. However, a model solely trained on these items (i.e., full exploration) is susceptible to noise and hence does not exhibit the best performance.} It can be seen that the exploration strategy and exploitation strategy need to be used \emph{together} to guarantee the reliable performance of the customs trade selection system. 

\looseness=-1
 
\subsection{Best Exploration Strategy for the Hybrid Model}
\label{sec:experiments:forHybrid}
% \brian{Can we combine 5.4 and 5.5 into a single section?} -> I also considered, but separating it would be better since the conclusions are different, and there is enough content.  
Next, we compare the performance of these exploration strategies by applying them with an exploitation strategy. Following Section~\ref{sec:experiments:exploitationFails}--\ref{sec:experiments:exploitationNotFails}, each hybrid strategy selects 90\% of the items by \dat{}, and the four exploration strategies select the remaining 10\% of the items. We also compare the strategies with \dat{} to show the long-term sustainability of the hybrid strategies. 

%   summarizes the overall simulation results. The average \textsf{Norm-Rev@n\%} and \textsf{Norm-Pre@n\%} over the previous 13 weeks are recorded. For the synthetic results, please refer to Figure~\ref{fig:synthetic-results}.
 
Figure~\ref{fig:exploration-for-hybrid} and Table~\ref{tab:performance_comparison_table} summarize the performance of the hybrid models with different exploration strategies. First, we can see that all hybrid strategies outperform a full-exploitation strategy \dat{} by some margin. For country \textsf{T}, where a staggering decline in the exploitation strategy performance is recorded, our hybrid strategy performs exceptionally stably, and the model ultimately improves. Even though the \dat{} model for exploitation remains effective for the other two countries, the 10\% trade-off for exploration does not hurt the overall performance; rather, this method slightly outperforms the exploitation algorithm. This proves our initial claim that even if we inspect suspicious items, we can guarantee similar performance by learning new patterns from the unknown items. 

Second, advanced exploration strategies can help to improve the performance of the whole hybrid model as much as possible. This is shown by the result that \dat{}+\bate{} achieves 1.6\% higher revenue than \dat{}+random in country \textsf{T}. It is noteworthy that the best exploration algorithm contributes the most to the hybrid strategy when it is used in the country with the largest trade volume and the highest illicit rate. \looseness=-1 
% However, it is noteworthy that the best exploration algorithm does not always contribute the most to the hybrid strategy.

Third, the hybrid model's performance with a random exploration strategy is still comparable to the hybrid model in general. In practice, we encourage customs administration to start with a simple exploration strategy without using additional computing power. In contrast to relying on single exploitation or exploration model, the customs trade selection model will be improved even more robustly with both strategies.

\begin{figure}[h!]
  \centering
  \includegraphics[width=1\columnwidth]{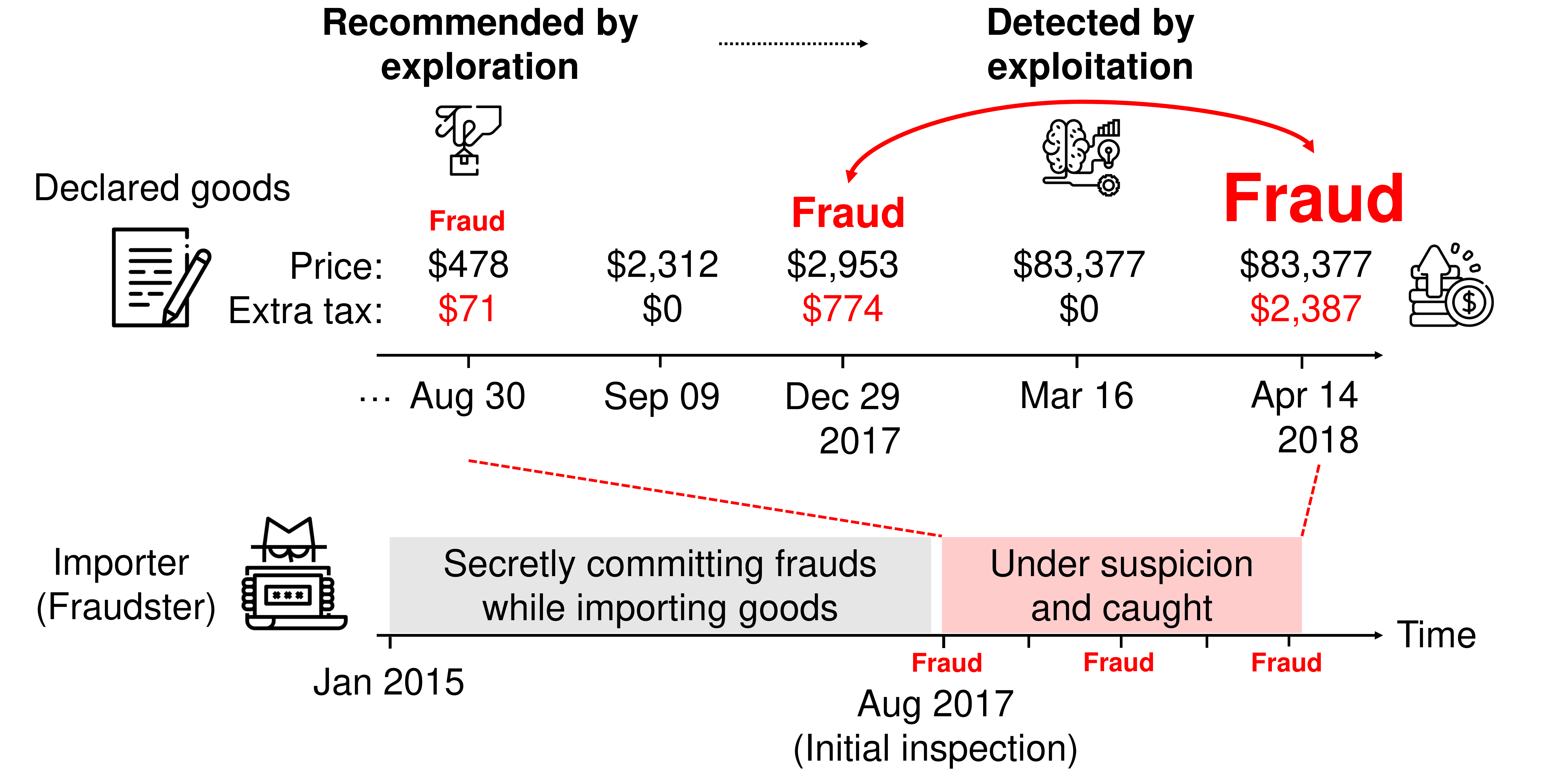}
  \caption{\sd{Trade log of a fraudster: Thanks to the exploration strategy which noticed the importer's fraud action, the hybrid targeting system is later able to detect major frauds he committed. The system that only operates the exploitation strategy could not detect the frauds until the end.}}
  \label{fig:case_studies}
\end{figure}

\subsection{\sd{Case Study}}
\label{sec:experiments:case-studies}
\sd{Timely exploration allows customs to inspect goods from unknown importers and extend their knowledge. Based on this input, the updated model can prevent potential frauds.
Figure~\ref{fig:case_studies} shows a successful case of detecting sequential frauds by our hybrid strategy. This example introduces a trade log of an importer who has imported goods since 2015. After 2.5 years, one of the transactions is subjected to physical inspection by exploration and was labeled a fraud.\footnote{\sd{From Jan 2015 to Aug 2017, the importer processed 59 transactions, including eight frauds, but none of these transactions was inspected yet.}} The importer mixed normal and fraudulent transactions to avoid further inspection. Yet, the newly updated exploitation strategy was able to catch his sequential frauds. Without being triggered by exploration (i.e., fully-exploitation), the targeting system would not have detected frauds from unidentified importers. In our experiment, 1,652 importers and their 170,683 items followed the same pattern (i.e., sequential and sporadic frauds) and were subjected to inspection by hybrid strategy. Among them, only 74 importers and their trades are inspected by the full-exploitation model.}
\begin{figure}[h!]
    \centering
    \resizebox{\linewidth}{!}{
    \subfloat[\sd{Breakdown}]{           
        \includegraphics[width=0.27\linewidth]{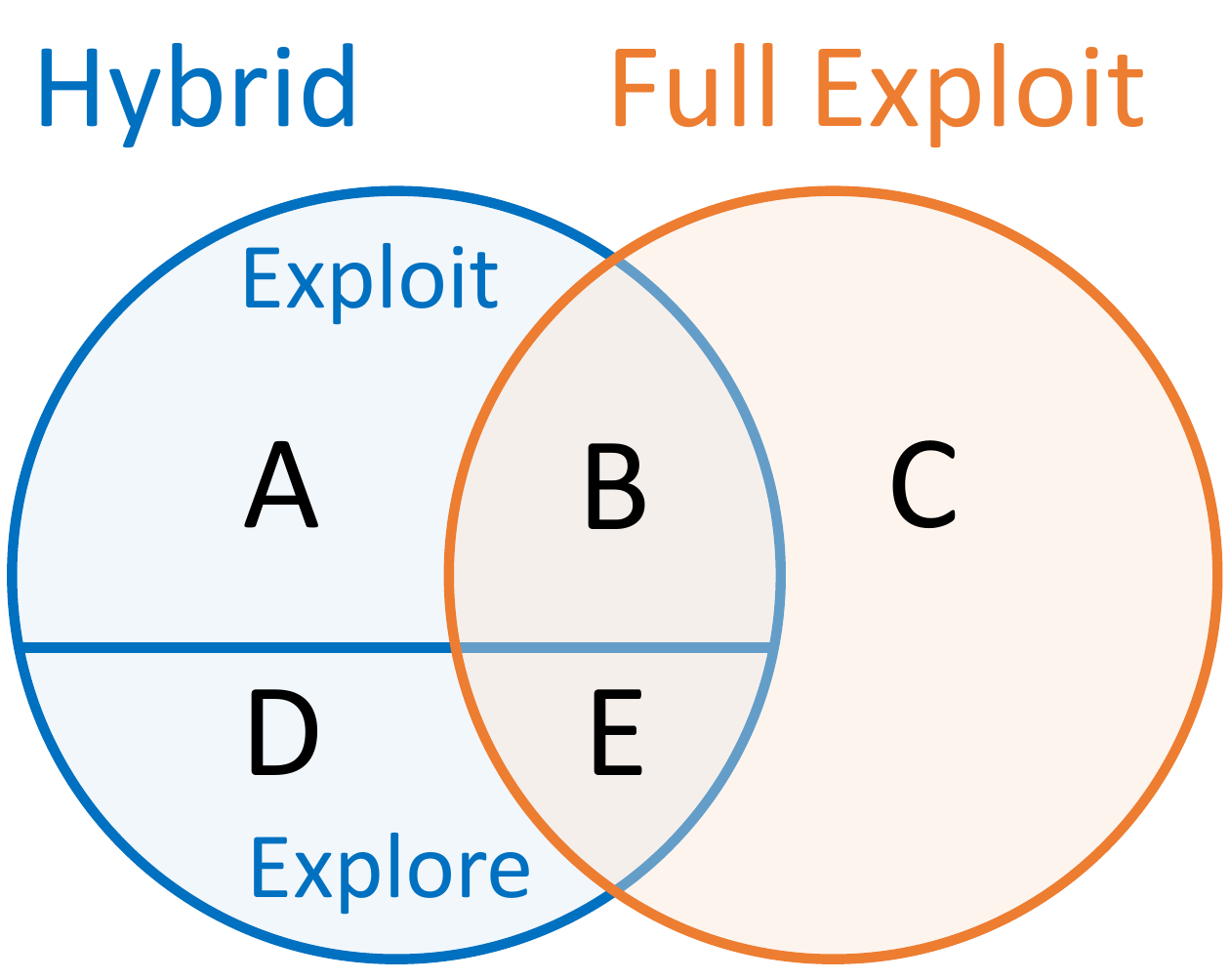}
        \label{fig:ma}          
        }
    \subfloat[\sd{Statistics of each component}]{
        \adjustbox{valign=b}{
        \resizebox{0.72\linewidth}{!}{%
        \begin{tabular}{ l r r r r r } \toprule
        Inspected items &  A  &  B  &  C  &  D  &  E  \\ \midrule
        \# imports & 216,935 & 152,914 & 255,221 & 38,286 & 2,814  \\ 
        \# importers &  7,547 & 1,216 & 785 & 8,509 & 235  \\ 
        Fraud rate & 36.4\% & 21.7\% & 5.9\% & 5.8\% & 5.5\%   \\ 
        Avg. revenue &  \$1,882 & \$782 & \$138 & \$180 & \$129   \\ 
        \bottomrule
        \end{tabular}
        \label{res-ma}
        }}}
    }
    \caption{\sd{The statistics of inspected items chosen by each algorithm in country T. For example, the group D represents the items explored and inspected by the hybrid model, but not selected by the full exploitation model. Note the gap between groups A and C.}} % Component-level result analysis in country T. 
    \label{fig:detailed-analysis}    
\end{figure}

% \begin{figure}[t!]
%     \begin{subfigure}[b]{0.33\columnwidth}
%         \centering
%         \includegraphics[width=\linewidth]{figures/detailed-analysis.pdf}
%     \end{subfigure}
%     \begin{subfigure}[b]{0.66\columnwidth}
%         \begin{tabular}{ l r r r r r } \toprule
%         Datasets &  (1)  & (2) & (3) & (4) & (5)  \\ \midrule
%         \# imports & 216,935 & 152,914 & 255,221 & 38,286 & 2,814  \\ 
%         \# importers &  7,547 & 1,216 & 785 & 8,509 & 235  \\ 
%         Avg. illicit rate & 36.4\% & 21.7\% & 5.9\% & 5.8\% & 5.5\%   \\ 
%         Avg. revenue &  1,882 & 782 & 138 & 180 & 129   \\ 
%         \bottomrule
%     \end{tabular}
%     \end{subfigure}
%     \caption{Detailed analysis.}
%     \label{fig:detailed-analysis}
% \end{figure}

% \begin{table}[t!]
% \centering
% % \scriptsize
% \caption{Statistics of the datasets}
% \label{tab:detailed-analysis}
% \resizebox{\linewidth}{!}{%

% }\end{table}
\begin{figure*}[t!]
\centering
    \begin{subfigure}[b]{.325\linewidth}
        \includegraphics[width=\linewidth]{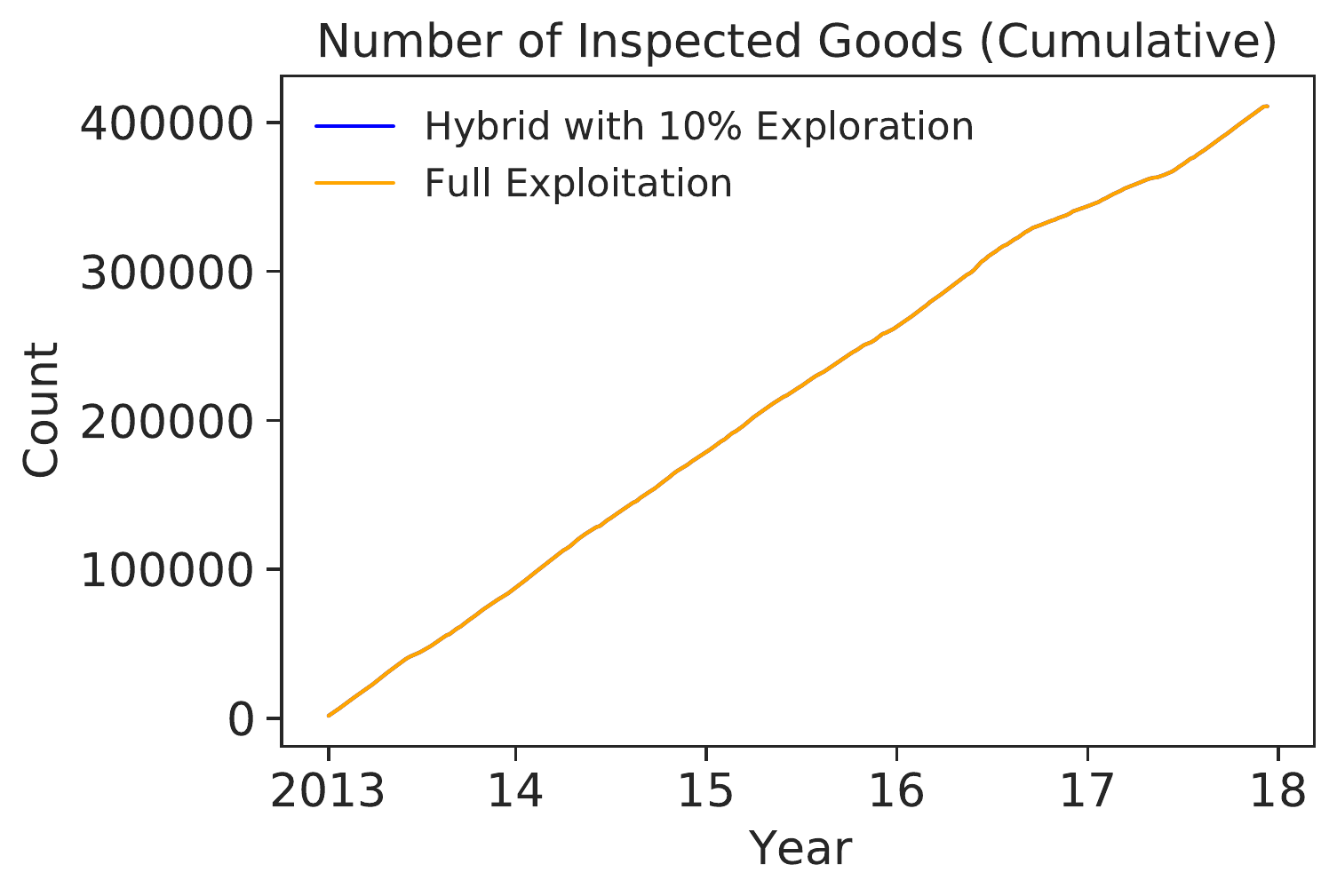}
        \centering\captionsetup{width=.91\linewidth}%
        \caption{\sd{Both strategies inspect 10\% of weekly imports, so the cumulative number of inspected goods between them is the same.}}
    \end{subfigure}
    \begin{subfigure}[b]{.325\linewidth}
        \includegraphics[width=\linewidth]{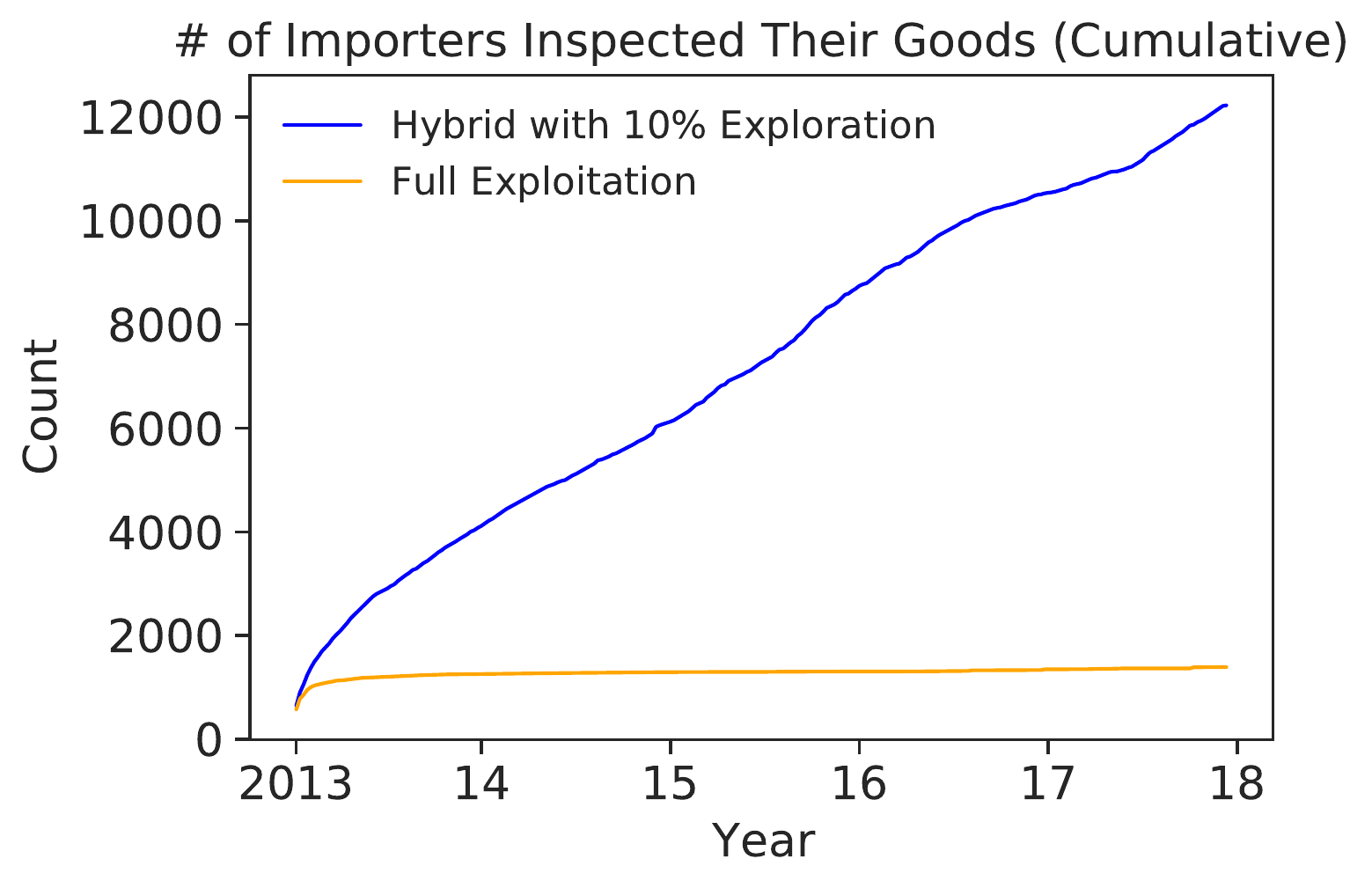}
        \centering\captionsetup{width=.91\linewidth}%
        \caption{\sd{With hybrid strategy, nearly ten times more importers are subjected to inspection compared to the fully-exploitation strategy.}}
    \end{subfigure}
    \begin{subfigure}[b]{.325\linewidth}
        \includegraphics[width=\linewidth]{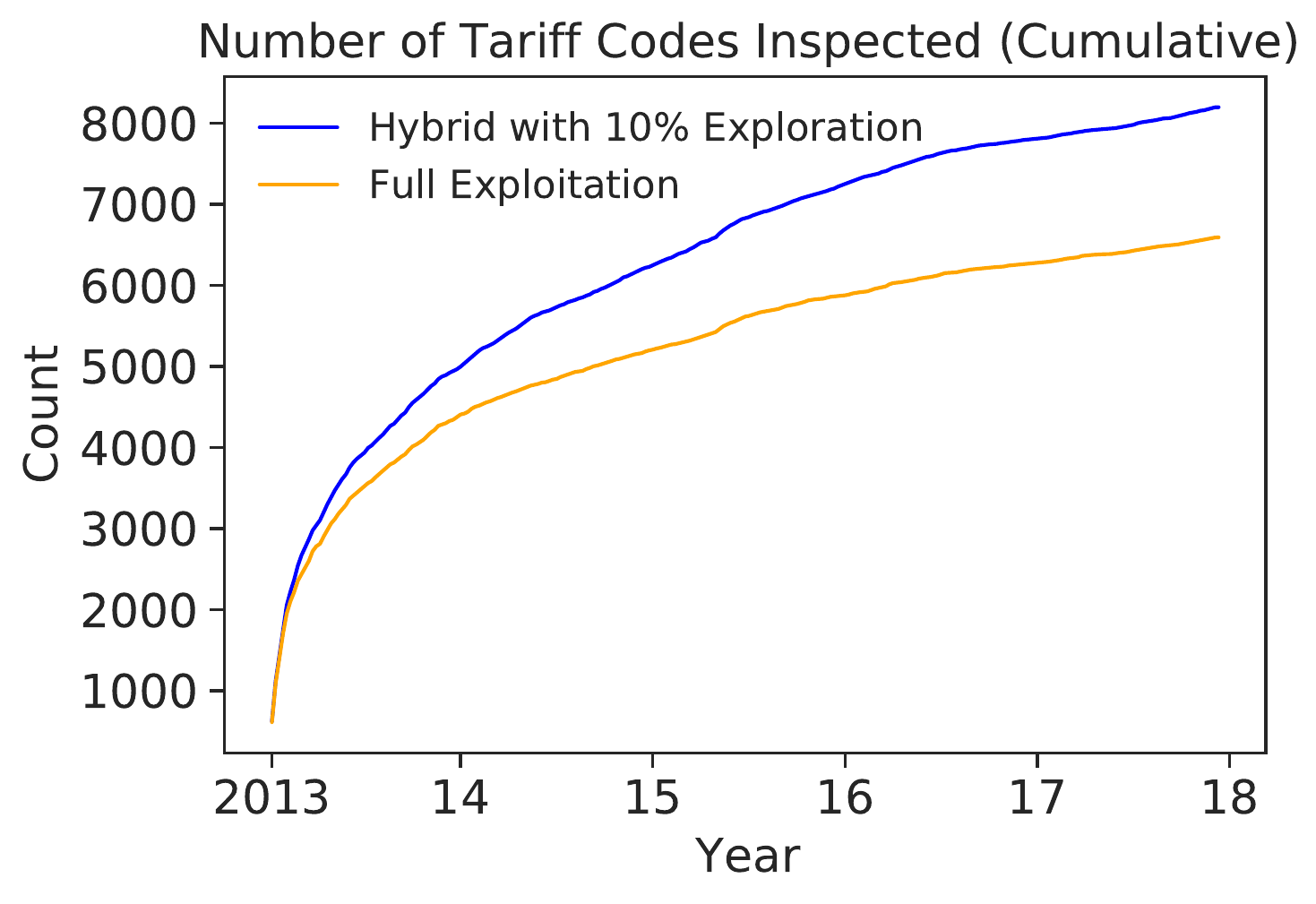}
        \centering\captionsetup{width=.91\linewidth}%
        \caption{\sd{Similarly, hybrid strategies inspect items with diverse tariff codes (categories). After five years, more than 2,000 new categories are discovered.}}
    \end{subfigure}
    \vspace{2mm}
    
    \begin{subfigure}[b]{.325\linewidth}
        \includegraphics[width=\linewidth]{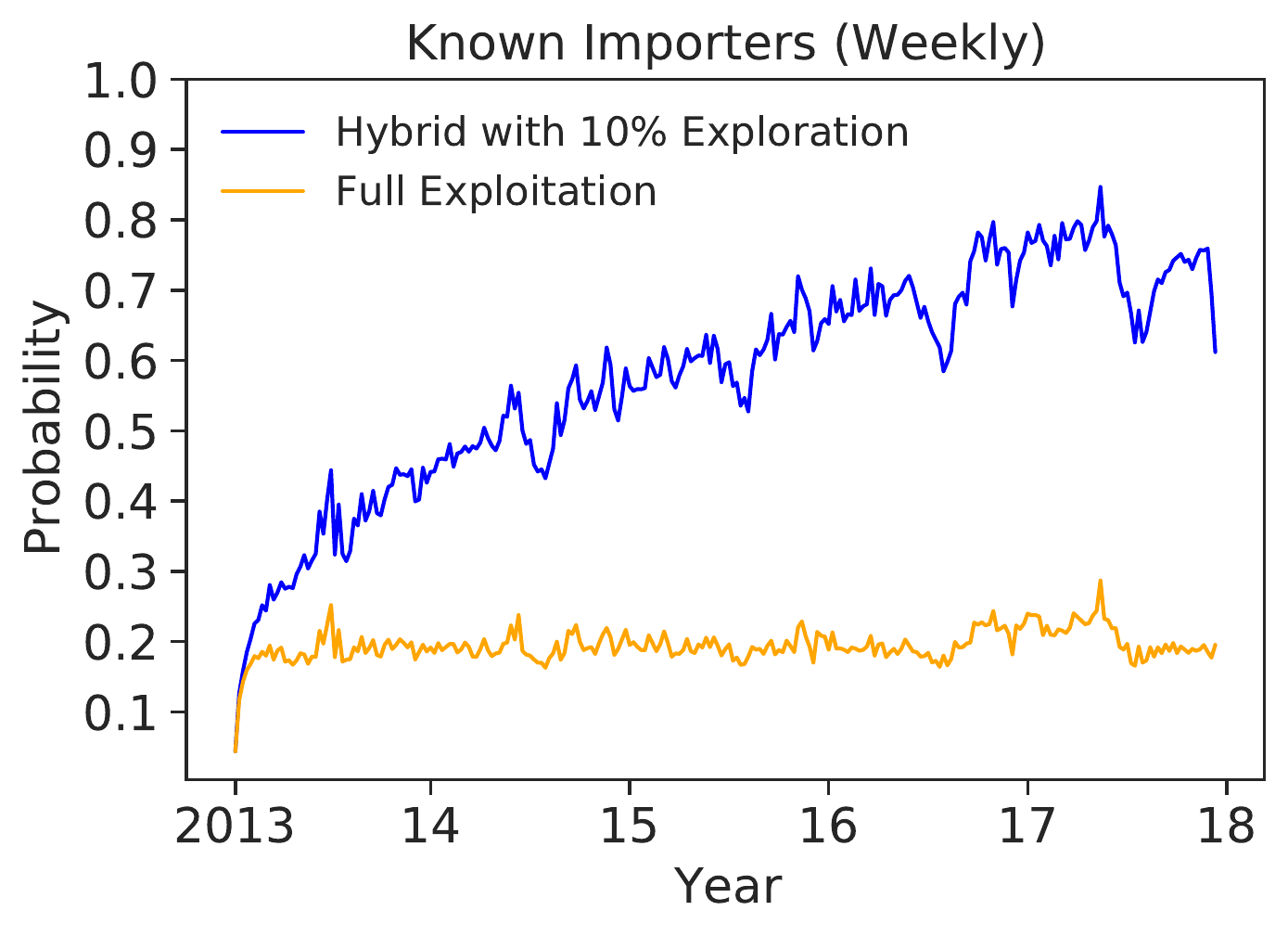}
        \centering\captionsetup{width=.91\linewidth}%
        \caption{\sd{After three years, the hybrid model operates by knowing 70\% of the importers who declare goods in the following week.}}
    \end{subfigure}
    \begin{subfigure}[b]{.325\linewidth}
        \includegraphics[width=\linewidth]{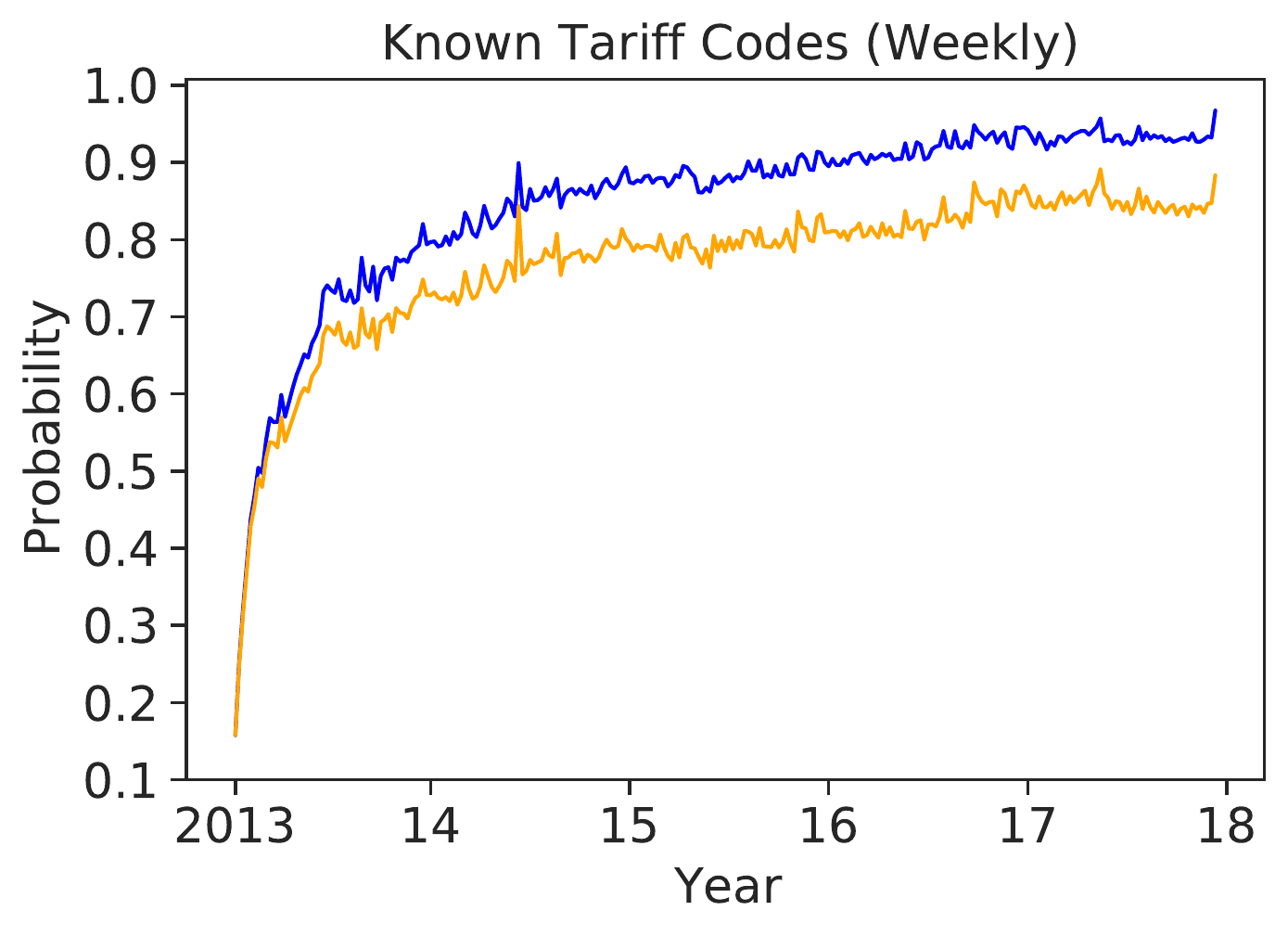}
        \centering\captionsetup{width=.91\linewidth}%
        \caption{\sd{Similarly, compared to the exploitation model, the hybrid model operates by knowing more tariff codes that declared items belong to.}}
    \end{subfigure}
    \begin{subfigure}[b]{.325\linewidth}
        \includegraphics[width=\linewidth]{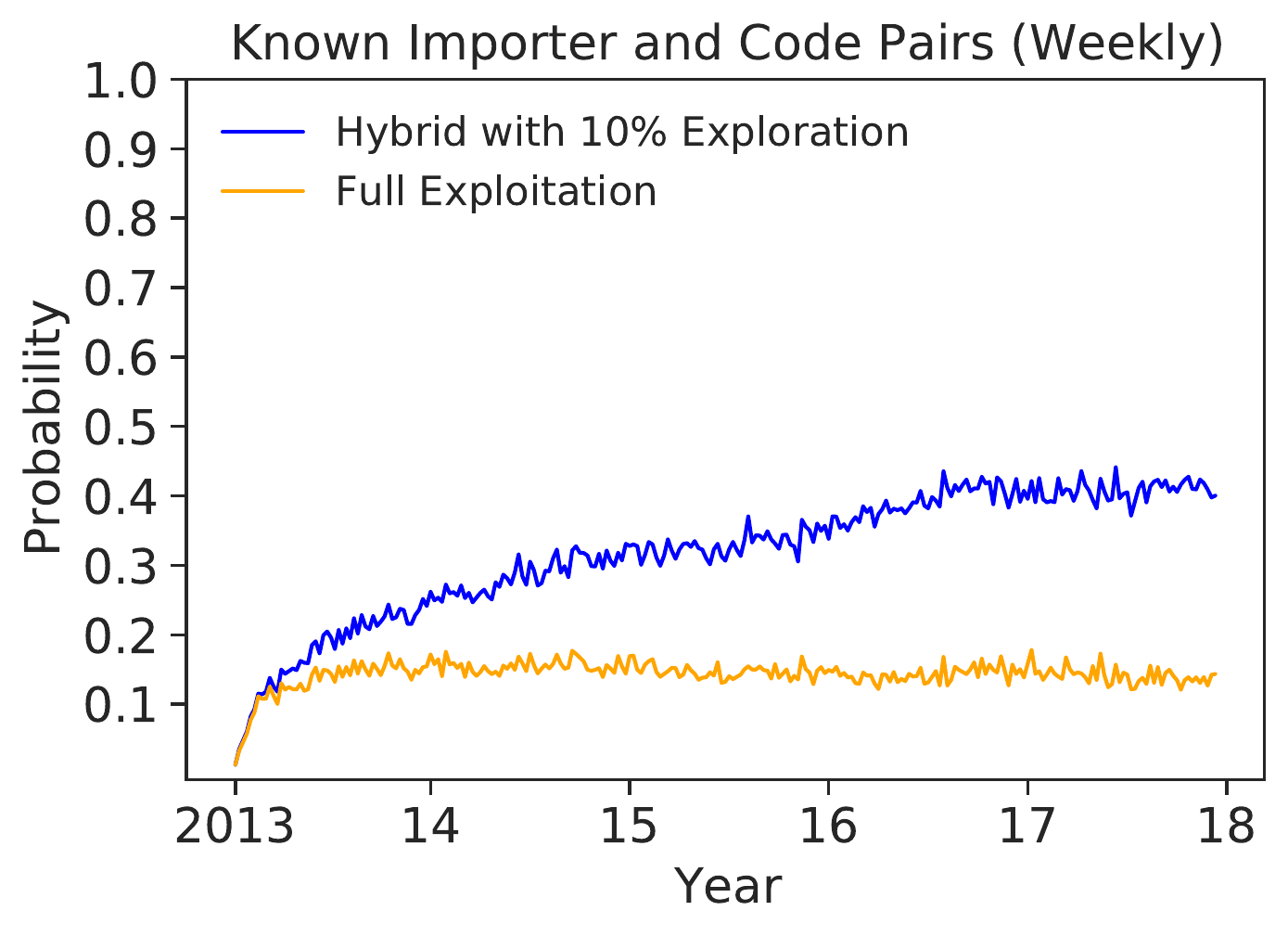}
        \centering\captionsetup{width=.91\linewidth}%
        \caption{\sd{The coverage of the hybrid model is significantly larger, when the amount of known importer and tariff code pairs is concerned.}}
    \end{subfigure}
    \vspace{2mm}
    
    \begin{subfigure}[b]{.325\linewidth}
        \includegraphics[width=\linewidth]{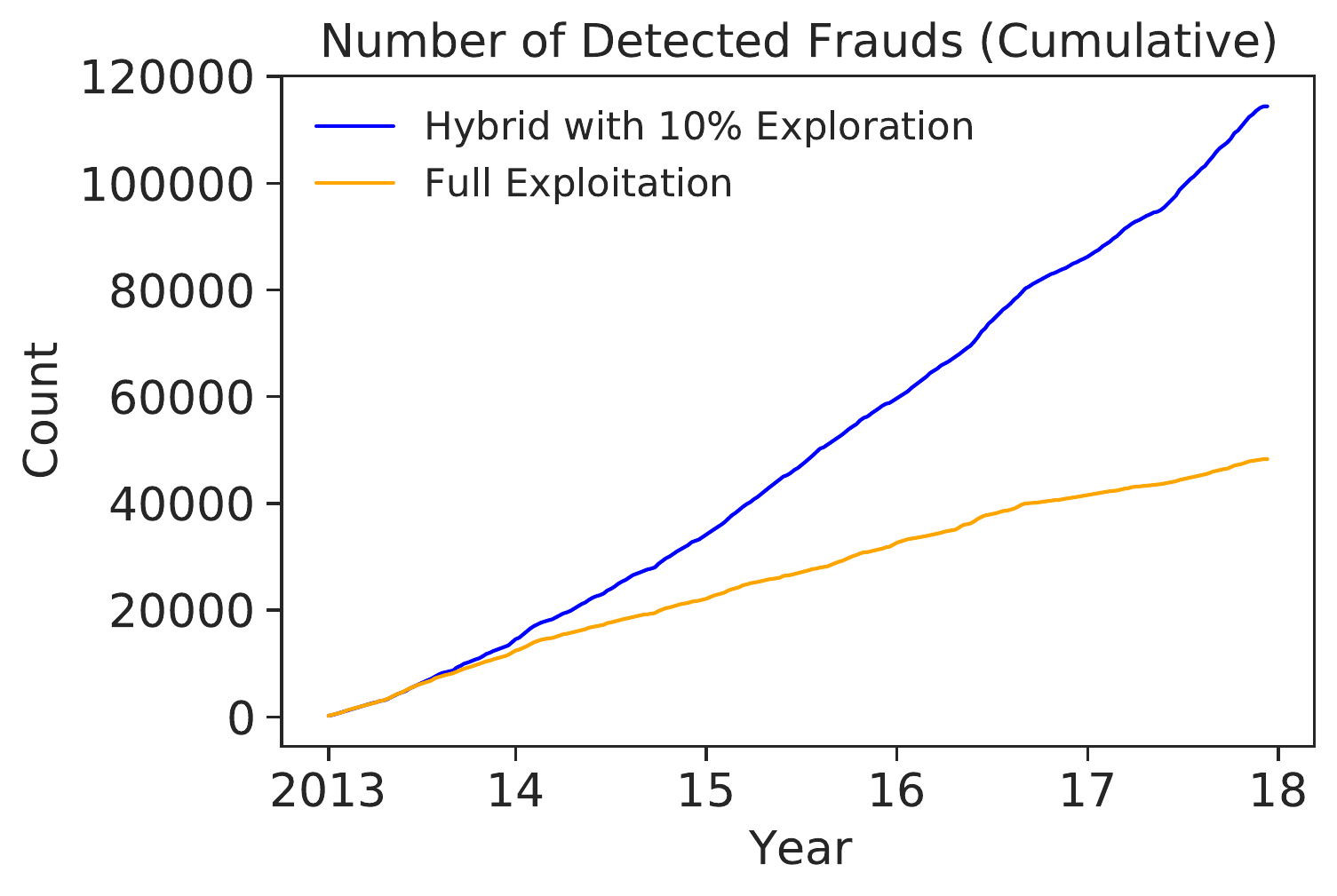}
        \centering\captionsetup{width=.91\linewidth}%
        \caption{\sd{Over the five years, the hybrid model is able to detect frauds more than two times compared to the fully-exploitation model.}}
    \end{subfigure}
    \begin{subfigure}[b]{.325\linewidth}
        \includegraphics[width=\linewidth]{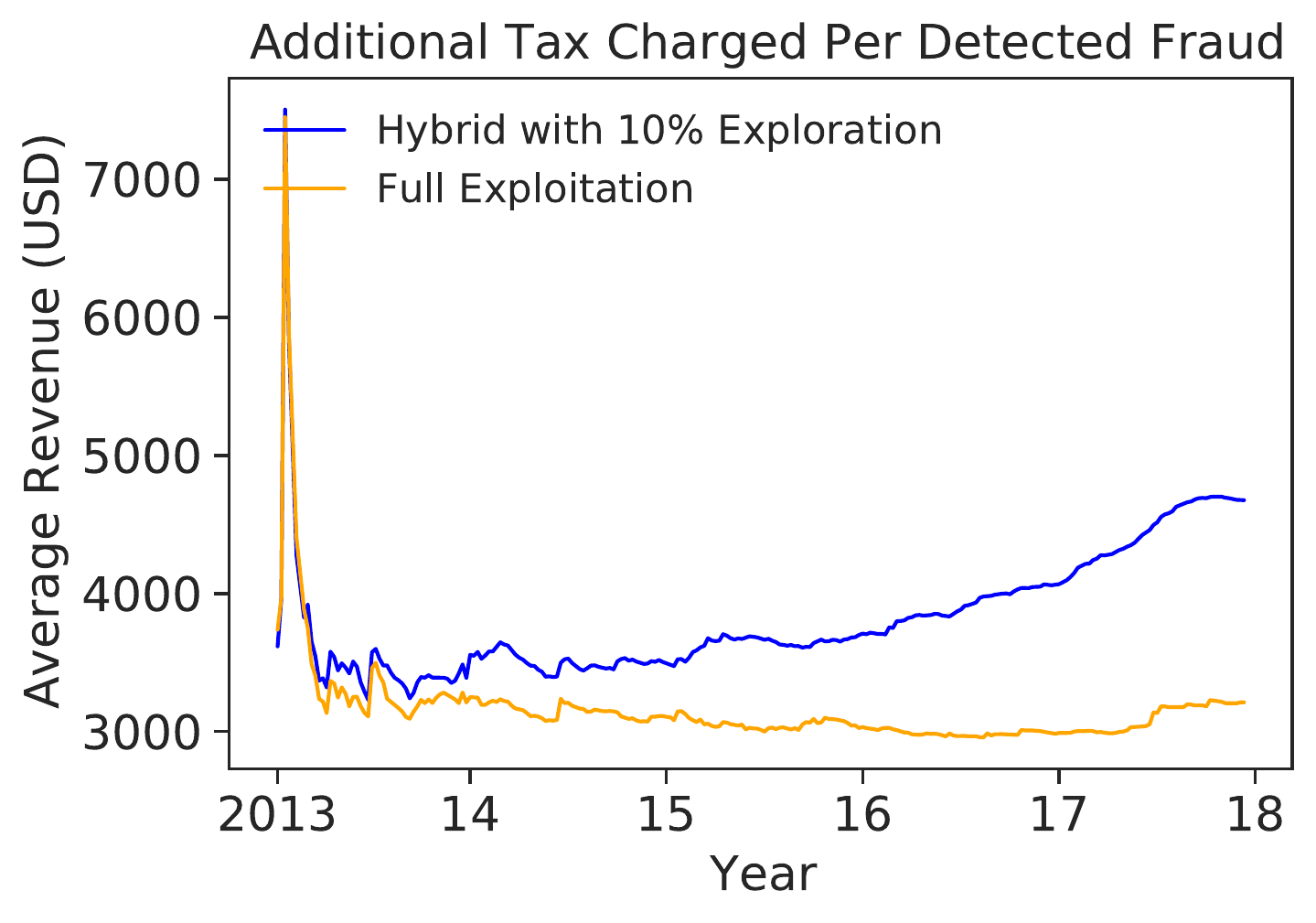}
        \centering\captionsetup{width=.91\linewidth}%
        \caption{\sd{Additionally, each fraud detected by the hybrid model has higher values on average. The value difference also increases over time.}}
    \end{subfigure}
    \begin{subfigure}[b]{.325\linewidth}
        \includegraphics[width=\linewidth]{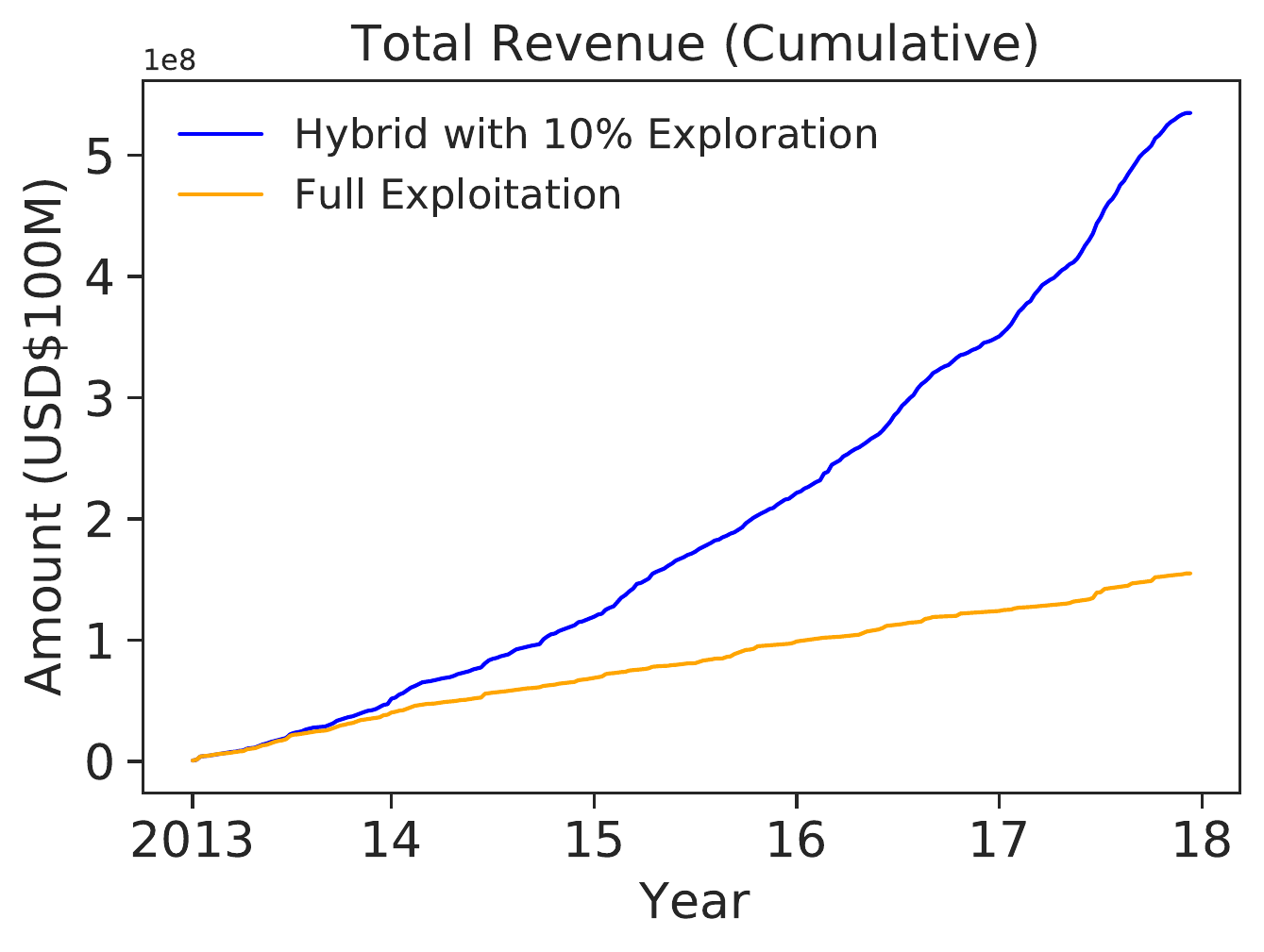}
        \centering\captionsetup{width=.91\linewidth}%
        \caption{\sd{Combining these effects, the hybrid model is able to secure three times more revenue compared to the fully-exploitation model.}}
    \end{subfigure}
    \caption{\sd{Detailed comparison between the hybrid and fully-exploitation-based targeting systems in country T.}}
    \label{fig:stats-exploration-effect}
\end{figure*}
\subsection{\sd{Detailed Analysis}}
\sd{
This case study demonstrates that a timely exploration triggers targeting systems to cope better with frauds from new importers. We further compared the statistics of the selected items between two targeting systems (i.e., hybrid and full exploitation). Figure~\ref{fig:detailed-analysis} illustrates how we break down the results into five components and shows their statistics. In the left figure, $A \cup B$ is a set of items selected by hybrid exploitation, and $D \cup E$ is a set of items selected by hybrid exploration. Likewise, $B \cup C \cup E$ is a set of items selected by the full-exploitation model.} 

\sd{The hybrid targeting system makes better trade selection based on the inputs it receives through exploration. Since the exploration strategy inspected 41,100 items from 8,744 importers ($D \cup E$), the exploitation module selected 369,849 suspicious items from 7,944 importers ($A \cup B$). In contrast, the full-exploitation model operated from a limited importer pool. Total 410,949 items were selected from merely 1,392 importers ($B \cup C \cup E$). In addition, the detection rate of the hybrid model and the corresponding revenue per item are higher than those of the full-exploitation model. For comprehensive understanding, we also compared the performance of the hybrid model and the full-exploitation model on various criteria over time. Detailed results are summarized in Figure~\ref{fig:stats-exploration-effect} with explanations.}

\section{Concluding Remarks}
\label{sec:conclusion}
This paper investigates the human-in-the-loop online active learning problem, where the indicators of the annotated samples are the key criteria for evaluation. One such example can be found in customs inspection, where customs officers need to decide which new cargo to examine (i.e., an exploration strategy) while retaining the history of existing illicit trades (i.e., an exploitation strategy). We present a selection strategy that efficiently combines exploration and exploitation strategies. Our numerical evaluation, based on multiyear transaction logs, provides insights for practical guidelines for setting model parameters in the context of customs screening systems. \looseness=-1 

To facilitate the proposed approach in customs administrations, the model code is open source. It currently supports diverse exploitation and exploration strategies with various tunable parameters ranging from models to simulation settings so that users can confirm whether our proposed work is well suited for their data. With minor adjustments, our code can also support various decision-making problems with constrained resources. 
% Broader audiences of our research include governments dealing with limited budgets. For example, the problem of allocating budgets to treat confirmed patients or to develop vaccines is crucial and it is analogous to our settings. 
Refer to the supplementary material for the code and data availability. Our forthcoming work will include the following two areas:
\begin{itemize}[leftmargin=*]
\setlength\itemsep{0.3em} %
\item \emph{Determining right balance for hybrid strategies~\cite{mai2021drift}}: In this paper, the ratio between exploration and exploitation is set empirically. The model performance is sensitive to this ratio, and the performance numbers vary depending on the dataset~(Fig.~\ref{fig:exploitation-exploration-ratio}). An adaptive algorithm for selecting this ratio will manage this trade-off better. The RP1 algorithm~\cite{RP1} leverages an online learning mechanism with an exponential weight framework~\cite{exponentialweight1994} to dynamically tune this ratio, which could be applicable in our model.
\begin{figure}[h!]
    \begin{subfigure}[b]{.49\linewidth}
        \includegraphics[width=\linewidth]{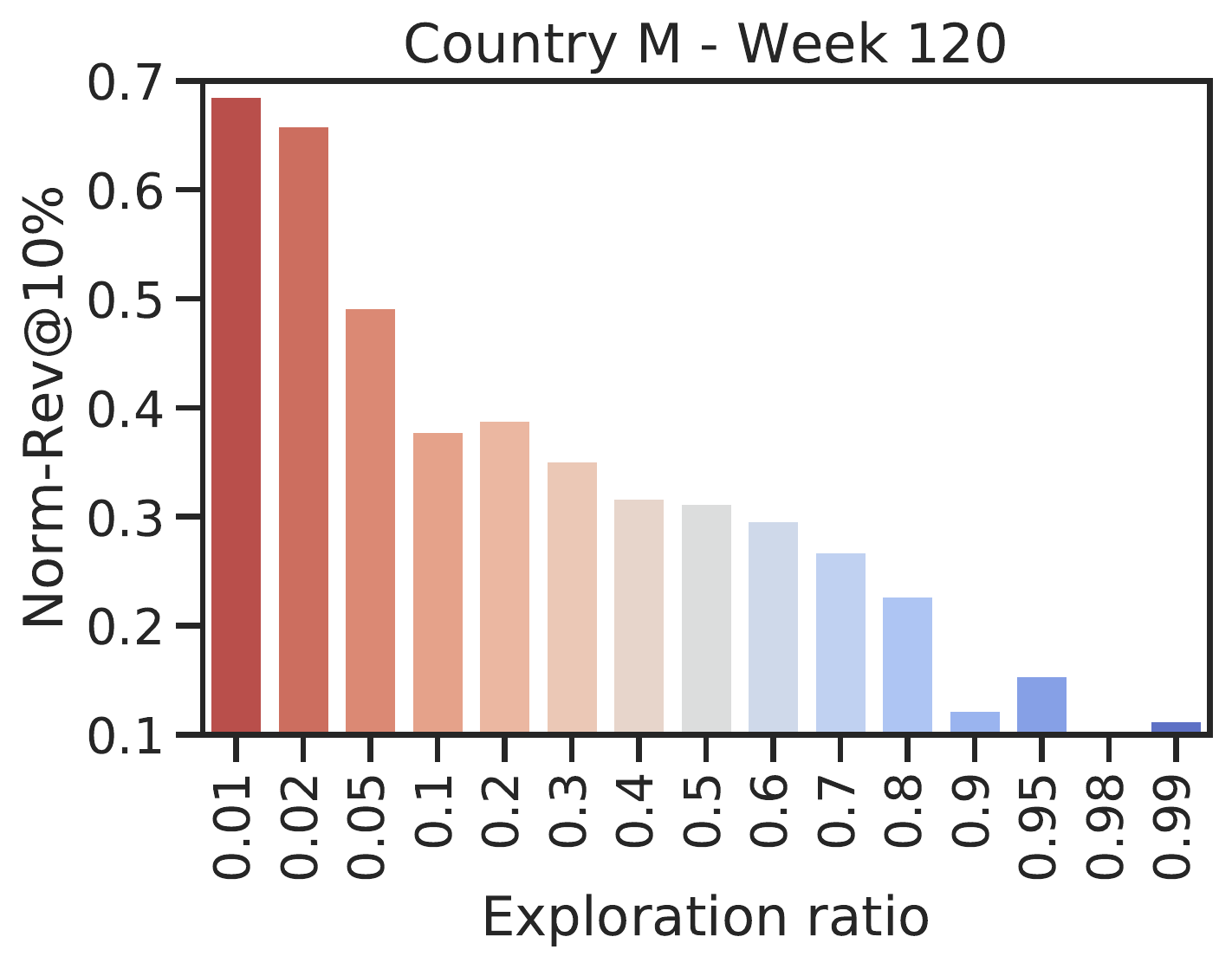}
        \centering\captionsetup{width=.9\linewidth}%
        \caption{In country M, the model performs the best with 1\% of exploration.}
    \end{subfigure}
    \begin{subfigure}[b]{.49\linewidth}
        \includegraphics[width=\linewidth]{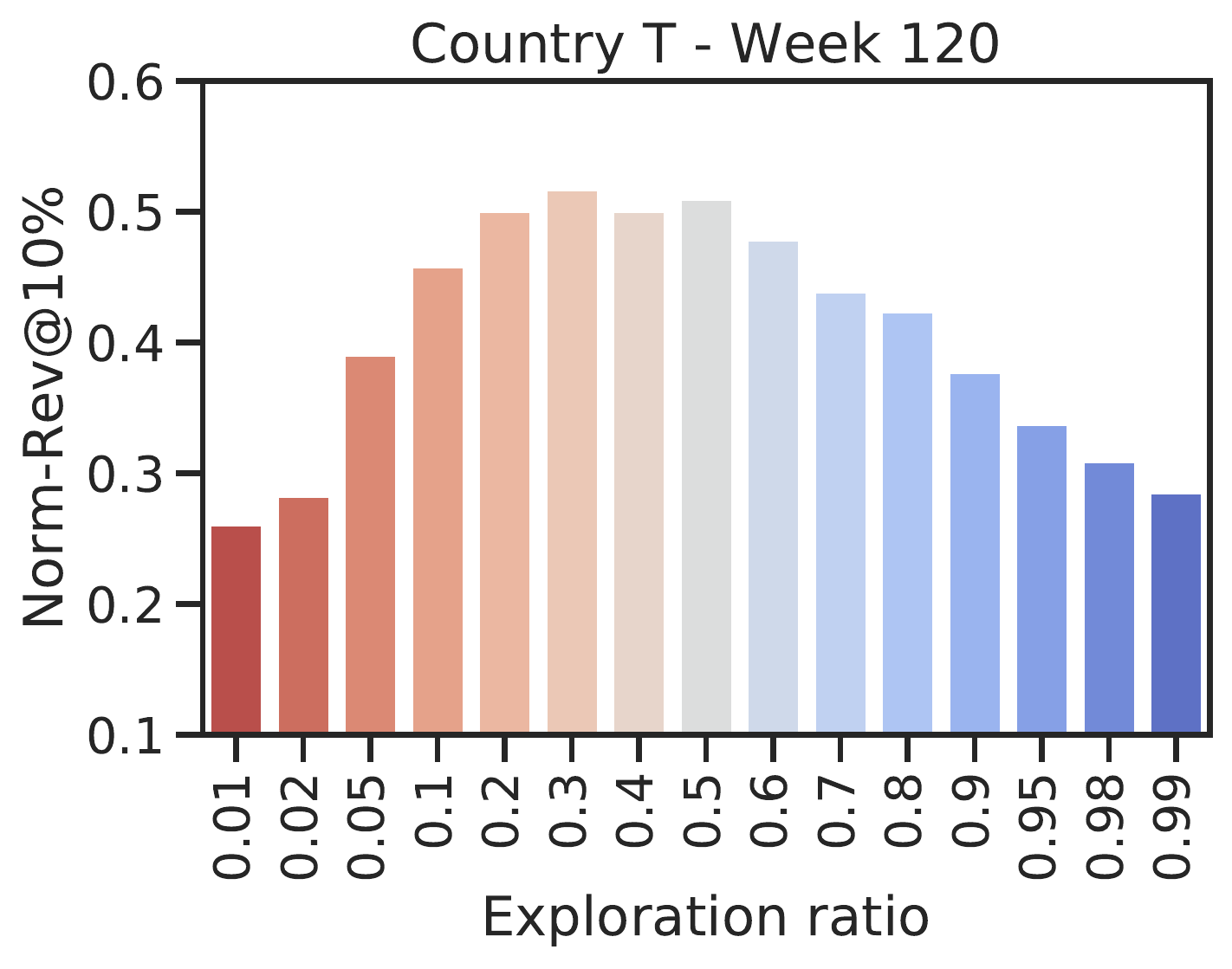}
        \centering\captionsetup{width=.9\linewidth}%
        \caption{In country T, The model performs the best with 30\% of exploration.}
    \end{subfigure}
    \caption{Best performing exploration ratio differs by data. In the case which the exploitation strategy does not work well, increasing an exploration ratio helps~(Country \textsf{T}).}
    \label{fig:exploitation-exploration-ratio}
\end{figure}
% Semi-supervised learning using both inspected and uninspected items was excluded from the scope of this research. However, h
\item \emph{Incorporating semi-supervised techniques}: Higher performance can be achieved by using richer information from a set of uninspected imports by incorporating a semi-supervised learning strategy in our framework. Building a set of augmented customs data and learning from it would be a key challenge for devising a semi-supervised learning model. 
\end{itemize}

%Together with the analysis results, the sequential prediction framework we used was disclosed. We hope that customs will use our software many, and member countries will participate in the BACUDA project to share data or model and apply AI well.

\section*{Reproducibility}
\subsection*{Code and Data Availability}
In line with this study, we prepared a GitHub repository for simulating customs selection considering the needs of customs administration. Our code is released at \sd{\url{https://github.com/Seondong/Customs-Fraud-Detection}}.

The import transaction data used in the paper cannot be made public due to nondisclosure agreements. Nevertheless, the source code runs compatibly with the synthetic data we included in the repository. % \footnote{The dataset was first released with \dat{}~\cite{kim2020date}.}
% Since the data providers are reviewing positively, we will soon release additional synthetic data based on import trades from anonymized countries. 
In the next section, we will share a step-by-step guide for running our code with synthetic data.

\subsection*{How to Run the Code}
Instructions for running the code and reproducing our experiments are as follows:

% \noindent Block use:
% \begin{pygments}{bash}
% $ wget http://tex.stackexchange.com
% \end{pygments}
% And after the block.

\begin{enumerate}[leftmargin=0.5cm]
\item Setup the Python environment: e.g., Anaconda Python \newline \texttt{\$ source activate py37}

\item Install the requirements: 
\newline \texttt{(py37) \$ pip install -r requirements.txt}

\item Run the simulation: Run \texttt{main.py} by selecting the query strategies defined in \texttt{./query\_strategies/}. The command below runs on synthetic data with a hybrid strategy consisting of 90\% \dat{} and 10\% \bate{}. By running the command, the orange line in Figure~\ref{fig:synthetic-results}(c) can be reproduced.
\newline {\small \texttt{(py37) \$ export CUDA\_VISIBLE\_DEVICES=3 \&\& python main.py --data synthetic --semi\_supervised 0 --batch\_size 512 --sampling hybrid --subsamplings DATE/bATE --weights 0.9/0.1 --mode scratch --train\_from 20130101 --test\_from 20130201 --test\_length 7 --valid\_length 28 --initial\_inspection\_rate 100 --final\_inspection\_rate 10 --epoch 10 --closs bce --rloss full --save 0 --numweeks 100 --inspection\_plan fast\_linear\_decay}}

\item Check the results: The simulation summaries are saved in \textsf{.csv} format in \texttt{./results/performances/}. The figures in this paper can be drawn by running Jupyter Notebooks in the \texttt{./analysis/} directory.

\item For further usage: the \texttt{.sh} files in the \texttt{./bash} directory will give you some ideas for running repeated experiments. See \texttt{main.py} for hyperparameter descriptions. Customs officers can simulate our strategies using their data by plugging them into the \texttt{./data} directory and adding an argument in \texttt{main.py}. The framework can support new selection strategies; The simple XGBoost selection method is found in \texttt{./query\_strategies/xgb.py}).  
\end{enumerate}

% \newpage
\subsection*{Testing on the Synthetic Dataset}
\subsubsection*{Dataset}
\begin{figure*}[h!]
\centering
    \begin{subfigure}[b]{0.44\columnwidth}
        \centering
        \includegraphics[width=\linewidth]{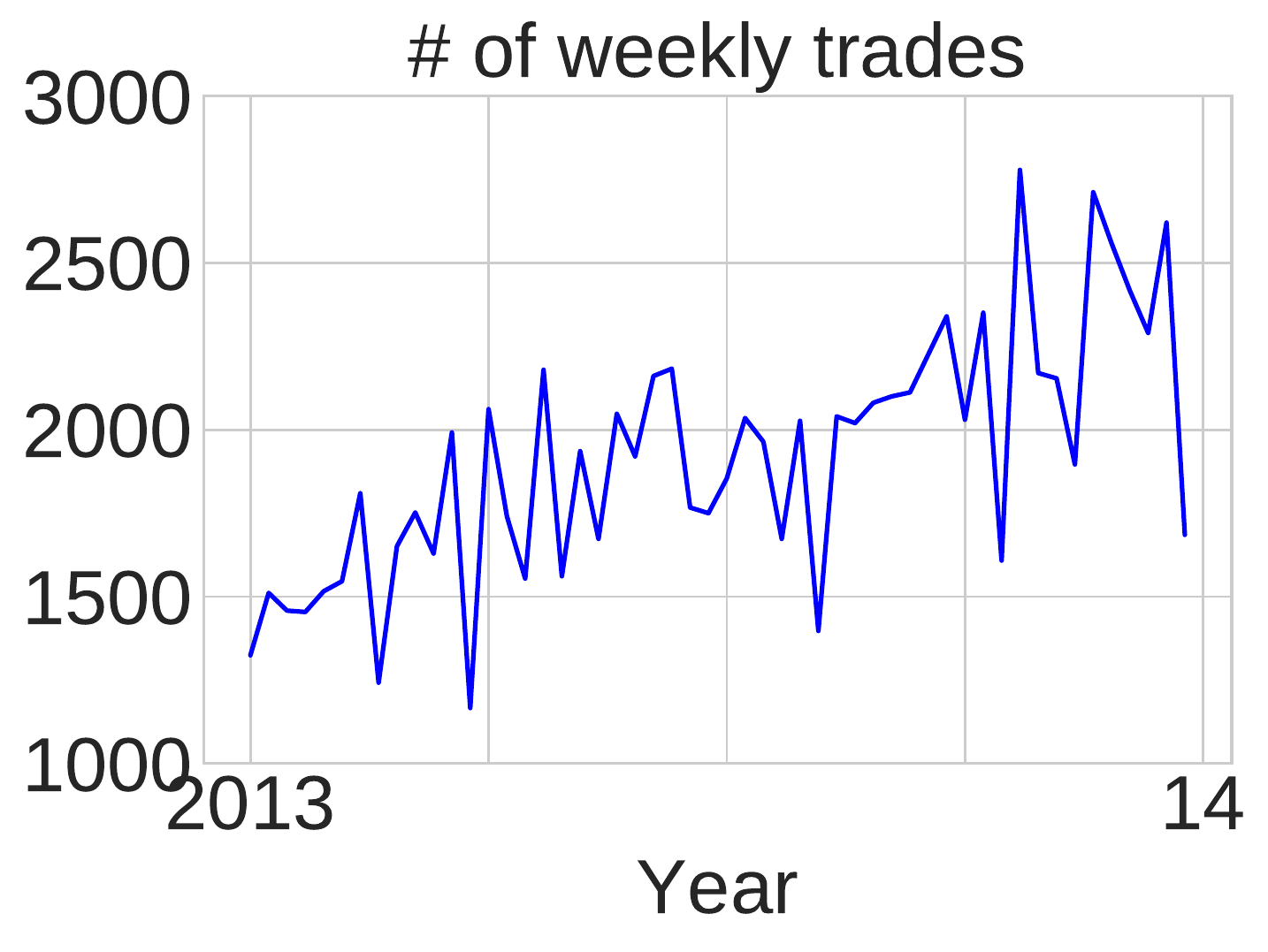}
    \end{subfigure}
    \begin{subfigure}[b]{0.44\columnwidth}
        \centering
        \includegraphics[width=\linewidth]{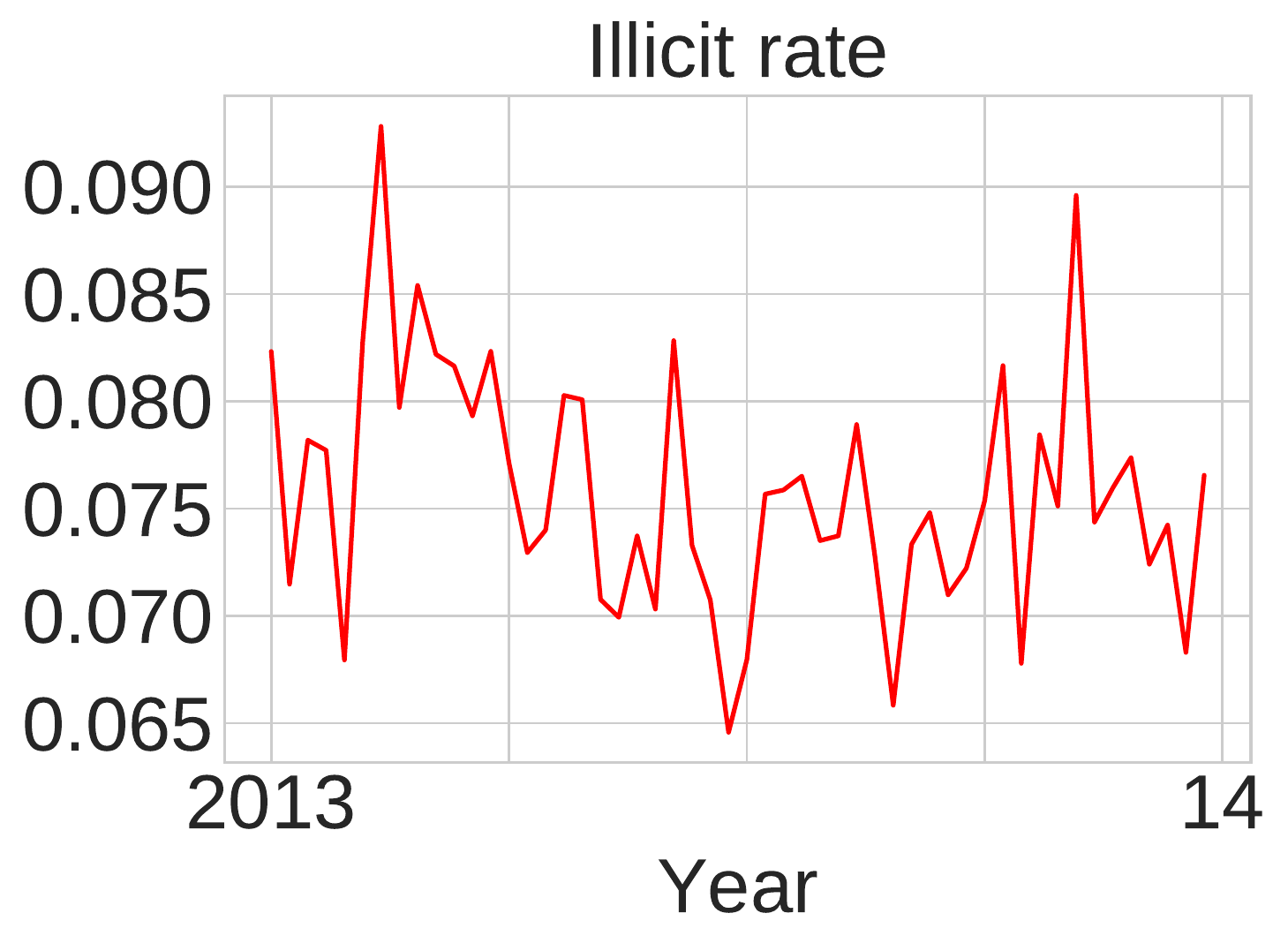}
    \end{subfigure}
    \begin{subfigure}[b]{0.42\columnwidth}
        \centering
        \includegraphics[width=\linewidth]{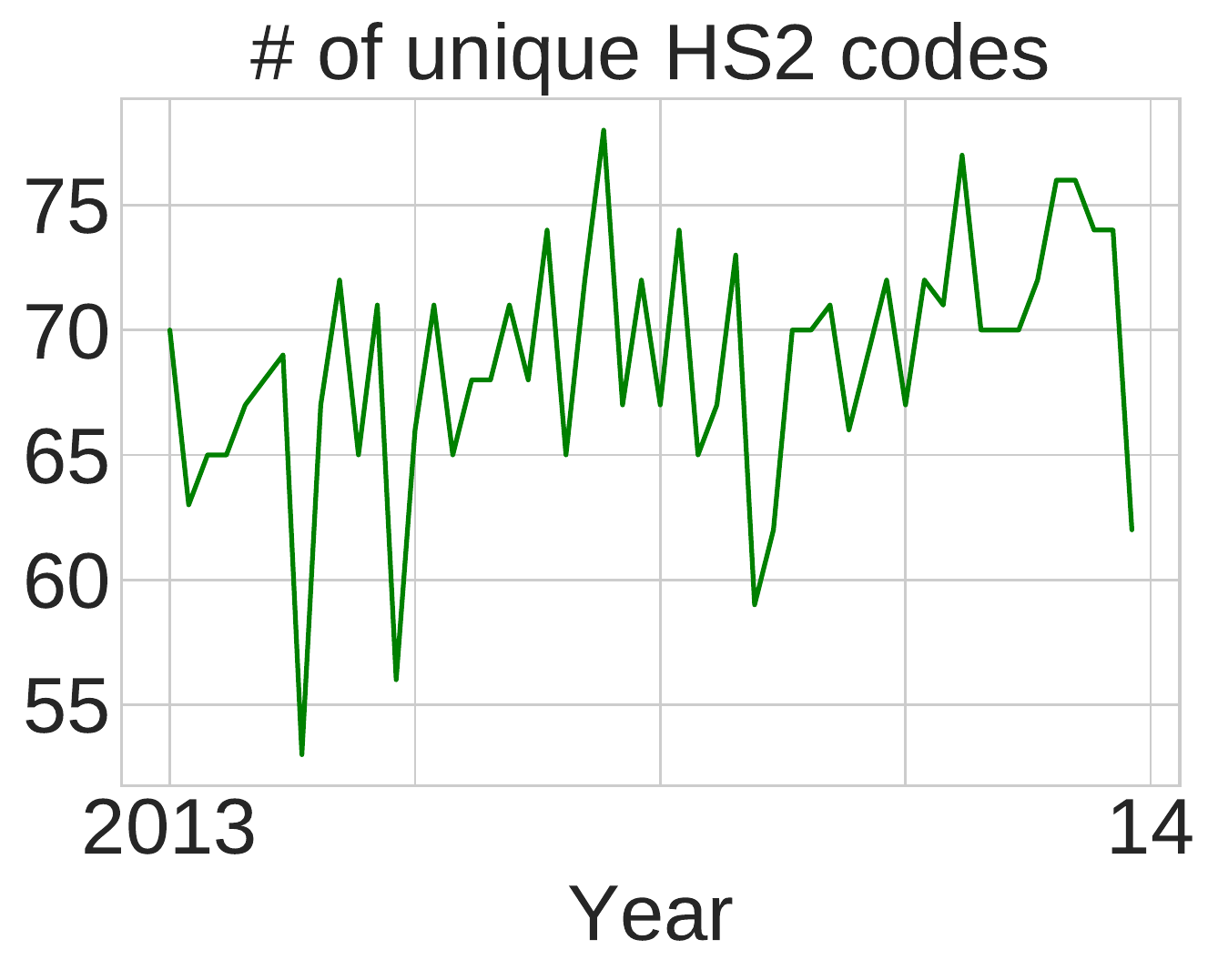}
    \end{subfigure}
    \begin{subfigure}[b]{0.44\columnwidth}
        \centering
        \includegraphics[width=\linewidth]{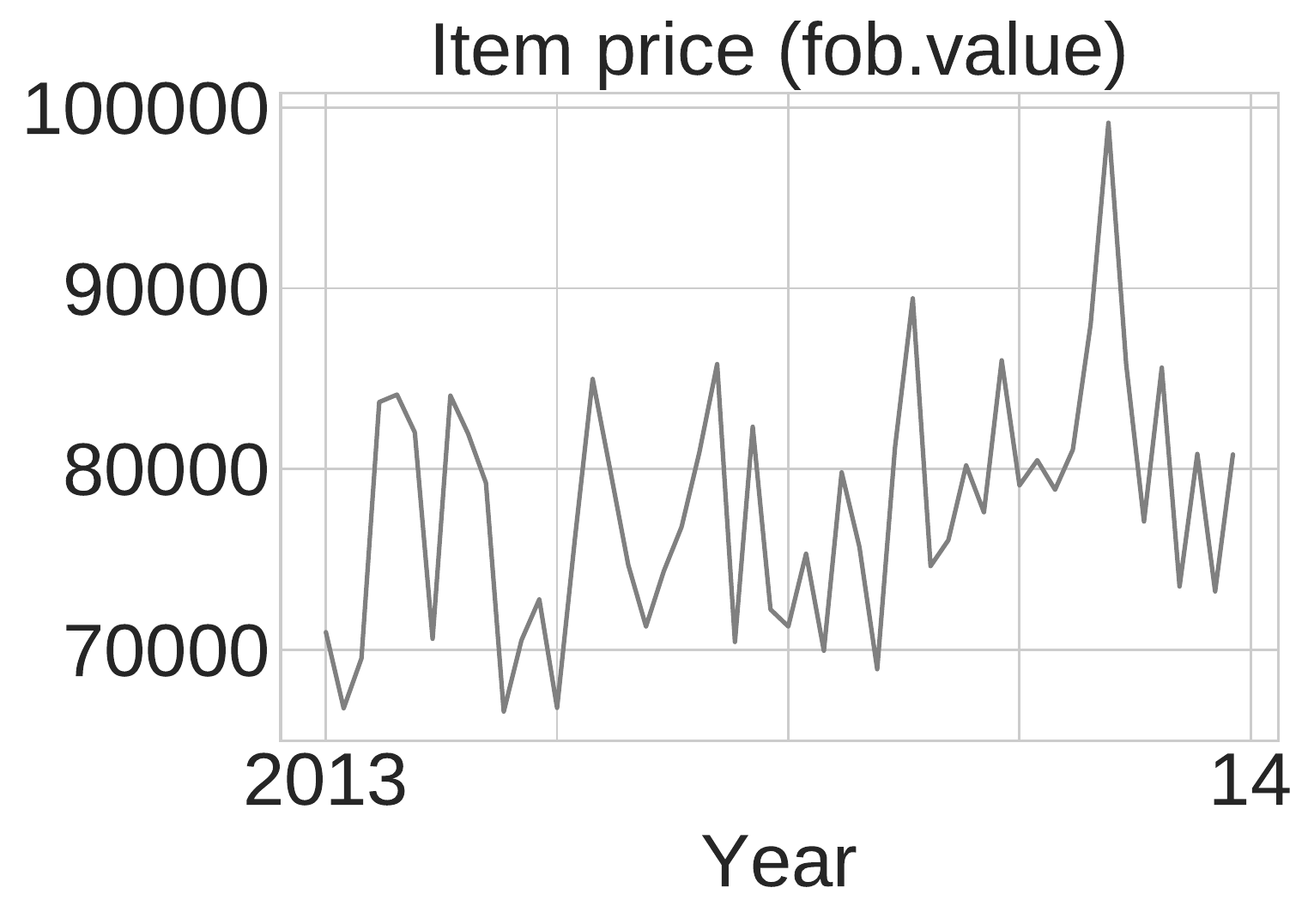}
    \end{subfigure}
    \caption{Weekly statistics of the synthetic data.}
    \label{fig:weeklystats-synthetic}
\end{figure*}

For reproducibility, we provide the experimental results using synthetic import declarations. The dataset is generated by CTGAN~\cite{xu2019} and shares similar data fields with real datasets. It consists of 100,000 artificial imports collected from Jan 2013 to Dec 2013. The number of unique importers is 8,653, and the average illicit rate is 7.6\%. Figure~\ref{fig:weeklystats-synthetic} depicts the weekly statistics of the dataset. 
\begin{figure*}[t!]
\centering
    \begin{subfigure}[b]{.30\linewidth}
        \centering\captionsetup{width=.95\linewidth}%
        \includegraphics[width=\linewidth]{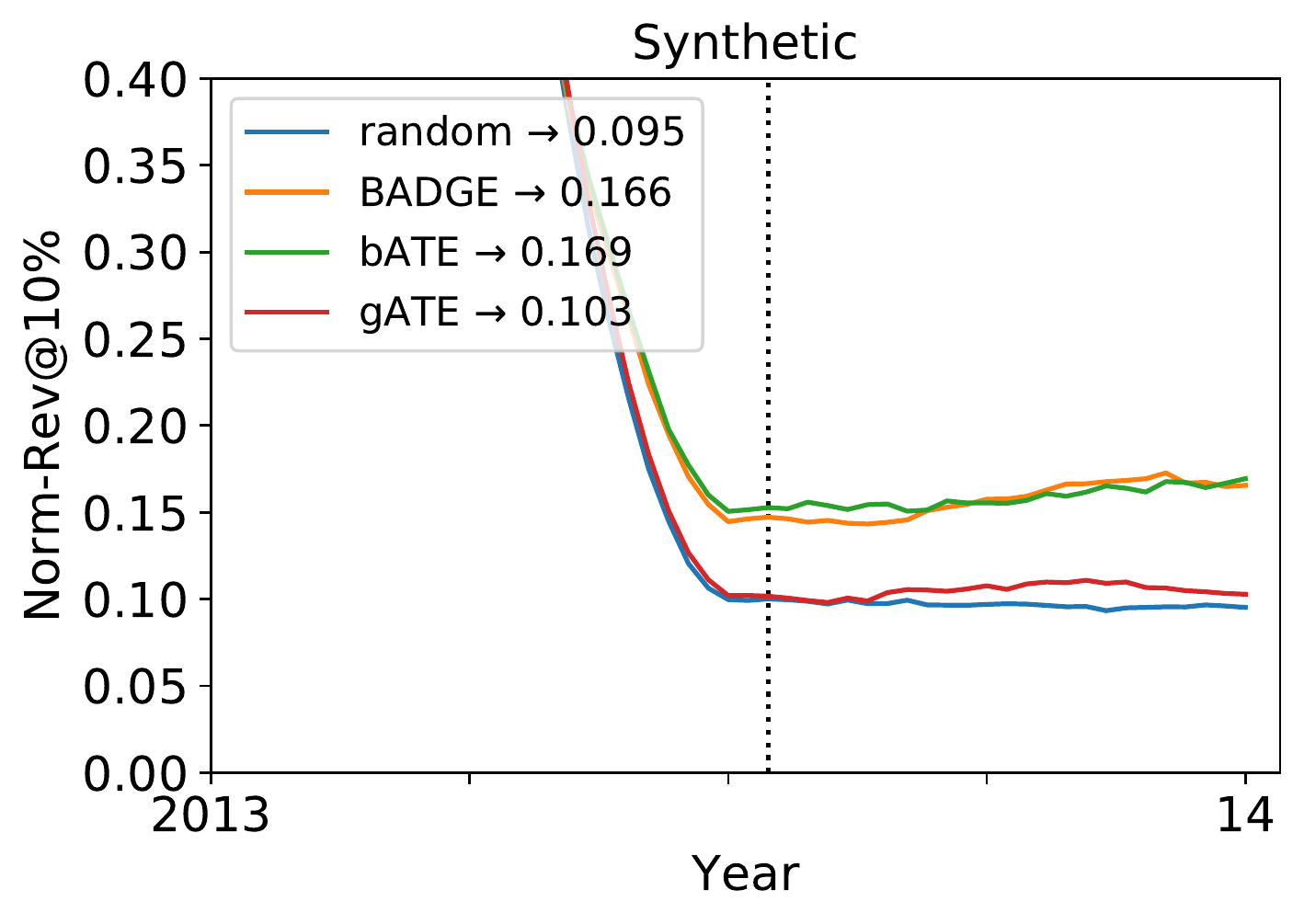}
        \caption{Pure exploration.}
    \end{subfigure}
    \begin{subfigure}[b]{.30\linewidth}
        \centering\captionsetup{width=.95\linewidth}%
        \includegraphics[width=\linewidth]{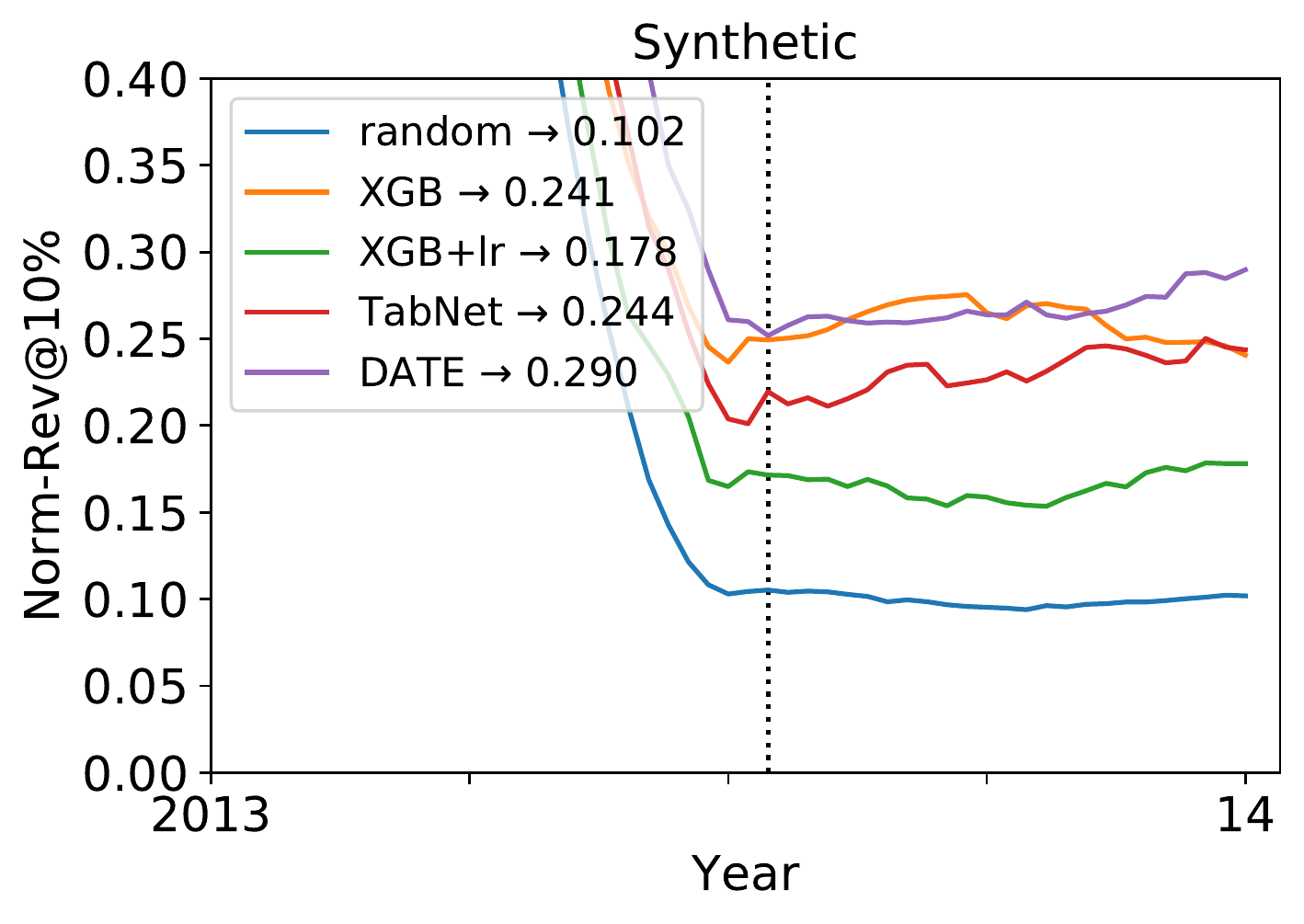}
        \caption{Pure exploitation.}
    \end{subfigure}
    \begin{subfigure}[b]{.30\linewidth}
        \centering\captionsetup{width=.95\linewidth}%
        \includegraphics[width=\linewidth]{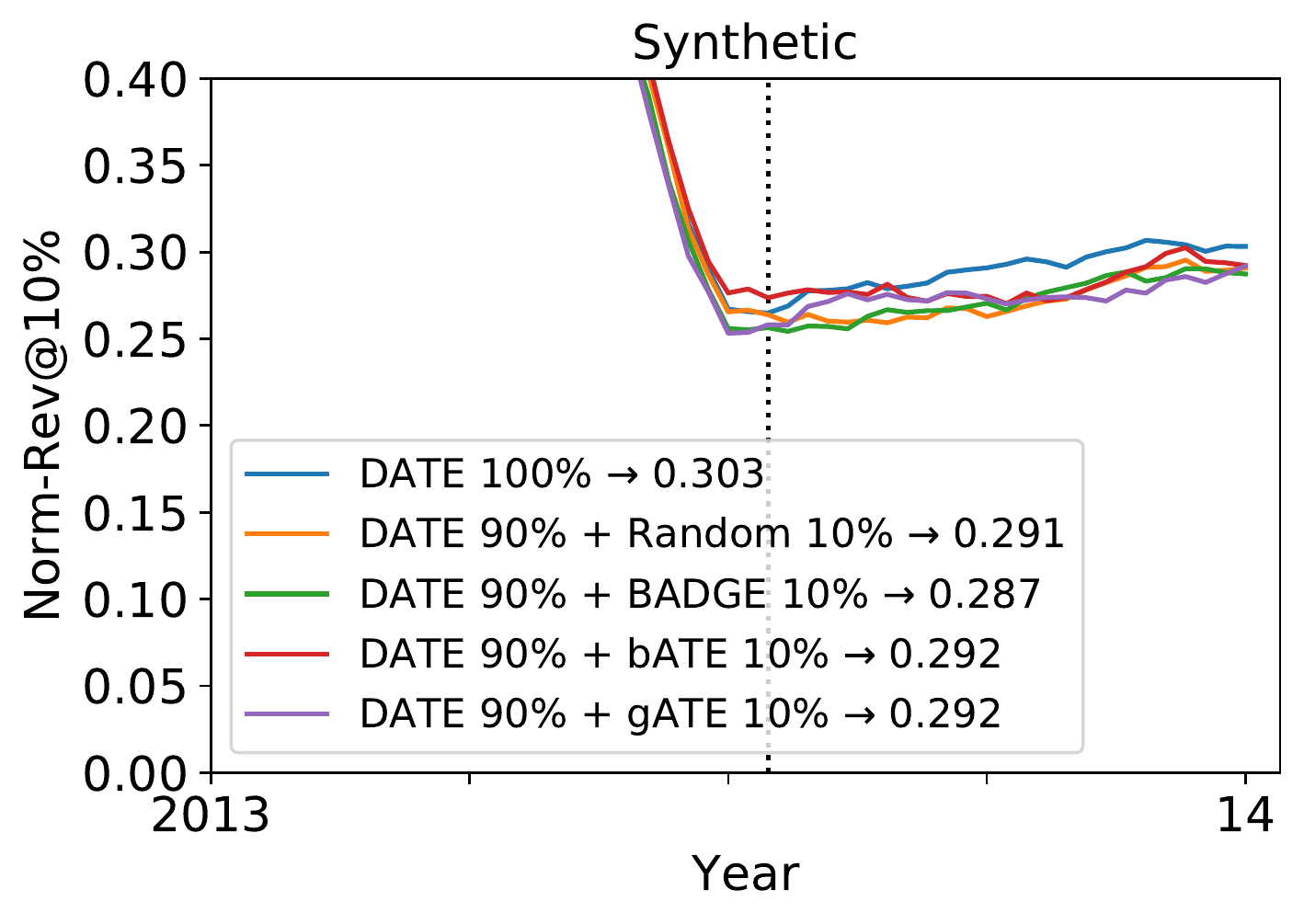}
        \caption{Hybrid strategies.} 
    \end{subfigure}
    \caption{Experimental results on the synthetic dataset.}
    \label{fig:synthetic-results}
\end{figure*}

% As shown, synthetic data의 데이터 분포 역시 꽤 다이나믹하다. Figure 10에서 볼 수 있음 
% Figure 9에 하나 더 추가할 만한 것: Synthetic data를 활용한 EDA 그림 몇개 추가해도 될듯

% \subsubsection*{Setting}
% We used the same setting in Sec.~\ref{sec:experiments:settings:scenario} with hyper-parameters listed in the above command. Additional parameters are set default in \texttt{main.py}.  
% 실험 세팅 역시 같다. 트레이닝 길이 등등 써주면 됨. 그래프 어떻게 그렸는지도 써 주면 좋음

\subsubsection*{Results} 
\label{sec:app:results}
Figure~\ref{fig:synthetic-results} shows the experimental results on synthetic data. We confirm that the synthetic data we introduce can help simulate customs selection. According to Figure~\ref{fig:synthetic-results}, the advanced model showed higher performances, supporting our statement that `Synthetic data also has its fraudulent patterns'. Looking into the details further, we can re-establish our findings from Section~\ref{sec:experiments:bestExplore}--\ref{sec:experiments:forHybrid} with the synthetic data. We run all the experiments five times and report their averages.
% \begin{aligned}
%     \forall k, \textsf{Norm-Rev@k\%} \gtrsim \textsf{Rev@k\%}, \\
%     \lim_{k \to 100\%} \dfrac{\textsf{Norm-Rev@k\%}}{\textsf{Rev@k\%}} = 1
% \end{aligned}

Figure~\ref{fig:synthetic-results}(a)--(b) compares the performance between exploration strategies and exploitation strategies. Among the exploration strategies, the state-of-the-art active learning approaches---\badge{} and \bate{}---outperform random learning by a large margin.
% \footnote{Theoretically, random selection's expected $\textsf{Norm-Rev@10\%}=0.1$.} 
Additionally, \gate{} performs nearly randomly because its default hyperparameter $\theta = 0.3$ (Algorithm~\ref{alg:exploration_strategy}) is set too high for synthetic data. 

However, pure exploration is not comparable to exploitation due to the nature of our problem. Note the large performance gap between the performance of \bate{} in Figure~\ref{fig:synthetic-results}(a) and the performance of simple XGBoost~\cite{chen2016xgboost} in Figure~\ref{fig:synthetic-results}(b). Customs administration should secure short-term revenue by inspecting the most likely fraudulent and highly profitable items and inspecting uncertain items that bring new insights for changing traffic. \dat{} is well designed for that purpose, showing its effectiveness compared to the \emph{state-of-the-art} classification models for tabular data (e.g., XGB, XGB with logistic regression~\cite{he2014adkdd}, and TabNet~\cite{TabNet}). 

Since the data length is relatively short, it is difficult to say that mixing exploration boosts the customs selection performance---Figure~\ref{fig:synthetic-results}(c)---and yet simple exploration is effective enough to be used as a component of the hybrid strategy. Moreover, the benefit of inspecting 1\% of the uncertain items is meaningful enough to compensate for the loss of not inspecting 1\% of the fraudulent items. \looseness=-1

\section*{Acknowledgment}
This work was supported by the Institute for Basic Science (IBS-R029-C2, IBS-R029-Y4). We thank the World Customs Organization (WCO) and their partner countries to support their datasets. The views and conclusions contained herein are those of the authors and should not be interpreted as necessarily representing the official policies or endorsements, either expressed or implied, of WCO.

\bibliographystyle{IEEEtran}
\bibliography{IEEEabrv, reference}
$~$





% H.~Kopka and P.~W. Daly, \emph{A Guide to {\LaTeX}}, 3rd~ed.\hskip 1em plus

% \end{thebibliography}

\vspace{-1cm}
\begin{IEEEbiography}[{\includegraphics[width=1in,height=1.25in,clip,keepaspectratio]{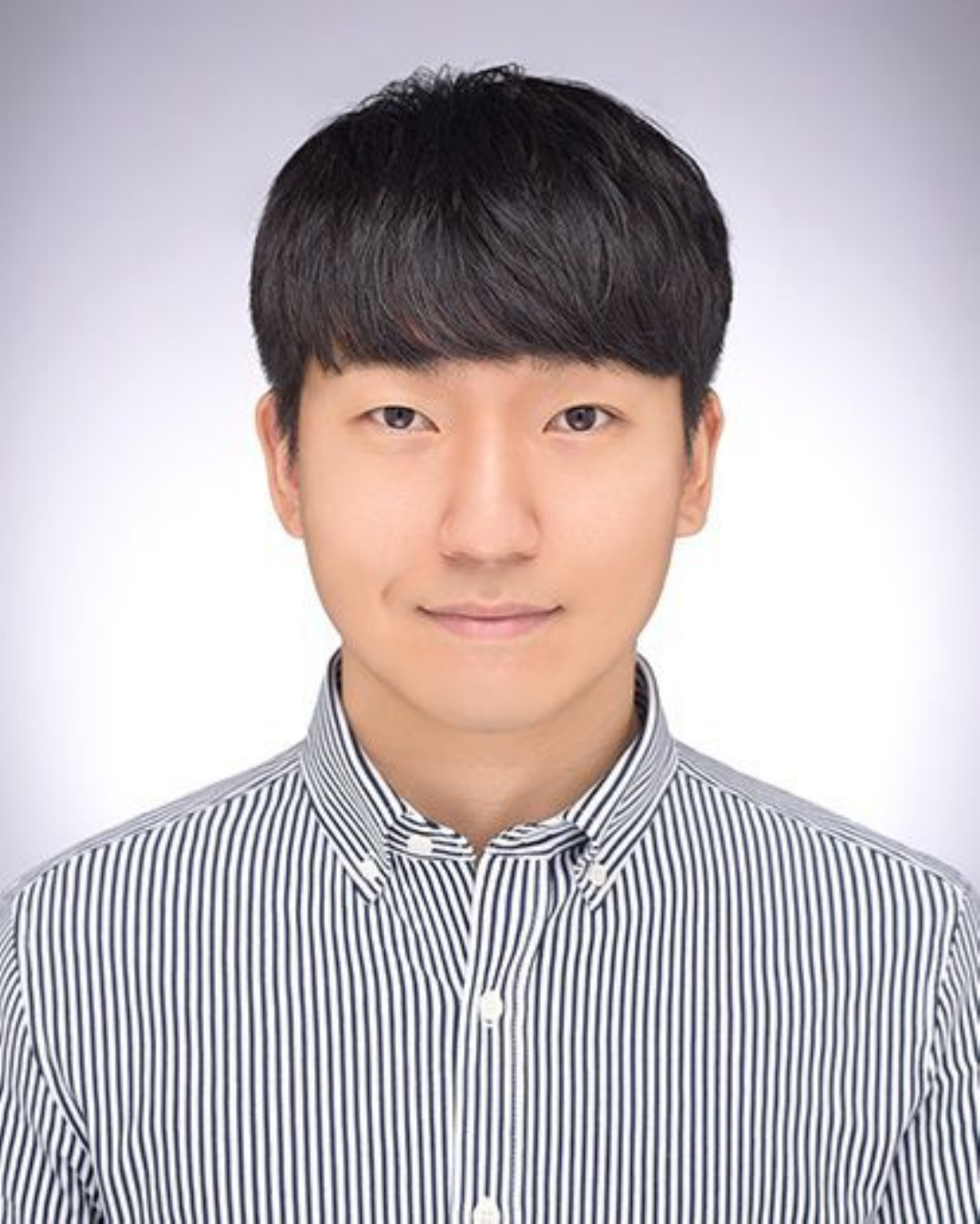}}]{Sundong Kim}
Sundong Kim is a senior researcher at Data Science Group, Institute for Basic Science. Before joining IBS, he obtained his Ph.D. at KAIST. His research interests include predictive analytics with real-world data with temporal and imbalanced in nature. He has published over 10 peer-reviewed articles in leading conferences and journals. He is a leading expert in the BACUDA initiative, developing fraud detection and category classification algorithms with the World Customs Organization. 
\end{IEEEbiography}
\vspace{-3mm}
\begin{IEEEbiography}[{\includegraphics[width=1in,height=1.25in,clip,keepaspectratio]{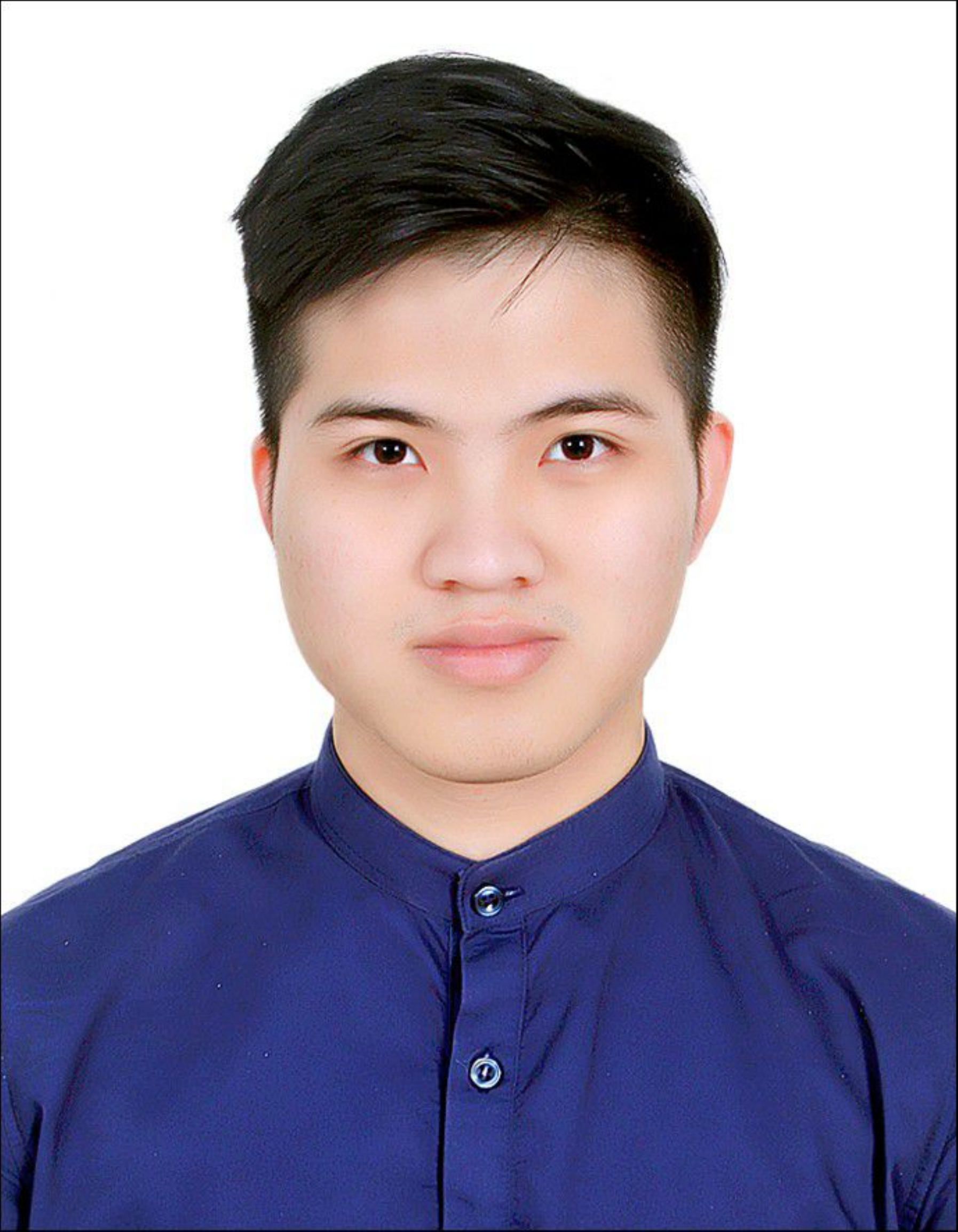}}]{Tung-Duong Mai} 
Tung-Duong Mai is an undergraduate student at School of Computing, KAIST. His research interests include predictive analytics with machine learning techniques. He has participated in the BACUDA project, developing customs fraud detection algorithms with the World Customs Organization. He worked on developing an algorithm to mitigate concept drift.
\end{IEEEbiography}
\vspace{-3mm}
\begin{IEEEbiography}[{\includegraphics[width=1in,height=1.25in,clip,keepaspectratio]{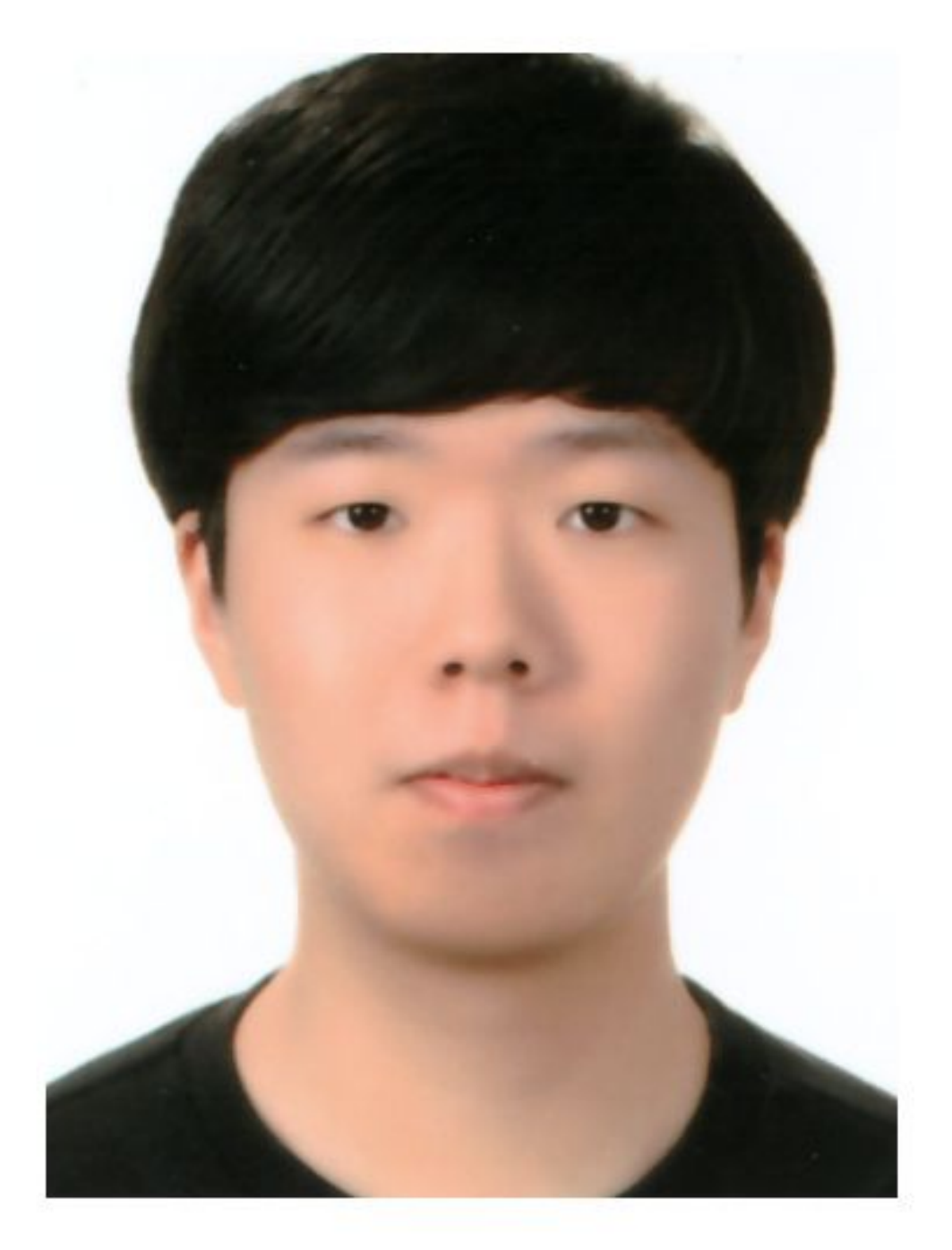}}]{Sungwon Han} 
 Sungwon Han is a Ph.D. student at School of Computing, KAIST. His research interests include developing robust representation learning algorithms to deal with data deficiency and corruption, which are common in publicly available datasets. He has worked on unsupervised learning algorithms for unstructured data and anomaly detection to discriminate out-of-distribution samples from data. He also developed a semi-supervised anomaly detection algorithm to discriminate illicit trades. 
\end{IEEEbiography}
\vspace{-3mm}
\begin{IEEEbiography}[{\includegraphics[width=1in,height=1.25in,clip,keepaspectratio]{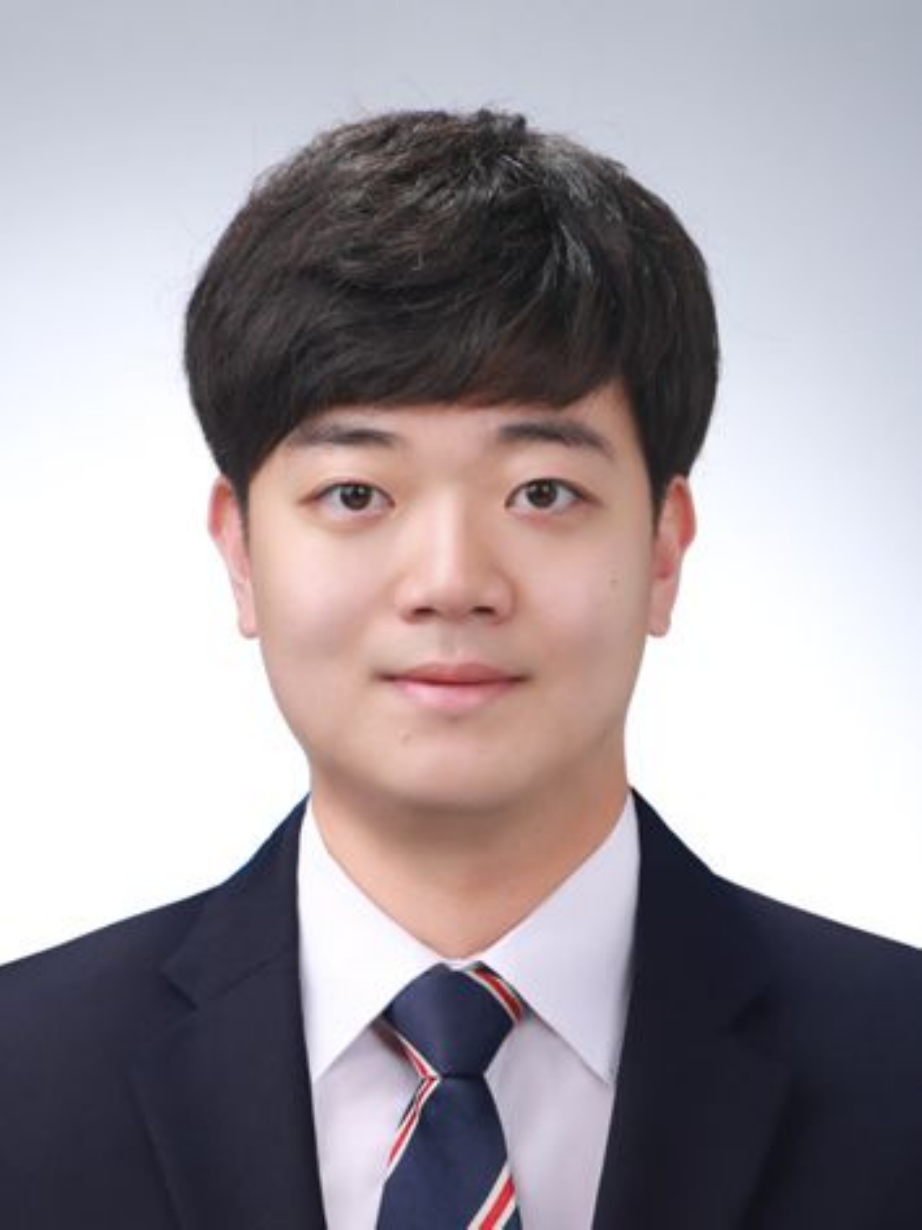}}]{Sungwon Park}
    Sungwon Park is a master student at School of Computing, KAIST. His research interests include general machine learning theory and machine learning application for social goods. He worked on a cross-national customs fraud detection model using domain generalization to support developing countries’ customs administration. % He also worked on computer vision task such as unsupervised clustering or economic scale prediction using satellite images. 
\end{IEEEbiography}
\vspace{-3mm}
% \newpage
\begin{IEEEbiography}[{\includegraphics[width=1in,height=1.25in,clip,keepaspectratio]{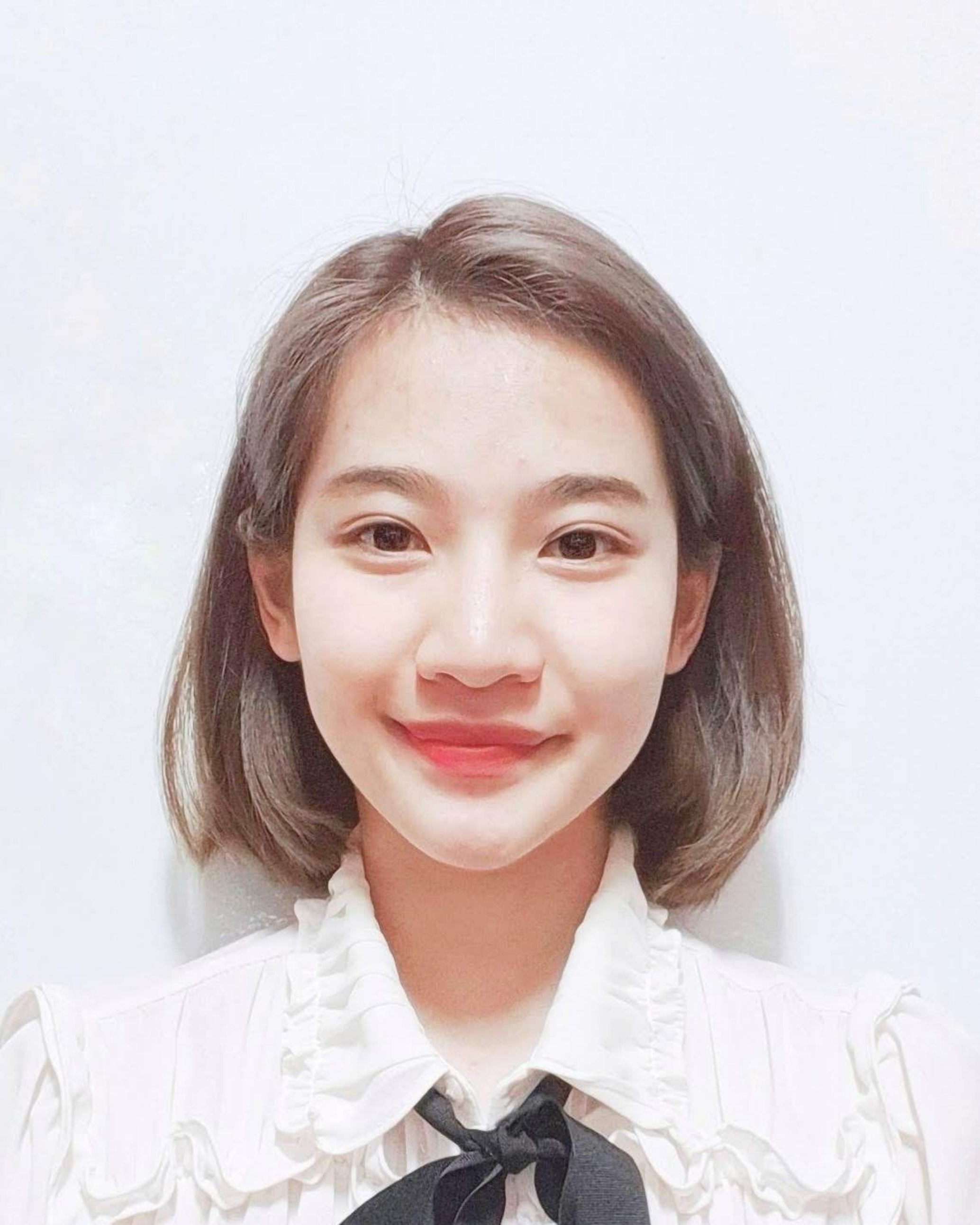}}]{Thi Nguyen D.K}
Nguyen Thi is an undergraduate student at School of Computing, KAIST. Her research interest is data science, especially prediction analytics. In this project, she worked on concept drift analysis, experimented with active customs trade selection algorithms.
\end{IEEEbiography}
\vspace{-3mm}
\begin{IEEEbiography}[{\includegraphics[width=1in,height=1.25in,clip,keepaspectratio]{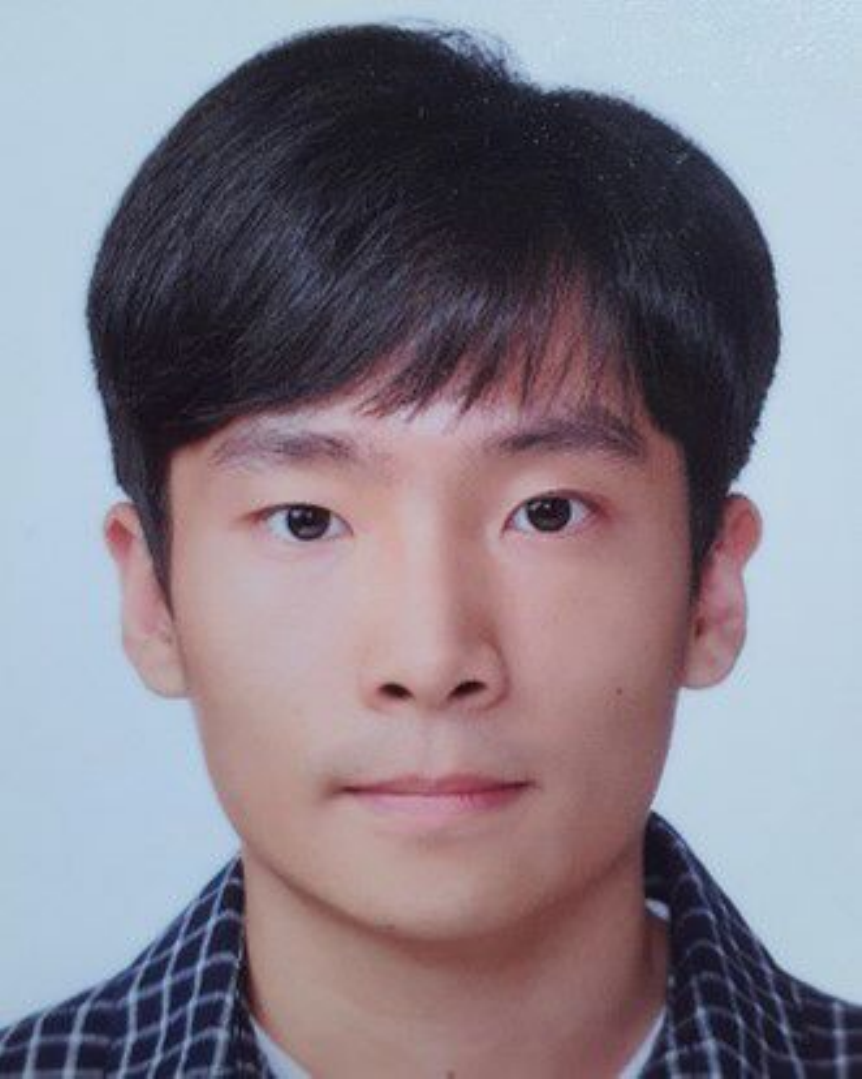}}]{Jaechan So}
Jaechan So is an undergraduate student majoring electrical engineering and computer science at KAIST. His research focuses include active learning, efficient and accurate online learning strategy, and data analysis in terms of uncertainty.
\end{IEEEbiography}
\vspace{-3mm}
\begin{IEEEbiography}[{\includegraphics[width=1in,height=1.25in,clip,keepaspectratio]{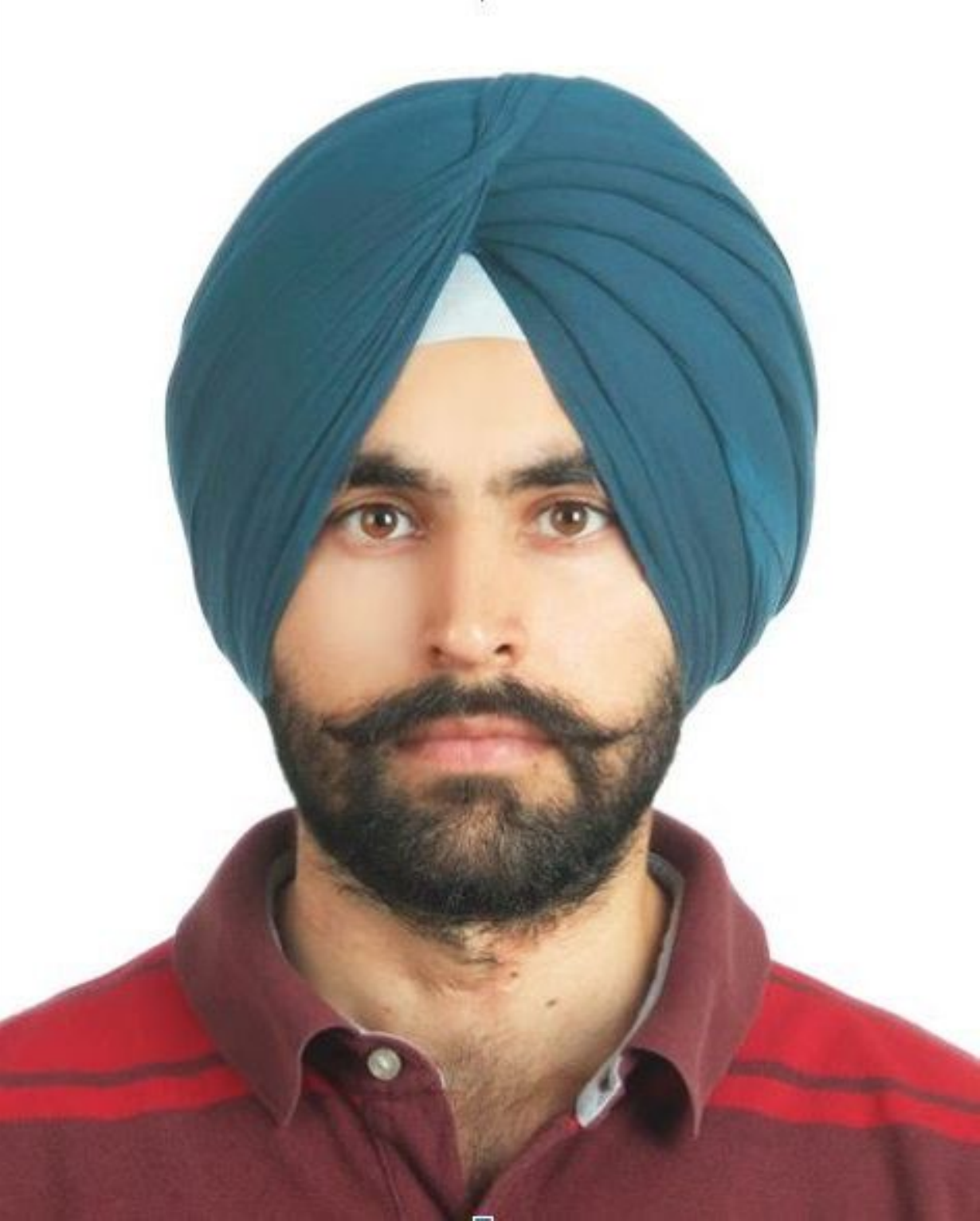}}]{Karandeep Singh}
Karandeep Singh is working as Senior Researcher at Data Science Group, Institute for Basic Science. He obtained his Ph.D. in 2019 from ETRI, Daejeon. His research interests include the application of computational techniques for networked systems and their information flows. Specific areas of application include graph-structured systems at all scales, such as interpersonal relationships in a social network, (mis)information flows on the web, and customs flows across different countries. 
\end{IEEEbiography} %molecular interactions in a protein, 
\vspace{-3mm}
\begin{IEEEbiography}[{\includegraphics[width=1in,height=1.25in,clip,keepaspectratio]{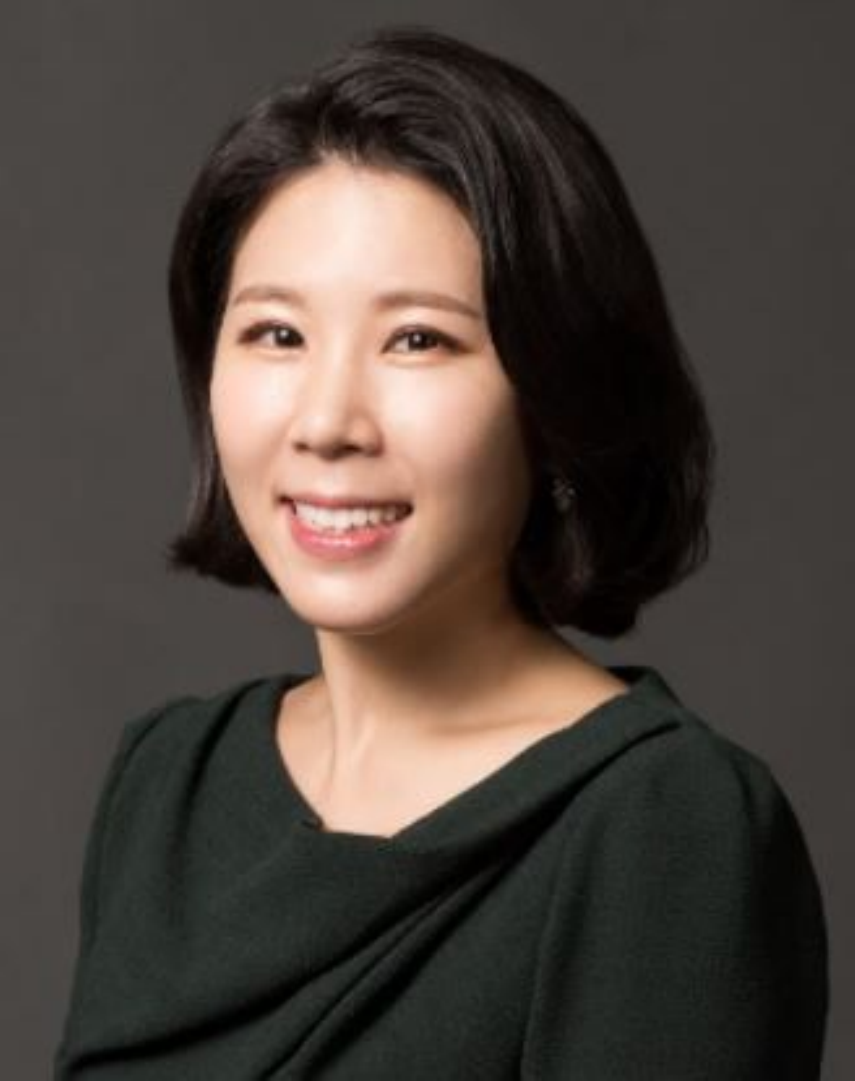}}]{Meeyoung Cha}
Meeyoung Cha is an associate professor at KAIST in the School of Computing and a Chief Investigator at the Institute for Basic Science. Her research focuses on network and data science with an emphasis on modeling, analyzing complex information propagation processes, machine learning-based computational social science, and deep learning. She has served on the editorial boards of PeerJ and ACM TSC.
\end{IEEEbiography}
\vfill

% \enlargethispage{-5in}

\end{document}